%% file: main.tex
\pdfoutput=1
\documentclass[11pt]{article}

\usepackage[]{acl}

\usepackage{times}
\usepackage{latexsym}

\usepackage[T1]{fontenc}
\usepackage[utf8]{inputenc}

\usepackage{times}
\usepackage{url}
\usepackage{multirow}
\usepackage{makecell}
\usepackage{graphicx}
\usepackage{url}
\usepackage{paralist}
\usepackage{tabularx}
\usepackage{qtree}

\usepackage{subcaption}
\usepackage{wrapfig,lipsum,booktabs}

\usepackage{colortbl}
\usepackage{siunitx}
\usepackage{color,soul}
\usepackage{makecell}

\usepackage{hhline}
\usepackage{comment}

\usepackage{adjustbox}
\usepackage{appendix}
\usepackage{placeins}
\usepackage{enumitem}
\usepackage{tikz}
\usepackage{amsmath}
\usepackage{hyperref}
\usepackage{cleveref}
\usepackage{arydshln}
\usepackage{amssymb}
\usepackage[normalem]{ulem}
\usepackage{graphicx}
\usepackage{pifont}
\usepackage{amsmath}
\useunder{\uline}{\ul}{}

\newlist{mylist}{enumerate}{1}%
\setlist[mylist]{label={\textbf{(\arabic*)}}}%

\newcolumntype{H}{>{\setbox0=\hbox\bgroup}c<{\egroup}@{}}
\newcommand\blfootnote[1]{%
  \begingroup
  \renewcommand\thefootnote{}\footnote{#1}%
  \addtocounter{footnote}{-1}%
  \endgroup
}

\newcommand{\sparrow}{SPARROW}

\title{The Skipped Beat: \\ A Study of Sociopragmatic Understanding in LLMs for 64 Languages}
\author{Chiyu Zhang$^{\xi}$ ~~~Khai Duy Doan$^{\lambda,*}$ ~~~Qisheng Liao$^{\lambda,*}$ ~~~Muhammad Abdul-Mageed$^{\xi,\lambda}$\\ 
  $^{\xi}$Deep Learning \& Natural Language Processing Group,
  The University of British Columbia\\\normalsize  $^{\lambda}$Department of Natural Language Processing \& Department of Machine Learning, MBZUAI\\ 
  \tt \{chiyuzh@mail,muhammad.mageed@\}.ubc.ca, \\
        \tt \{duy.doan,qisheng.liao\}@mbzuai.ac.ae}

\begin{document}
\maketitle
\input{abstract}
 ~\blfootnote{ $^{\star}$ {Equal contribution}}
\input{intro}
\input{related_work}
\input{benchmark}
\input{model}

\input{experiment}
\input{conclusion}
\input{limitation_ethic}
\bibliography{custom}
\bibliographystyle{acl_natbib}

\newpage
\appendix
\input{appendix}

\end{document}

%% file: abstract.tex
\begin{abstract}
Instruction tuned large language models (LLMs), such as ChatGPT, demonstrate remarkable performance in a wide range of tasks. Despite numerous recent studies that examine the performance of instruction-tuned LLMs on various NLP benchmarks, there remains a lack of comprehensive investigation into their ability to understand cross-lingual sociopragmatic meaning (SM), i.e., meaning embedded within social and interactive contexts. This deficiency arises partly from SM not being adequately represented in any of the existing benchmarks. To address this gap, we present SPARROW, an extensive multilingual benchmark specifically designed for SM understanding. SPARROW comprises $169$ datasets covering $13$ task types across six primary categories (e.g., anti-social language detection, emotion recognition). SPARROW datasets encompass $64$ different languages originating from $12$ language families representing $16$ writing scripts. We evaluate the performance of various multilingual pretrained language models (e.g., mT5) and instruction-tuned LLMs (e.g., BLOOMZ, ChatGPT) on SPARROW through fine-tuning, zero-shot, and/or few-shot learning. Our comprehensive analysis reveals that existing open-source instruction tuned LLMs still struggle to understand SM across various languages, performing close to a random baseline in some cases. We also find that although ChatGPT outperforms many LLMs, it still falls behind task-specific finetuned models with a gap of $12.19$ SPARROW score. Our benchmark is available at: \url{https://github.com/UBC-NLP/SPARROW}
\end{abstract}

%% file: intro.tex
\section{Introduction}\label{sec:intro}

\input{table_fig/data_compare}

Multilingual LLMs have recently transformed NLP, due to their powerful capabilities on a wide range of tasks~\cite{xue-2021-mt5, scao-2022-bloom}. Methods such instruction tuning and reinforcement learning from human feedback (RLHF)~\cite{ouyang-2022-instructgpt} have further enhanced the zero-shot generalizability of these models. Notably, ChatGPT exhibits impressive capabilities in this regard. Human language, however, is intrinsically ambiguous and far from solved. In fact, some forms of meaning are deeply embedded in social interactions. We collectively refer to this type of meaning as \textit{sociopragmatic meaning} (SM). To appreciate SM, consider how the meaning of an utterance in social interaction (e.g., on social media) can be highly subtle and how it incorporates both the social variation related to language users (from a sociolinguistics perspective)~\cite{tagliamonte2015making} and their communicative intentions (from a pragmatics perspective)~\cite{boxer-2021-social}. Although SM is quite established within linguistics, NLP systems still struggle with this type of meaning that is intertwined in social and interactive contexts~\cite{zhang-2021-improving}. The extent to which instruction tuned models such as ChatGPT can capture SM across languages remains largely unclear as these models are yet to be evaluated on appropriate datasets under standardized conditions easy to replicate. 

To facilitate evaluation of LLMs and enhance fairness of model comparisons and reproducibility, early work introduces evaluation benchmarks.  Most existing benchmarks, however, focus on the monolingual setting. These include GLUE~\cite{wang-etal-2018-glue}, SentEval~\cite{conneau-2018-senteval}, and TweetEval~\cite{barbieri-2020-tweeteval} for English, ARLUE~\cite{arabert-2021-abdulmageed} for Arabic, CLUE~\cite{xu-2020-clue} for Chinese, and IndoNLU~\cite{wong-2020-indonlu} for Indonesian. Although XTREME~\cite{hu-2020-xtreme} and XGLUE~\cite{liang-2020-xglue} introduce multilingual benchmarks, they only include a few SM tasks for a limited number of languages. They are also limited to standard language use (e.g., Wikipedia). \citet{barbieri-etal-2022-xlm} propose a multilingual sentiment analysis benchmark (UMSAB), but it solely contains tweet sentiment analysis datasets in only eight languages. As such, absence of a unified, diverse, and comprehensive benchmark and a fragmented evaluation landscape hamper NLP work for cross-lingual SM. 

Another challenge for SM research is the issue of \textit{data inaccessibility}~\cite{assenmacher2021benchmarking}. Although many studies release the IDs of posts (e.g., tweets), substantial amounts of these social posts become inaccessible over time due to deletion, etc.~\cite{zhang-2022-decay}. In our benchmark, we attempt to re-collect text contents of $25$ datasets by using their IDs but can only retrieve $58\%$ samples on average (see Table~\ref{tab:data_decay} in Appendix). This data decay also hinders fair comparisons in NLP for SM research. This issue has already become worse as corporations such as Twitter tighten access to their API, making it even harder to collect historical data. 
To address this bottleneck, we introduce a massively multilingual SM evaluation benchmark, dubbed~\textit{\sparrow}, that comprises $169$ datasets covering $64$ languages from $12$ language families, $16$ types of scripts, across diverse online platforms (e.g., Twitter, YouTube, and Weibo). We then perform an extensive evaluation of ChatGPT, comparing it to $13$ other models ranging in size between $110$M-$7$B parameters. Our evaluations allow us to answer multiple questions related to how it is that LLMs fare across languages on SM. To facilitate future comparisons, we also design a modular, interactive leaderboard on top of our new benchmark.

To summarize, the contributions of this paper are as follows: \textbf{(1)} We collect, uniformize, and responsibly release massively multilingual SM datasets in a \textit{new benchamark}; \textbf{(2)} Our~\sparrow~benchmark is essentially an archive of SM datasets that alleviates the serious issue of \textit{data decay}; \textbf{(3)} We evaluate a wide range of models on our~\sparrow~benchmark via fine-tuning SoTA encoder-only pretrained language models and zero-shot learning of a number of generative models, including instruction tuned models (e.g., BLOOMZ) as well as ChatGPT; and \textbf{(4)} We establish standard settings for future research in this area across a large number of languages and tasks. through a \textit{public leaderboard}. 

%% file: table_fig/data_compare.tex
\begin{figure}
  \centerline{\includegraphics[width=\linewidth]{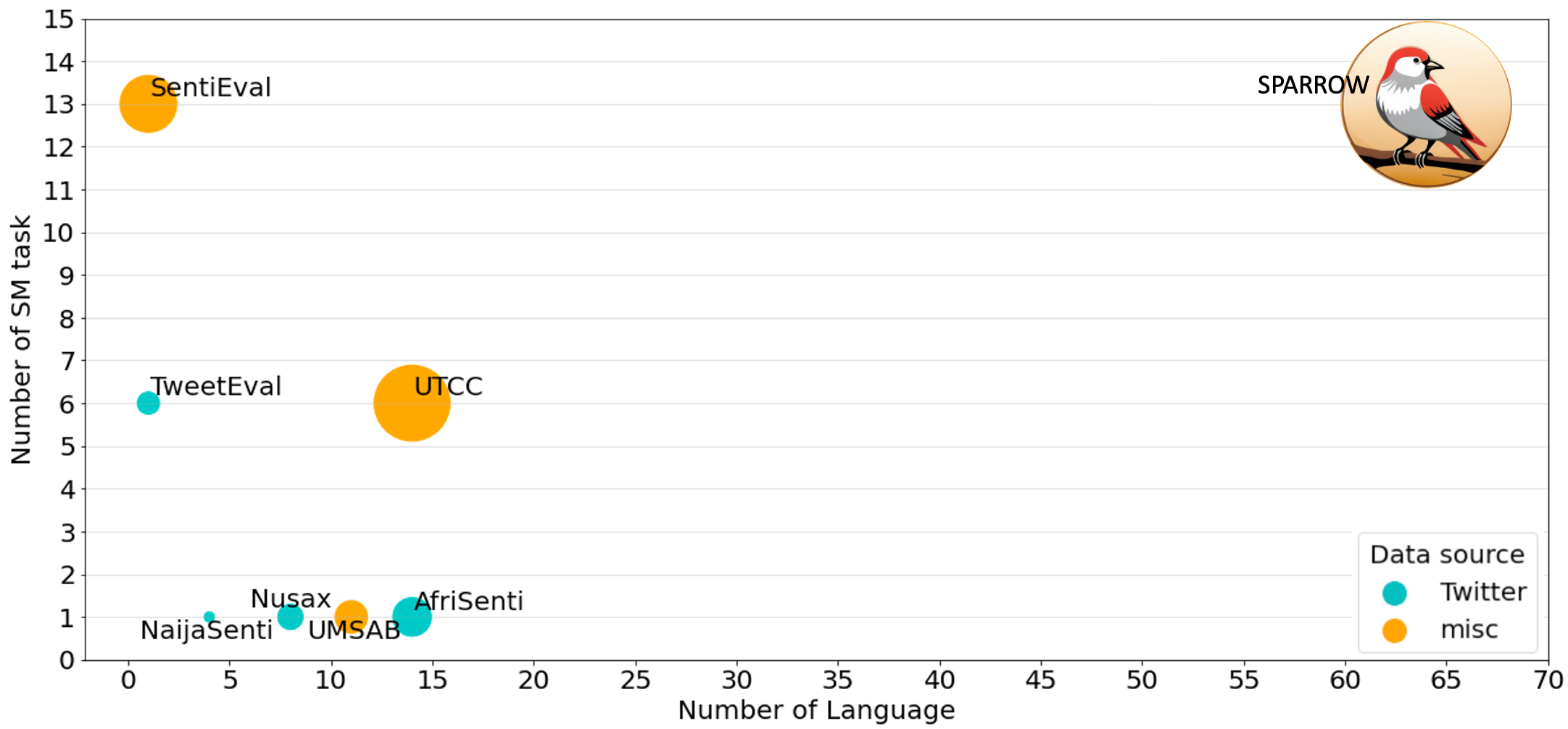}}
  \vspace{-5pt}
  \caption{Comparison of SM benchmarks with leaderboards. The bubble size indicates the number of datasets. Previous works: TweetEval~\cite{barbieri-2020-tweeteval}, UMSAB~\cite{barbieri-etal-2022-xlm}, Nusax~\cite{nusax-2020-genta}, UTCC~\cite{risch-etal-2021-toxic}, NaijaSenti~\cite{muhammad-2022-naijasenti}, AfriSenti~\cite{afrisenti-2023-muhammad}, SentiEval~\cite{sentieval-2023-zhang}.}
  \label{fig:sparrow_compare}
\end{figure}

%% file: related_work.tex
\section{Related Work}

\paragraph{Evaluation of LLMs.} 
There have been many attempts to evaluate ChatGPT and instruction tuned LLMs. \citet{qin-2023-chatgpt, systematic-2023-tahmid, zhong-2023-understand, wu-2023-lamini} utilize existing English evaluation benchmarks, such as GLUE~\cite{wang-etal-2018-glue} and BigBench~\cite{srivastava-2022-bigbench}, to evaluate LLMs' capacities on various NLP tasks. These studies find that although ChatGPT performs less effectively than the models finetuned specifically for each task, it demonstrates superior capabilities compared to other instruction tuned LLMs (e.g., FLAN~\cite{chung-2022-flan}). 
\citet{mega-2023-ahuja, multitask-2023-bang, systematic-2023-tahmid, beyond-2023-lai,huang2023languages} evaluate LLMs on more diverse languages using existing multilingual benchmarks (e.g.,  XNLI, PAWS-X, XLSum) involving monolingual NLP tasks and crosslingual tasks (e.g., machine translation). Their findings point to a large gap in performance of instruction tuned LLMs and ChatGPT, especially on low-resource languages and those with non-Latin scripts. 
\input{table_fig/related_work}

SM is still not adequately represented in existing benchmarks, hindering comprehensive evaluations on more languages. As we summarize in Table~\ref{tab:related_compare}, benchmarks used for listed evaluations only include a few SM tasks focusing on sentiment analysis. \citet{sentiment-2023-wang, sentieval-2023-zhang} investigate LLMs on a number of SM tasks (e.g., offensive language detection), but only on English. \citet{ziems-2023-can} investigate ChatGPT performance on a range of computational social science tasks covering subjects such as sociology, psychology, and linguistics, but they again focus only on English. \citet{das-2023-evaluating} extend evaluation of ChatGPT on hate speech detection to 11 languages. 
% \citet{qin-2023-chatgpt} evaluates ChatGPT on $20$ English NLP datasets encompassing seven task categories in a zero-shot setting. They find that ChatGPT demonstrates superior reasoning capabilities compared to other instructed tuned LLMs (e.g., FLAN) 
%However, they solely focus on English CSS tasks. 
Compared to these works, our objective is to investigate more diverse SM tasks on a massively multilingual setting. %To achieve this, we design SPARROW.  %We now discuss how our proposed benchmark distinguishes it from existing work. 

\paragraph{Sociopragmatic Meaning Benchmarks.}
Many previous works introduce unified benchmarks such as GLUE~\cite{wang-etal-2018-glue}, SentEval~\cite{conneau-2018-senteval}, XTREME~\cite{hu-2020-xtreme}, and XGLUE~\cite{liang-2020-xglue}. These benchmarks include a wide range of NLP tasks, but comprise a sole SM task (i.e., sentiment analysis). Some recent studies started to construct benchmarks focusing on SM: \citet{barbieri-2020-tweeteval} introduce TweetEval benchmark that contains seven English datasets of six SM tasks; \citet{sentieval-2023-zhang} develop SentiEval that involves $26$ English datasets of $13$ sentiment-related tasks. Beyond English, NusaX~\cite{nusax-2020-genta}, NaijaSenti~\cite{muhammad-2022-naijasenti}, and AfriSenti~\cite{afrisenti-2023-muhammad} propose benchmarks for sentiment analysis with eight Indonesian languages, four African languages, and $14$ African languages, respectively. UMSAB introduced by \citet{barbieri-etal-2022-xlm} contains $11$ sentiment analysis datasets in $11$ languages. For detecting antisocial online comments, \citet{risch-etal-2021-toxic} introduces a toxic comment collection that contains $43$ datasets of six antisocial detection tasks in $14$ languages. Compared to these works, our SPARROW benchmark includes significantly more SM tasks and languages, from more diverse sources (refer to Figure~\ref{fig:sparrow_compare} for a comparison).  

%% file: table_fig/related_work.tex
\begin{table}[t]
\scriptsize
\centering
\setlength\tabcolsep{2pt}
\begin{tabular}{@{}lcccccc@{}}
\toprule
\textbf{Studies}                                & \textbf{Lang.} & \textbf{Tasks} & \textbf{SM Tasks} & \textbf{Dataset} & \textbf{Models} & \textbf{LeaderBrd} \\ \midrule
\citet{zhong-2023-understand}                     & en              & 5              & 1                 & 8                & 5               & 
\ding{55}              \\
\citet{qin-2023-chatgpt}                       & en              & 7              & 1                 & 20               & 29              & \ding{55}               \\
\citet{mega-2023-ahuja}                       & 70             & 10             & 3                 & 16               & 11              & \ding{55}               \\
\citet{systematic-2023-tahmid}                & 12             & 12             & 2                 & 140              & 27              & \ding{55}               \\
\citet{multitask-2023-bang}                    & 8              & 8              & 1                 & 23               & 3               & \ding{55}               \\
\citet{beyond-2023-lai}                         & 37             & 7              & 0                 & 8                & 7               & \ding{55}               \\
\citet{das-2023-evaluating}                    & 11             & 2              & 2                 & 2                & 1               & \ding{55}               \\ 
\citet{sentiment-2023-wang}                    & en              & 5              & 5                 & 18               & 3               & \ding{55}               \\
\citet{sentieval-2023-zhang}                   & en              & 13             & 13                & 26               & 5               & \ding{51}              \\
\citet{ziems-2023-can}                          & en              & 24             & 18                & 24               & 13              & \ding{55}               \\\hline
Ours                                                            & 64             & 13             & 13                & 169              & 14              & \ding{51}              \\ \bottomrule
\end{tabular}
\caption{Our work in comparison.  }\label{tab:related_compare}
\end{table}

%% file: benchmark.tex
\section{SPARROW Benchmark}
In this section, we describe clusters of tasks in our benchmark as well as our preprocessing and unification. SPARROW consists of $13$ types of tasks in six main categories. It contains $169$ datasets from diverse online platforms and covers a wide period of time ($1986$-$2022$). We group different tasks in our benchmark by what we perceive to be an affinity between these tasks. For example, we group tasks of hate speech, offensive language, and dangerous language detection as anti-social language detection. Meanwhile, we keep particular tasks (such as sentiment analysis and emotion recognition) distinct due to the popularity of these tasks and since there are multiple datasets representing each of them. Table~\ref{tab:data_summary} summarizes statistics of SPARROW. We now briefly introduce our task clusters. We provide more information about languages in SPARROW in Table~\ref{tab:lang_sum} of the Appendix. We also provide detailed descriptions with full citations of all our datasets in Tables~\ref{tab:antisocial_data},~\ref{tab:emotion_data},~\ref{tab:humor_data},~\ref{tab:irony_data},~\ref{tab:sentiment_data}, and~\ref{tab:subj_data} in Appendix.

\input{table_fig/group_data}
\subsection{Task Clusters}

%\subsection{Antisocial Language Detection}

\paragraph{\textbf{Antisocial Language Detection.}} The proliferation of antisocial language (e.g., hate speech) toxifies public discourse, incites violence, and undermines civil society~\cite{sap2019risk, vidgen2020directions}. Antisocial language detection is thus a useful task. %task aims to identify toxic language automatically and prevent the spread of malicious content. %Thus, it is essential to develop an effective automated system to identify toxic language and prevent the spread of malicious content.
We include under the umbrella of antisocial language the following: \textbf{(1)} aggressive language~\cite{kumar-2018-aggression}, \textbf{(2)} dangerous language~\cite{alshehri-etal-2020-understanding},  \textbf{(3)} hate speech (e.g., \citet{waseem-2016-hateful, deng-2022-cold}), \textbf{(4)} offensive language (e.g., \citet{mubarak-etal-2020-overview, kralj2021slovenian}), \textbf{(5)} offense type identification (e.g.,~\citet{zampieri-2019-predicting}), and \textbf{(6)} offense target identification (e.g.,~\citet{ ousidhoum2019multilingual, jeong-2022-kold}). %More information is in Table~\ref{tab:antisocial_data} of Appendix.  

%\subsection{Emotion Recognition}
\paragraph{\textbf{Emotion Recognition.}}
Emotion affects our decision-making as well as mental and physical health~\cite{abdul-2017-emonet}. %An effective emotion recognition system can improve a wide range of applications, such as in customer relations management, disaster management, and human-computer interaction~\cite{mohammad2012emotional}. 
SPARROW includes $26$ emotion datasets in $17$ languages (e.g.,~\citet{kajava2018cross, bianchi2021feel}). %We provide descriptions of these in Table~\ref{tab:emotion_data} in Appendix. 

 %\subsection{Humour Detection}
\paragraph{\textbf{Humor Detection.}}
 Humor is a type of figurative language which induces amusing effects, such as laughter or well-being sensations. We include four humor detection datasets in four languages (e.g.,~\citet{blinov-2019-large, meaney2021hahackathon}). %We describe these datasets in Table~\ref{tab:humor_data} in Appendix. 

\paragraph{\textbf{Irony \& Sarcasm Detection.}} Irony and sarcasm also involve figurative language. An ironic/sarcastic expression intentionally uses diametric language to signify implied meaning. We include \textbf{(1)} nine irony detection datasets in seven languages (e.g.,~\citet{xiang2020ciron}), \textbf{(2)} ten sarcasm detection datasets in four languages (e.g.,~\citet{walker2012corpus}), and \textbf{(3)} an irony type identification dataset~\cite{van-hee2018semeval}. %Table~\ref{tab:irony_data} in Appendix offers detailed information about these datasets.  
 
\paragraph{\textbf{Subjectivity and Sentiment Analysis.}}
Subjectivity analysis aims to understand the opinions, feelings, judgments, and speculations expressed via language~\cite{abdul2014samar}. Our benchmark includes six subjectivity analysis datasets in five different languages (e.g.,~\citet{pang-2004-sentimental, priban-2022-czech}). Subjectivity incorporates sentiment. Sentiment analysis~\cite{poria2020beneath} is one of the most popular tasks in SM understanding where the objective is to identify the underlying sentiment of a given text. Our benchmark contains $77$ sentiment analysis datasets in $58$ languages (e.g.,~\citet{pang2005seeing, mounika-2022-resource}). %More detailed information about these sentiment and subjectivity analysis datasets are in Tables~\ref{tab:sentiment_data} and~\ref{tab:subj_data} in Appendix, respectively. 

% \subsection{Subjectivity Analysis}

\subsection{Preprocessing, Splits, and Metrics}\label{subsec:preprocess}
We apply light normalization on all the samples by converting user mentions and hyperlinks to `USER' and `URL', respectively. 
We standardize label names for consistency across datasets without reassigning nor aggregating the original labels of the datasets. For instance, in certain sentiment analysis datasets, we map `\textit{0}' and `\textit{1}' to `\textit{Negative}' and `\textit{Positive}' respectively. 
Regarding data splits, if the dataset already has Train, Dev, and Test sets, we maintain the same splits. If the original dataset does not include a Dev set, we randomly select $10\%$ of training data to be a Dev set. In cases without pre-defined splits, we use an $80\%$ Train, $10\%$ Dev, and $10\%$ Test random split. For computing constraints,  we also prepare a smaller Test set for each dataset by randomly sampling $500$ samples from Test (if its size exceeds $500$). We refer to this smaller test set as Test-S.%\footnote{We will release both Test and Test-S.} 

%\subsection{Evaluation Metrics}
We evaluate on each dataset using its original metric as Tables~\ref{tab:antisocial_data},~\ref{tab:emotion_data},~\ref{tab:humor_data},~\ref{tab:irony_data},~\ref{tab:sentiment_data}, and~\ref{tab:subj_data} in Appendix summarize.\footnote{We select the macro-average F\textsubscript{1} score as the main metric if the original paper utilizes more than one metric.} We report the performance on individual datasets, aggregate datasets into $13$ tasks, and report an average score over each task. Moreover, we introduce a metric for each main category, calculated as the average of dataset-specific metrics within that category. Inspired by previous evaluation benchmarks like GLUE~\cite{wang-etal-2018-glue}, we define a global metric called \textit{SPARROW score}, which represents the unweighted average of all dataset-specific metrics. The SPARROW score provides an overall indication of performance on SM tasks.

\input{image_fig/prompt_demon}

%% file: table_fig/group_data.tex
% \usepackage{booktabs}
% \usepackage{multirow}
\begin{table}[t]
\scriptsize
\centering
\begin{tabular}{llrrrr}
\toprule
\multicolumn{1}{c}{\textbf{}}     & \multicolumn{1}{c}{\textbf{Tasks}} & \multicolumn{1}{c}{\textbf{Dataset}} & \multicolumn{1}{c}{\textbf{Lang.}} & \multicolumn{1}{c}{\textbf{LF}} & \multicolumn{1}{c}{\textbf{Scr}} \\ \midrule
\multirow{7}{*}{\rotatebox[origin=c]{90}{\textbf{Antisocial}}}      & Aggressive                         & 1                                       & 1                                        & 1                                                 & 1                                       \\
                                  & Dangerous                         & 1                                       & 1                                        & 1                                                 & 1                                       \\
                                  & Hate                               & 16                                      & 11                                       & 6                                                 & 5                                       \\
                                  & Offense                            & 7                                       & 6                                        & 3                                                 & 3                                       \\
                                  & H/O-Group                          & 3                                       & 3                                        & 2                                                 & 3                                       \\
                                  & H/O-Target                         & 8                                       & 8                                        & 4                                                 & 7                                       \\  \cdashline{2-6}
                                  & Antisocial                                 & 36                                      & 20                                       & 7                                                 & 10                                      \\ \hline
\multicolumn{2}{l}{\textbf{Emotion}}                                            & 26                                      & 17                                       & 7                                                 & 5                                       \\\hline
\multicolumn{2}{l}{\textbf{Humor}}                                              & 4                                       & 4                                        & 1                                                 & 2                                       \\\hline
\multirow{4}{*}{\rotatebox[origin=c]{90}{\textbf{Irn\&Sarc}}} & Irony                              & 9                                       & 7                                        & 3                                                 & 3                                       \\
                                  & Sarcasm                            & 10                                      & 4                                        & 3                                                 & 3                                       \\
                                  & Irony-Type                         & 1                                       & 1                                        & 1                                                 & 1                                       \\\cdashline{2-6}
                                  & Irony\&Sarcasm                               & 20                                      & 8                                        & 3                                                 & 3                                       \\\hline
\multicolumn{2}{l}{\textbf{Sentiment}}                                          & 77                                      & 58                                       & 10                                                & 15                                      \\\hline
\multicolumn{2}{l}{\textbf{Subjectivity}}                                       & 6                                       & 5                                        & 2                                                 & 2                                       \\\Xhline{3\arrayrulewidth}
\multicolumn{2}{l}{\textbf{SPARROW}}                                            & 169                                     & 64                                       & 12                                                & 16                                      \\ \bottomrule
\end{tabular}
\caption{Summary of datasets in SPARROW. \textbf{Lang:} number of languages, \textbf{LF:} number of language families, \textbf{Scr:} number of scripts. }\label{tab:data_summary}
\end{table}

%% file: image_fig/prompt_demon.tex
\begin{figure*}
  \centerline{\includegraphics[width=\textwidth]{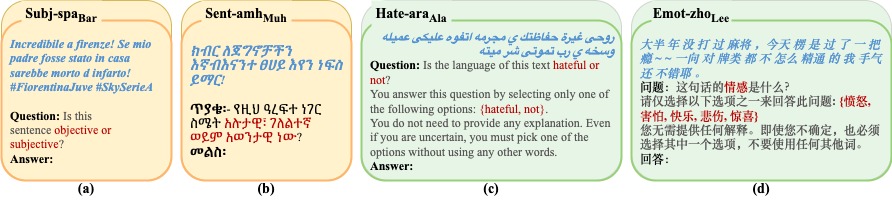}}
  \vspace{-5pt}
  \caption{Examples of prompts used for zero-shot evaluation with \texttt{lm-evaluation-harness} (\colorbox{yellow!80}{yellow}) and ChatGPT (\colorbox{green!40}{green}). We use an English prompt (Figures a, c) and machine translated the prompt in the corresponding language (Figures b, d), repectively. The prompts construct each task as question-and-answer tasks. The actual input sample is in \textcolor{blue!90}{blue}, and the label options are in \textcolor{red}{red}.} \label{fig:prompt_demo}
\end{figure*}

%% file: model.tex
\section{Evaluation Methods}
\subsection{Finetuning on Encoder-Only Models}
%\subsection{SoTA M-PLMs}
%Research on NLP has been revolutionized by transformer-encoder-based PLMs such as BERT~\cite{devlin-2019-bert}.  M-PLMs~\cite{devlin-2019-bert,conneau2020unsupervised} can learn powerful cross-lingual representations and transfer learned knowledge across languages. PLMs are initially trained on a large corpus with self-supervised objectives (e.g., masked language modelling [MLM] and next sentence predication [NSP]). After that, a PLM is fine-tuned on downstream tasks with a labeled dataset. %To evaluate the transferability of PLMs, previous work introduces evaluation benchmarks~\cite{superglue-2019-wang, wang-etal-2018-glue}. However, previous studies have not systematically investigated model capacity on SM tasks. To fill this gap, 
We evaluate the following Transformer-encoder-based multilingual models on SPARROW:
\textbf{(1) Multilingual-BERT} (mBERT)~\cite{devlin-2019-bert} (Base, $110$M parameters), \textbf{(2) XLM-RoBERTa\textsubscript{Base}} (XLM-R)~\cite{conneau2020unsupervised} ($270$M parameters), \textbf{(3) Bernice}~\cite{alexandra-2022-bernice}, a $270$M-parameter model trained with $2.5$B tweets in $66$ languages, and \textbf{(4) InfoDCL}~\cite{zhang2023contrastive}, a SoTA for SM understanding, which further trains XLM-R with $100$M tweets in $66$ languages with contrastive learning. 
More details about all models are in Appendix~\ref{append:models}. 

% \textbf{(3) XLM-Twitter} (XLM-T)~\cite{barbieri-etal-2022-xlm} takes XLM-R\textsubscript{B} model and continues pretraining it on $198$M tweets with MLM objective. 

% \textbf{(4) TwHIN-BERT}~\cite{zhang-2022-twhinbert} is trained on $7$B tweets from $100$ languages with MLM objective and a graph-based contrastive social objective. TwHIN-BERT contains $280$M parameters and a SentencePiece tokenizer with a vocabulary size of $250$K.

%We finetune each model on the full Train set of each task, identify the best model based on the model's Dev performance, and report the Test performance of the best model. 

\subsection{Zero- and Few-Shot on LLMs}
We investigate zero-shot performance on a wide range of generative models, including \textbf{\textit{pretrained} generative models}: \textbf{(1) BLOOM}~\cite{scao-2022-bloom}, \textbf{(2) mT5}~\cite{xue-2021-mt5}, \textbf{(3) LLaMA}~\cite{touvron-2023-llama}, \textbf{\textit{instruction tuned} models}: \textbf{(4) BLOOMZ}~\cite{scao-2022-bloom}, a BLOOM-initialized model tuned with multilingual xP3 corpus, \textbf{(5) BLOOMZ-P3}~\cite{muennighoff-2022-crosslingual}, a BLOOM-initialized model tuned with English-only P3 corpus, \textbf{(6) BLOOM-Bactrian}~\cite{li-2023-bactrian}, a BLOOM-initialized model tuned with 
$3.4$M instruction-following samples in $52$ languages, \textbf{(7) mT0}~\cite{muennighoff-2022-crosslingual}, an mT5 model tuned with xP3 corpus, \textbf{(8) Alpaca}~\cite{alpaca-2023-taori}, a LLaMA-initialized model tuned with $52$K English instruction-following samples, \textbf{(9) Vicuna}~\cite{vicuna2023-chiang}, a LLaMA-initialized model on $70$K conversational data, and \textbf{(10) ChatGPT}, for which we use the \texttt{gpt-3.5-turbo-0301} version via OpenAI API.\footnote{In rest of this paper, ChatGPT refers to \texttt{gpt-3.5-turbo-0301}.} We use $7$B-size version of BLOOM- and LLaMA-based models and $4$B-size version of mT5-based models. We also evaluate six open-source LLMs (i.e, BLOOM, BLOOMZ-P3, mT5, mT0, LLaMA, and Vicuna) via few-shot in-context learning. 
% For all PLMs, we utilize their public released checkpoints from Huggingface Models.\footnote{\url{https://huggingface.co/models}}

% \subsection{Intermediate Fine-Tuning for SM}
% While previous studies have trained m-PLMs with billions of social media data, these m-PLMs are not intentionally trained for capturing SM. \citet{felbo2017using, zhang-2022-improving} show that intermediate fine-tuning of a PLM on surrogate label prediction (SLP), e.g., emoji prediction objective, results in rich representations for many SM tasks. This approach also needs with significantly less data than training a domain-specific model from scratch. We thus adopt intermediate fine-tune of XLM-R on SLP. To this end, we prepare an emoji-based surrogate labeled dataset that contains a set of $1,067$ emojis in $111$M tweets from $66$ languages\footnote{Language is labeled by Twitter API. More details about this dataset are in Section~\ref{sec:data_slp} in Appendix.}. We fine-tune XLM-R on the SLP objective for ten epochs with a batch size of $4,096$ and a constant learning rate of $5e-5$. We train this model on four NVIDIA V100 $32$G GPUs, and each epoch takes $19$ hours. We refer to the resulting model as SLP. 

%% file: experiment.tex
\section{Experiments}
\subsection{Implementation}
% \textbf{Hyperparameters.}
\paragraph{\textbf{Finetuning.}} To keep computation reasonable, we randomly select $45$ datasets for hyper-parameter tuning and only tune the learning rate of each model.\footnote{For more information, refer to Section~\ref{sec:hyper} in Appendix.} For all experiments, we finetune a pretrained model with an arbitrary batch size of $32$ and sequence length of $128$ tokens. Each model is finetuned on the full Train set of each dataset for $20$ epochs (with patience $=5$) based on performance on Dev set. We run each experiment \textit{three} times with different random seeds and identify the best model on Dev in each run. We report the average performance on Test-S over three runs.\footnote{We also report the performance of Dev and standard deviations in Appendix Table~\ref{tab:finetune_more_results}.}
%The best learning rate for each model is identified based on the average score of Dev set over the $45$ selected datasets. We then use the best learning rate to fine-tune each PLM on all datasets in SPARROW and report the Test performance over three runs.

\paragraph{\textbf{Zero-shot Evaluation.}}
We perform a zero-shot evaluation on SPARROW for BLOOM-, mT5-, and LLaMA-based models using language model evaluation harness (lm-evaluation-harness~\citet{eval-harness}).\footnote{\url{https://github.com/EleutherAI/lm-evaluation-harness}} 
 While we do not tailor prompts specifically for each model, customized prompts are employed for each set of tasks. These prompts follow the structure of question-and-answer tasks, where we present sample content alongside a task-specific question, as shown in Figure~\ref{fig:prompt_demo}.  The prompts are summarized in Appendix Table~\ref{tab:prompt_new}. We then instruct the model to generate an answer based on the provided option labels. Each option label represents a potential answer, and we calculate the log-likelihood for each candidate. The prediction with the highest log-likelihood is chosen as the model's final prediction. For the evaluation of ChatGPT, we draw inspiration from previous practices for prompt design~\cite{ziems-2023-can}, and incorporate additional instructions to guide its generation of the desired labels. As shown in Figure~\ref{fig:prompt_demo}, we provide an instruction that compels ChatGPT to select a single label for the given input text without providing any additional explanation. We set the temperature to $0$ to generate \textit{deterministic and reproducible results} from ChatGPT. For a few instances, we observe that ChatGPT is unable to provide a direct answer. In these cases, we randomly assign a false label to each sample. 
In addition, we also use machine translation to translate English prompts and label names to the corresponding language of each dataset.\footnote{We use Google Translate for most languages. NLLB model is used to translate the languages of \texttt{ace, ban, bjn, bug}, and \texttt{min} because Google Translate does not cover these. The prompts of \texttt{pcm} datasets are translated by a native speaker.}

\paragraph{\textbf{Few-shot Evaluation.}} We utilize lm-evaluation-harness tool with the same prompts employed in zero-shot evaluation to explore the few-shot in-context learning abilities of open-source LLMs. Before the actual test examples, we prepend $m$ examples from the Train set. Each example consists of an input text, task-specific instruction, and the corresponding answer. We set $m$ to either 3 or 5.

\input{table_fig/group_res}

\subsection{Results}
We present the aggregated performance of Test-S on each task and main category, respectively, in Table~\ref{tab:main_res}. We also present test results on all datasets and compare to dataset-specific SoTA performance in Tables~\ref{tab:anti-social_res}, \ref{tab:emotion_res}, \ref{tab:humor_res}, \ref{tab:irony_res}, \ref{tab:sentiment_res}, and \ref{tab:subjectivity_res} in Appendix. 

% \paragraph{\textbf{Overall Performance.}} 
\paragraph{\textbf{(1) How is the overall performance over different models?}} 
\textit{All the fully finetuned models surpass the zero-shot generative models as well as ChatGPT, as shown in Table~\ref{tab:main_res}}. The most superior among the finetuned models is InfoDCL, which achieves a SPARROW score of $71.60$ and outperforms ChatGPT with $11.56$ points SPARROW score. %But the gap between ChatGPT and mBERT model is only $6.56$ SPARROW points. 
On the other hand, the open-source models (i.e., BLOOM, mT5 and LLaMA) still significantly lag behind on multilingual SM understanding with performance close to a random baseline. Meanwhile, the instruction tuned multilingual LLMs (BLOOMZ and mT0) only slightly perform better than the random baseline. 

\paragraph{\textbf{(2) Can instruction tuning enhance LLMs' ability on SM understanding?}}
\textit{Yes, but it depends on the instruction training data.} Following instruction tuning on the English-only P3 dataset, BLOOMZ-P3 demonstrates an improvement of $7.76$ SPARROW score compared to BLOOM. Also, BLOOMZ improves $5.85$ points over BLOOM (but falls short of BLOOMZ-P3). MT0 also outperforms mT5. However, there remains a substantial gap between all instruction tuned models and finetuned models. BLOOM-Bactrian performs worse than BLOOMZ and BLOOMZ-P3, which are instruction tuned with NLP tasks. This indicates that the general purpose instruction-response dataset is not very useful for SM understanding. 

To further probe how instruction tuning improves BLOOM-based models, we compare BLOOM with BLOOMZ-P3 and BLOOMZ in terms of individual tasks, finding  sentiment analysis to exhibit the most significant improvement. BLOOMZ-P3 and BLOOMZ achieve a sentiment score improvement of $16.37$ and $12.36$, respectively, based on average calculation across $77$ sentiment analysis datasets. However, BLOOM-Bactrian obtains an improvement of only $1.79$ sentiment score, perhaps implying that the Bactrian instruction-response data is not all that useful for some SM tasks. After tuning mT5 on xP3 dataset, mT0 also experiences a $13.88$ improvement in the sentiment score. These may be stemming from inclusion of five English sentiment analysis datasets in both P3 and xP3 during the training phase. For example, we observe that BLOOM, BLOOMZ, BLOOMZ-P3, mT5, and mT0 obtain an accuracy of $56.4$, $92.2$, $93.00$, $49.00$, and $76.8$ on Sent-eng\textsubscript{Soc} (not included in either xP3 or P3), respectively and that BLOOM-Bactrian still performs poorly (accuracy$=53.60$) after instruction tuning. Again, these results indicate that it is still important to include task-related datasets in the instruction tuning stage.

\begin{figure}[t]
\centering
\begin{subfigure}[]{.23\textwidth}
  \centering
  \frame{\includegraphics[width=\linewidth]{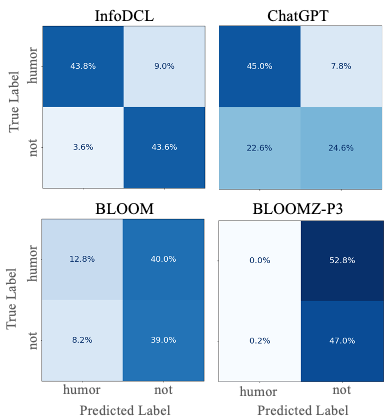}}
  \caption{Humo-spa\textsubscript{Chi}}\label{fig:confusion_humor}
\end{subfigure} 
\begin{subfigure}[]{.23\textwidth}
  \centering
   \frame{\includegraphics[width=\linewidth]{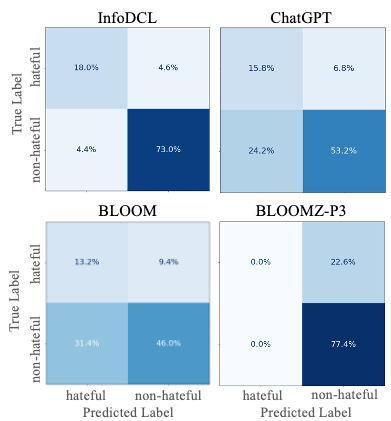}}
  \caption{Hate-ara\textsubscript{Ala}}\label{fig:confusion_hate}
\end{subfigure} 
  \caption{Confusion matrices of two datasets. }
  \label{fig:confusion}
\end{figure}

\paragraph{\textbf{(3) How do LLMs perform across different SM tasks?}} \textit{They are inferior at humor and antisocial language detection while being relatively better at sentiment and emotion recognition tasks.}
BLOOMZ-P3, BLOOMZ, and mT0 exhibit considerable enhancements ($>5$ points) on sentiment and emotion when compared to their respective initialization models. On the other hand, we find that instruction tuned models perform significantly worse on aggressive language detection and humor detection tasks. BLOOMZ-P3, BLOOMZ, BLOOM-Bactrian, and mT0 all incur a loss of more than $5$ points on these two tasks.
Upon investigating the predictions of instruction tuned models, we find that they tend to assign negative labels (i.e., non-aggressive or non-humor) which results in many false negative predictions. For a concrete example, we show that BLOOMZ-P3 predict most samples as non-humor in Figure~\ref{fig:confusion_humor} shows. 

ChatGPT outperforms the open-source LLMs on all tasks except dangerous language detection. Comparing ChatGPT to  InfoDCL, we find gaps favoring InfoDCL in subjectivity analysis (a difference of $9.47$), emotion recognition (a difference of $10.68$), and irony \& sarcasm detection (a difference of $10.70$). ChatGPT also largely lags behind InfoDCL in humor detection (a difference of $15.40$) and antisocial language detection (a difference of $14.06$). As the example shows in Figure~\ref{fig:confusion_hate}, ChatGPT makes more false positive errors (classifies non-hateful as hateful).

\input{table_fig/sample_language_wise}

% \paragraph{\textbf{Language-wise Performance.}}
\paragraph{\textbf{(4) How do LLMs perform across different languages?}}
We now examine the impact of instruction finetuning on the model's language-wise performance. We categorize the performance of each dataset based on language and calculate the average language scores across all datasets within a language. Since each language contains different tasks and datasets, a direct comparison across languages is not feasible. Therefore, we compare the relative performance between different models for each language. By comparing the instruction tuned models to their initial models, we observe that \textit{most languages experience improvement}. However, we also observe a significant decline in performance for the Amharic (amh) dataset among these models. Specifically, BLOOMZ-P3, BLOOMZ, and mT0 experience a deterioration of $36.07$, $24.99$, and $26.12$ points, respectively, compared to their respective initial models. We hypothesize that this deterioration can be attributed to catastrophic forgetting after instruction tuning, where Amharic was not included in the training set and does not share the writing scripts with the other included languages. Similarly, the Filipino (fil) tasks exhibit an average decline of approximately $11$ points on both BLOOMZ-P3 and BLOOMZ, as Filipino is not included in the xP3 dataset. Although Hindi is included in the xP3 dataset, the three instruction tuned models still show a decline in performance. Upon examining the individual performance of Hindi datasets, we find that the major deteriorations occur in the aggressive language detection and humor detection tasks, while the emotion recognition and sentiment analysis tasks show improvement. 
The instruction-response data for training Alpaca and Vicuna consist solely of English language. Therefore, we compare the performance of Alpaca and Vicuna to that of LLaMA using both English and non-English datasets. We observe that Alpaca and Vicuna outperform LLaMA when evaluated on English datasets, achieving average scores of $8.30$ and $5.51$, respectively. However, their performance declines when tested on non-English datasets, resulting in average decreases of $1.53$ and $0.33$, respectively.
Compared to task-specific InfoDCL, ChatGPT performs poorly in $63$ out of $64$ languages, sometimes with a large gap (e.g., $45.06$ lower on Amharic, $38.67$ lower on Malayalam, and $36.91$ lower on Buginese), as Table~\ref{tab:language_wise_results} shows.% However, it is worth noting that ChatGPT surpasses the task-specific finetuned InfoDCL in terms of performance on the Maltese dataset ($10.46$ points better). Compared to the fully finetuned mBERT, ChatGPT is able to outperform $14$ languages, e.g., Finnish, Moroccan Arabic, and Albanian.

We also investigate how different models perform on SM tasks across various languages. Results for two tasks, hate speech detection (top) and humor detection (bottom), are presented in Figure~\ref{fig:task-language}. The dataset for each task is grouped according to language, and the average score of each language is obtained. The relative gain of each model against the random baseline is shown, allowing us to compare across these languages.\footnote{We note that different annotation artifacts across the different languages still make direct comparisons challenging.} We observe that InfoDCL is the best model across various tasks and languages, with the exception of hate speech in Polish where ChatGPT outperforms it. As Figure~\ref{fig:task-language} shows, ChatGPT performs better for Western languages on hate speech detection. We can also observe wider gaps in hate speech detection between ChatGPT and InfoDCL on Arabic and Korean. Similarly, while ChatGPT demonstrates satisfactory performance in English humor, it remains at significant distance behind InfoDCL in Hindi humor.

\input{image_fig/language_wise_fig}

\paragraph{\textbf{(5) Do machine translated prompts help LLMs?}}
\textit{Not in general, but they do help in a few cases.} We find, in Table~\ref{tab:main_res}, that the SPARROW score of ChatGPT with machine translated prompts is $6.14$ points lower than ChatGPT with English prompts. Meanwhile, a few tasks such as humor and sarcasm acquire improvements. We also observe a similar pattern for BLOOMZ and mT0, as Table~\ref{tab:main_res} shows. The low-resource languages with non-Latin scripts experience more performance drops in general, which is in line with findings by~\citet{beyond-2023-lai}. Hebrew (heb) and Greek (ell) get the largest performance drops (over $25$ points in each case), as shown in Table~\ref{tab:language_wise_results}. %While most languages experience a performance drop by using machine translated prompts, we also see a few languages get improvement, especially Amharic gets an improvement of $27.50$. \chiyu{any reason?}  

\input{instruction_ft_models}
\input{table_fig/case_study}
\input{table_fig/fewshot_res}
\paragraph{\textbf{(6) Does GPT-4 outperform ChatGPT?}} \textit{Yes, it does.} 
We provide a study on probing GPT-4's capacities. We exploit $20$ datasets from two tasks (i.e., hate speech and humor detection) in $12$ languages, only choosing samples whose labels ChatGPT predicted incorrectly. We refer to this test set as \texttt{GPTHard} and provide samples from it to GPT-4 in their original language, employing the same English prompts as those used by ChatGPT. As Table~\ref{tab:case-study} shows, GPT-4 significantly outperforms ChatGPT (McNemar's test with $\alpha<0.01$) on $19$ datasets.\footnote{An exception is one dataset where a significance test is not possible due to small sample size ($n=2$).}  

\paragraph{\textbf{(7) Can translating input samples into English help improve ChatGPT's predictions?}}
\textit{Yes, it can}. Here, we use the non-English part of \texttt{GPTHard} ($16$ datasets). We translate these test samples into English using ChatGPT and subsequently employ the translated text and English prompt for classification. As Table~\ref{tab:case-study} shows, we acquire a noteworthy enhancement in ChatGPT's performance (McNemar's test with $\alpha<0.01$) when using the translated input. We also observe that when fed with these English-translated samples, ChatGPT is able to surpass GPT4 with the original inputs in three datasets (i.e., Hate-ara\textsubscript{Ala}, Hate-spa\textsubscript{Bas}, Hate-ara\textsubscript{Mub}). These results suggest that although ChatGPT has inferior ability on several languages in terms of detecting SM, a translate-then-detect approach may be possible. %Consistent with the findings by~\cite{jiao-2023-translator}, leveraging English as a pivotal language can enhance the machine translation performance of ChatGPT. However, to the best of our knowledge, no research has investigated its impact on classification tasks.

\paragraph{\textbf{(8) How do open-source LLMs perform with few-shot in-context learning?}}
As Table~\ref{tab:few-shot} shows, we compare three-shot and five-shot results with zero-shot results. Based on SPARROW score, we observe that few-shot learning does enhance the performance of BLOOM, mT5, LLaMA, and Vicuna. With the increasing number of shots, the performance of LLaMA and Vicuna increases. Vicuna obtains SPARROW scores of $29.36$, $39.44$, and $41.97$ with zero, three, and five shots, respectively. However, BLOOMZ-P3 and mT0 do not improve with few-shot learning. 
% We suspect this is because BLOOMZ-P3 and mT0 were finetuned only on NLP datasets, which makes them different from Vicuna that is finetuned with open-ended instructions. 
We suspect this is because the instruction finetuning of these two models only uses a zero-shot template that hurts their few-shot learning capacities. BLOOMZ-P3 and mT0 are also different from BLOOM and LLaMA in that they are finetuned on several NLP tasks only one of which is an SM task (i.e., sentiment analysis). This probably biases the behavior of these two models.

\paragraph{\textbf{(9) Are the open-source LLMs sensitive to prompts used?}} 
We carry out a study to probe the open-source LLMs' sensitivity to prompts. We curate $55$ datasets across four tasks from SPARROW and evaluate six models with the prompts we used for evaluating ChatGPT. As Table~\ref{tab:prompt_sensitive} in Appendix shows, we find that BLOOM, LLaMA, and Vicuna incur sizable performance drops ($>6$ points decrease across $55$ datasets), while BLOOMZ-P3, mT5, and mT0 demonstrate performance levels akin to those observed in previous experiments ($<2$ points different). We leave a more comprehensive evaluation of prompt sensitivity as future work.% and hope SPARROW will be a useful benchmark for such analyses. }

%We believe that the model performance is not only affected by the choice of prompts but also by which model is used and which dataset is tested on. We hope SPARROW will be a useful benchmark for investigating multilingual SM in future work including opportunities to evaluate different types of prompts.}

\section{Public Leaderboard}
To facilitate future work, we design a public leaderboard for scoring models on SPARROW. Our leaderboard is \textit{interactive} and offers \textit{rich metadata} about the various datasets in our benchmark. It also encourages users to submit information about their models (e.g., number of parameters, time to convergence, pretraining datasets). We also distribute a new \textit{modular toolkit} for finetuning or evaluating models on \sparrow. %The toolkit is built around standard tools including PyTorch~\cite{paszke2019pytorch} and HuggingFace datasets hub~\cite{lhoest2021datasets}.

% Performance on different data sources.
% Performance on different training size.

%% file: table_fig/group_res.tex
\begin{table*}[t]
\centering
\scriptsize
\setlength\tabcolsep{3pt}
\begin{tabular}{llc|cccc|ccccc:ccc:ccc:cc}
\toprule
                                           &                &   \multicolumn{1}{c}{\textbf{Rand.}} & \multicolumn{4}{c}{\textbf{Finetuning}}                                                                                         & \multicolumn{11}{c}{\textbf{Zero-shot}}                                  \\ \cmidrule(r){3-3} \cmidrule(r){4-7} \cmidrule(r){8-20} 
\textbf{}                         & \multicolumn{1}{c}{\textbf{Tasks}} & \multicolumn{1}{c}{\textbf{---}} & \multicolumn{1}{c}{\textbf{mB.}} & \multicolumn{1}{c}{\textbf{X-R}} & \multicolumn{1}{c}{\textbf{Ber.}} & \multicolumn{1}{c}{\textbf{InfoD}} & \multicolumn{1}{c}{\textbf{BM}} & \multicolumn{1}{c}{\textbf{BMZ}} & \multicolumn{1}{c}{\textbf{\begin{tabular}[c]{@{}c@{}}BMZ\\ (MT)\end{tabular}}} & \multicolumn{1}{c}{\textbf{\begin{tabular}[c]{@{}c@{}}BMZ\\ P3\end{tabular}}} & \multicolumn{1}{c}{\textbf{\begin{tabular}[c]{@{}c@{}}BM\\ Bac.\end{tabular}}} & \multicolumn{1}{c}{\textbf{mT5}} & \multicolumn{1}{c}{\textbf{mT0}} & \multicolumn{1}{c}{\textbf{\begin{tabular}[c]{@{}c@{}}mT0\\ (MT)\end{tabular}}} & \multicolumn{1}{c}{\textbf{LLa.}} & \multicolumn{1}{c}{\textbf{Alp.}} & \multicolumn{1}{c}{\textbf{Vic.}} & \multicolumn{1}{c}{\textbf{CG}} & \multicolumn{1}{c}{\textbf{\begin{tabular}[c]{@{}c@{}}CG\\ (MT)\end{tabular}}} \\ 
                                  &                                    & \multicolumn{1}{l}{}              & \multicolumn{1}{c}{110M}        & \multicolumn{1}{c}{270M}         & \multicolumn{1}{c}{270M}          & \multicolumn{1}{c}{270M}           & \multicolumn{1}{c}{7B}          & \multicolumn{1}{c}{7B}           & \multicolumn{1}{c}{7B}      &            \multicolumn{1}{c}{7B}                                       & \multicolumn{1}{c}{7B}                                                         & \multicolumn{1}{c}{4B}           &
                                  \multicolumn{1}{c}{4B}           & \multicolumn{1}{c}{4B}           & \multicolumn{1}{c}{7B}            & \multicolumn{1}{c}{7B}            & \multicolumn{1}{c}{7B}            & \multicolumn{1}{c}{175B}         & \multicolumn{1}{c}{175B}                                                        \\ \midrule
\multirow{7}{*}{\rotatebox[origin=c]{90}{\textbf{Antisocial}}}        & Aggressive                         & 43.14                             & 72.71                           & {74.64}                            & \textbf{75.45}                             & 73.96                              & 51.06                           & \textcolor{red}{15.82} & \textcolor{red}{15.82}                           & \textcolor{red}{18.72}                                                                         & \textcolor{red}{16.37}                                                                          & 53.67                           & \textcolor{red}{15.82} & \textcolor{red}{22.00}                           & \textcolor{red}{18.31}                             & 49.29                             & \textcolor{red}{25.07}                             & \textbf{63.53}                            & {54.36}                                                                           \\
                                  & Dangerours                         & 42.06                             & 62.36                           & 63.57                            & \textbf{67.13}                             & {65.23}                              & 46.87                           & 46.87 & \bf 50.84                           & 46.87                                                                         & 46.87                                                                          & 49.31                            & 46.87 & 46.87                          & 46.87                             & 46.87                             & 46.87                             & \textcolor{red}{37.93}                            & \textcolor{red}{33.68}                                                                           \\
                                  & Hate                               & 43.62 & 72.97 & 74.37 & 76.76          & \textbf{75.85}                              & \textcolor{red}{39.83}                           & \textcolor{red}{39.44} & \textcolor{red}{37.76}                           & \textcolor{red}{38.52}                                                                         & \textcolor{red}{42.23}                                                                          & \textcolor{red}{23.29}                            & \textcolor{red}{37.33} & \textcolor{red}{39.05}                           & \textcolor{red}{37.80}                             & 44.31                             & \textcolor{red}{41.59}                             & \textbf{66.06}                            & 58.74                                                                           \\
                                  & Offense                            & 39.48 & 77.53 & 75.88 & 78.45          & \textbf{78.88}                              & 41.06                           & 40.42 & \textcolor{red}{20.28}                           & \textcolor{red}{38.59}                                                                         & 40.43                                                                          & \textcolor{red}{24.99}                            & 39.90 & \textcolor{red}{21.11}                           & 39.85                             & \textcolor{red}{16.82}                             & 48.70                             & \textbf{67.31}                            & 52.70                                                                           \\
                                  & H/O-Group                          & 14.82 & 46.18 & 42.39 & \textbf{51.15} & 50.24                               & \textcolor{red}{13.63}                           & 17.26 & \textcolor{red}{14.23}                           & 21.23                                                                         & \textcolor{red}{14.81}                                                                          & \textcolor{red}{7.02}                             & 16.25 & 17.01                           & \textcolor{red}{12.35}                             & \textcolor{red}{14.13}                             & \textcolor{red}{9.26}                              & \textbf{39.66}                            & 26.74                                                                           \\
                                  & H/O-Target                         & 20.39 & 53.16 & 57.67 & \textbf{60.96} & 60.79                               & \textcolor{red}{18.73}                           & \textcolor{red}{19.03} & \textcolor{red}{18.74}                           & \textcolor{red}{16.89}                                                                         & \textcolor{red}{18.77}                                                                          & \textcolor{red}{6.69}                             & 20.58 & \textcolor{red}{17.99}                           & \textcolor{red}{19.32}                             & \textcolor{red}{16.83}                             & \textcolor{red}{17.01}                             & \textbf{35.89}                            & 28.67                                                                           \\ \cdashline{2-20}
                                  & \textbf{AS}                                 & 35.20 & 66.92 & 67.99 & \textbf{71.14} & 70.61                              & \textcolor{red}{33.70}                           & \textcolor{red}{32.80} & \textcolor{red}{27.93}                           & \textcolor{red}{31.97}                                                                         & \textcolor{red}{33.79}                                                                        & \textcolor{red}{20.14}                            & \textcolor{red}{32.02} & \textcolor{red}{28.79}                           & \textcolor{red}{31.68}                             & \textcolor{red}{30.55}                             & \textcolor{red}{34.50}                             & \textbf{56.55}                            & 47.40                                                                           \\\hline
\multicolumn{2}{l}{\textbf{Emotion}}                                            & 15.86 & 61.42 & 66.87 & 68.13          & \textbf{69.27}                              & \textcolor{red}{9.71}                            & 17.18 & \textcolor{red}{13.85}                           & \textcolor{red}{15.07}                                                                         & \textcolor{red}{15.19}                                                                          & \textcolor{red}{7.75}                             & 27.87 & 24.21                           & \textcolor{red}{15.14}                             & 31.80                             & 18.12                             & \textbf{59.58}                            & 50.85                                                                            \\\hline
\multicolumn{2}{l}{\textbf{Humor}}                                              & 49.65 & 84.35 & 85.19 & 86.75          & \textbf{87.05}                              & \textcolor{red}{41.78}                           & \textcolor{red}{33.12} & \textcolor{red}{33.82} & \textcolor{red}{33.17}                           & \textcolor{red}{33.04}                                                                         & \textcolor{red}{35.91}                                                                          & \textcolor{red}{43.60}                            & \textcolor{red}{33.12}                            & \textcolor{red}{39.78}                             & \textcolor{red}{41.72}                             & \textcolor{red}{46.19}                             & 71.65                            & \textbf{72.70}                                                                           \\\hline
\multirow{4}{*}{\rotatebox[origin=c]{90}{\textbf{I\&S}}} & Irony                              & 42.39 & 64.24 & 65.53 & 66.88          & \textbf{68.38}                             & \textcolor{red}{36.63}                           & \textcolor{red}{35.15} & \textcolor{red}{38.69}                           & 44.46                                                                         & \textcolor{red}{36.18}                                                                          & \textcolor{red}{36.52}                            & \textcolor{red}{34.69} & \textcolor{red}{33.99}                           & \textcolor{red}{40.78}                             & \textcolor{red}{27.49}                             & 47.48                             & \textbf{58.23}                            & 56.24                                                                           \\
                                  & Sarcasm                            & 45.48 & 72.41 & 73.40 & 74.78          & \textbf{74.94}                              & \textcolor{red}{43.00}                           & \textcolor{red}{41.62} & \textcolor{red}{32.23} & \textcolor{red}{32.22}                           & \textcolor{red}{41.68}                                                                         & 46.34                                                                          & \textcolor{red}{36.09}                            & \textcolor{red}{41.62}                            & \textcolor{red}{41.17}                             & \textcolor{red}{32.48}                             & 47.67                             & 65.34                            & \textbf{65.55}                                                                           \\
                                  & Irony-Type                         & 22.36 & 47.35 & 46.43 & 56.04          & \textbf{57.58}                              & \textcolor{red}{18.83} & \textcolor{red}{18.83} & \textcolor{red}{18.83}                         & \textcolor{red}{18.83}                            & \textcolor{red}{18.83}                                                                         & \textcolor{red}{18.83}                                                                          & \textcolor{red}{18.83}                            & \textcolor{red}{18.83}                            & \textcolor{red}{18.83}                             & \textcolor{red}{18.83}                             & \textcolor{red}{18.83}                             & \textbf{30.81}                            & 30.81                                                                           \\\cdashline{2-20}
                                  & \textbf{I\&S}                               & 42.93 & 67.48 & 68.51 & 70.29          & \textbf{71.12}                              & \textcolor{red}{38.92}                           & \textcolor{red}{37.57} & \textcolor{red}{34.46}                           & \textcolor{red}{41.79}                                                                         & \textcolor{red}{40.39}                                                                          & \textcolor{red}{35.42}                            & \textcolor{red}{37.36} & \textcolor{red}{32.35}                           & \textcolor{red}{39.87}                             & \textcolor{red}{29.56}                             & 46.14                             & \textbf{60.41}                            & 59.63                                                                           \\\hline
\multicolumn{2}{l}{\textbf{Sentiment}}                                          & 34.68 & 66.34 & 69.58 & 70.44          & \textbf{71.64}                              & \textcolor{red}{26.67}                           & 39.03 & \textcolor{red}{28.61}                           & 43.03                                                                         & \textcolor{red}{28.46}                                                                          & \textcolor{red}{20.77}                            & \textcolor{red}{34.65} & \textcolor{red}{32.76}                           & \textcolor{red}{27.55}                             & \textcolor{red}{25.84}                             & \textcolor{red}{25.02}                             & \textbf{60.34}                            & 54.94                                                                           \\ \hline
\multicolumn{2}{l}{\textbf{Subjectivity}}                                       & 41.41 & 72.54 & 74.45 & 74.80          & \textbf{75.73}                              & 44.12                           & \textcolor{red}{29.45} & \textcolor{red}{30.69}                           & \textcolor{red}{30.73}                                                                         & \textcolor{red}{39.65}                                                                          & \textcolor{red}{37.35}                            & 41.64 & \textcolor{red}{36.16}                           & 42.30                             & \textcolor{red}{30.44}                             & \textcolor{red}{38.73}                             & \textbf{66.26}                            & 59.33                                                                           \\ \Xhline{3\arrayrulewidth}
\multicolumn{2}{l}{\textbf{SPARROW}}                                                 & 33.47                             & 66.60                           & 69.38                            & 70.85                             & \textbf{71.60}                              & \textcolor{red}{27.94}                           & 33.79 & \textcolor{red}{27.17}                           & 35.70                                                                         & \textcolor{red}{29.45}                                                                          & \textcolor{red}{21.45}                            & 33.63 & \textcolor{red}{30.85}                           & \textcolor{red}{28.75}                             & \textcolor{red}{28.79}                             & \textcolor{red}{29.36}                             & \textbf{60.04}                            & 53.90                                                                           \\ \bottomrule
\end{tabular}
\caption{SPARROW benchmark Test-S results. We report the average of dataset-specific metrics in a task and a category, respectively. \textbf{Rand.:} random baseline, \textbf{mB.:} mBERT, \textbf{X-R:} XLM-R, \textbf{Ber.:} Bernice, \textbf{InfoD:} InfoDCL, \textbf{BM:} BLOOM, \textbf{LLa.:} LLaMA, \textbf{Alp.:} Aplaca, \textbf{Vic.:} Vicuna, \textbf{CG:} ChatGPT, \textbf{MT:} using machine translated prompts. The best performance in each setting is \textbf{Bold}. The \textcolor{red}{red font} denotes a performance lower than the random baseline. }%We report the Test performance and standard deviation of Test results. %The best performance of models with \texttt{Base} architecture is \textbf{Bold}. %\textbf{Underscore} indicates that XLM-R\textsubscript{L} outperforms the best \texttt{Base} model.}
\label{tab:main_res}
\end{table*}

%% file: table_fig/sample_language_wise.tex
\begin{table}[t]
\centering
\scriptsize
\setlength\tabcolsep{2pt}
% \renewcommand*{\arraystretch}{0.8}   %to squeeze table content
% \resizebox{!}{0.90\linewidth}{
\begin{tabular}{@{}cc|c|ccc:cc@{}}
\toprule
\multicolumn{1}{c}{\textbf{Lang}} & \textbf{Random} & \textbf{InfoDCL} & \textbf{BMZ-P3} & \textbf{mT0} & \textbf{Vicuna} & \textbf{CG} & \textbf{CG-MT} \\  \hline
amh                               & 37.95         & \bf 65.68         & \textcolor{red}{16.05}           & \textcolor{red}{22.49}        & \textcolor{red}{2.99}          & \textcolor{red}{20.62}       & \colorbox{green!30}{46.82}          \\
bug                               & 30.77         & \bf 71.55         & 34.60           & \textcolor{red}{18.27}        & \textcolor{red}{12.90}         & \colorbox{green!30}{34.63}       & 30.86          \\
ell                               & 41.24         & \bf 79.13         & 46.71           & 45.47        & 48.21         & \colorbox{green!30}{60.94}       & \textcolor{red}{34.98}          \\
eng                               & 37.90         & \bf 75.48         & 43.32           & 39.23        & 39.75         & \colorbox{green!30}{66.51}       & ---          \\
fil                               & 52.37         & \bf 79.01         & \textcolor{red}{34.47}           & \textcolor{red}{34.47}        & \textcolor{red}{34.47}         & \colorbox{green!30}{69.13}       & 66.67          \\
heb                               & 47.60         & \bf 95.80         & 71.20           & 76.60        & \textcolor{red}{40.80}         & \colorbox{green!30}{84.20}       & 57.40          \\
hin                               & 35.24         & \bf 67.55         & \textcolor{red}{28.92}           & \textcolor{red}{26.20}        & \textcolor{red}{29.06}         & \colorbox{green!30}{52.63}       & 48.30          \\
mal                               & 31.68         & \bf 82.70         & 43.84           & 41.65        & \textcolor{red}{24.85}         & \colorbox{green!30}{44.03}       & \textcolor{red}{31.44}          \\ \bottomrule
\end{tabular}
\caption{Language-wise model performance for sample languages. The complete results are in Table~\ref{tab:full_lanuage_wise} in Appendix. Best performance in each language is \textbf{bold}, and the second best is in \colorbox{green!30}{green highlight}. The \textcolor{red}{red font} denotes a performance lower than the random baseline. }\label{tab:language_wise_results}
\end{table}

%% file: image_fig/language_wise_fig.tex
\begin{figure}[t]
\centering
% \begin{subfigure}[]{.45\textwidth}
%   \centering
%   \includegraphics[width=\linewidth]{image_fig/hate_task.jpg}
%   \vspace{-25pt}
%   % \caption{Hate speech detection}
%   \label{fig:sub1}
% \end{subfigure}

% \begin{subfigure}[]{.45\textwidth}
%   \centering
%   \includegraphics[width=\linewidth]{image_fig/humor_task.jpg}
%   % \caption{Humor detection}
% \end{subfigure} 
% \\
% \begin{subfigure}[]{.48\textwidth}
%   \centering
%   \includegraphics[width=\linewidth]{image_fig/irony_task.jpg}
%   \caption{Irony detection}
% \end{subfigure}
\centerline{\includegraphics[width=\linewidth]{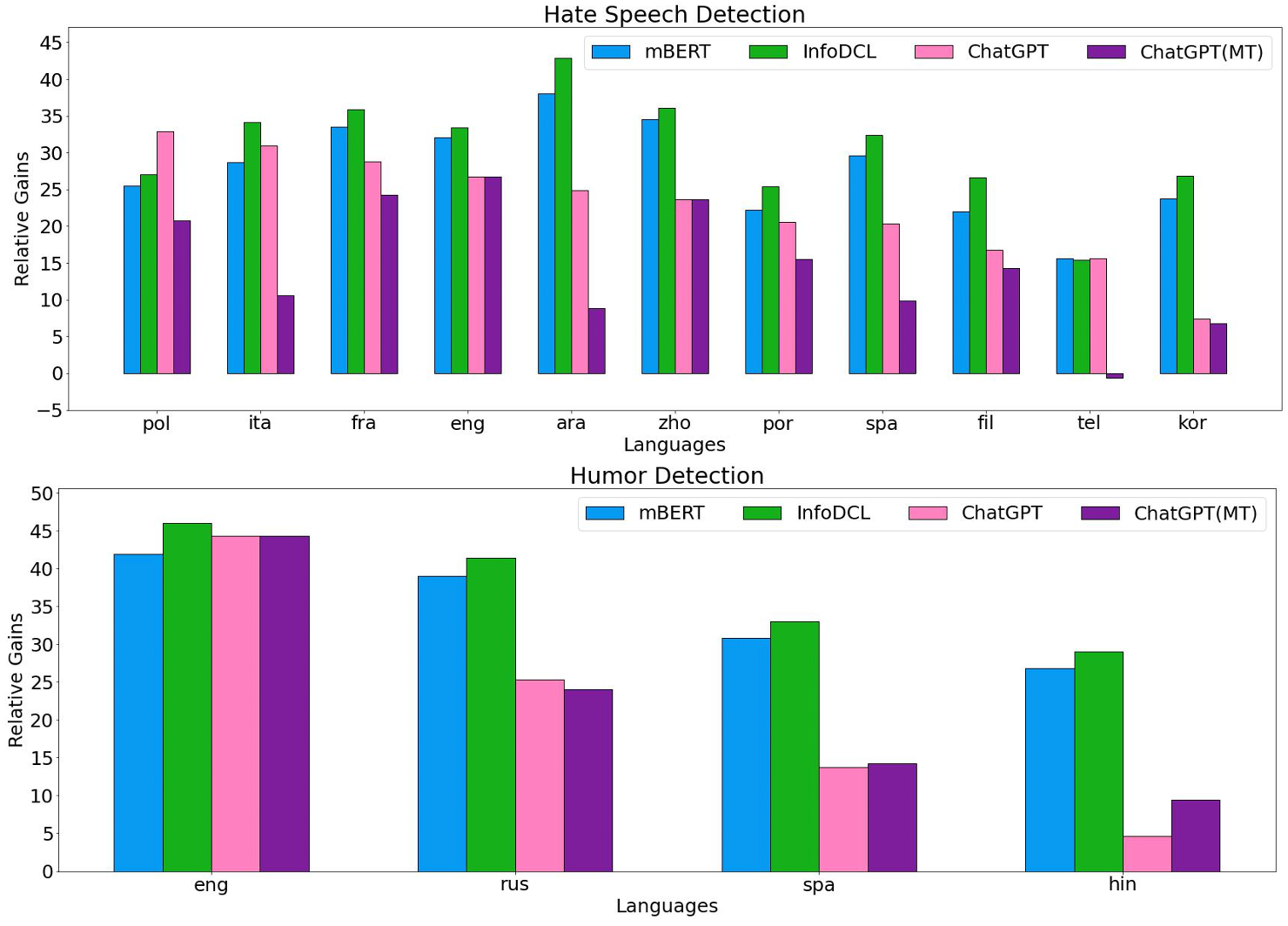}}
\vspace{-5pt}
\caption{A comparison of different models on multiple tasks across various languages. We show the relative gain of each model compared to the random baseline.}
\label{fig:task-language}
\end{figure}

%% file: table_fig/case_study.tex
\begin{table}[t]
\scriptsize
\centering
\begin{tabular}{@{}lrcc@{}}
\toprule
\multicolumn{1}{c}{\textbf{Task}}             & \multicolumn{1}{c}{\textbf{\#Sample}} & \multicolumn{1}{c}{\textbf{CG (sample MT)}} & \multicolumn{1}{c}{\textbf{GPT-4}} \\ \midrule
Hate-eng\textsubscript{Was}   & 96                                       & \multicolumn{1}{c}{---}                     & 28.82                              \\
Hate-eng\textsubscript{Dav}   & 168                                      & \multicolumn{1}{c}{---}                     & 76.30                              \\
Hate-ara\textsubscript{Ala}   & 153                                      & \textbf{38.82}                                       & 35.76                              \\
Hate-ita\textsubscript{Bos}   & 104                                      & 29.75                                       & \textbf{38.44}                              \\
Hate-fil\textsubscript{Cab}   & 145                                      & 15.76                                       & \textbf{23.96}                              \\
Hate-ara\textsubscript{Mul}   & 127                                      & 29.49                                       & \textbf{36.33}                              \\
Hate-eng\textsubscript{Bas}   & 178                                      & \multicolumn{1}{c}{---}                     & 19.23                              \\
Hate-spa\textsubscript{Bas}   & 174                                      & \textbf{21.06}                                       & 17.35                              \\
Hate-por\textsubscript{For}   & 142                                      & 35.37                                       & \textbf{37.34}                              \\
Hate-pol\textsubscript{Pta}   & 54                                       & 35.62                                       & \textbf{49.57}                              \\
Hate-kor\textsubscript{Moo}   & 250                                      & 14.20                                       & \textbf{15.85}                              \\
Hate-ara\textsubscript{Mub}   & 89                                       & \textbf{20.54}                                       & 16.74                              \\
Hate-zho\textsubscript{Den}   & 125                                      & 29.24                                       & \textbf{39.24}                              \\
Hate-kor\textsubscript{Jeo}   & 169                                      & 27.75                                       & \textbf{44.84}                              \\
Hate-tel\textsubscript{Mar}   & 2                                        & \textbf{100.00}                                      & \textbf{100.00}                             \\
Sexi-fra\textsubscript{Chi} & 113                                      & 34.10                                       & \textbf{43.69}                              \\ \hline
Humo-hin\textsubscript{Agg}  & 220                                      & 22.73                                       & \textbf{30.91}                              \\
Humo-rus\textsubscript{Bli}  & 136                                      & 26.89                                       & \textbf{47.31}                              \\
Humo-spa\textsubscript{Chi}  & 152                                      & 30.68                                       & \textbf{36.80}                              \\
Humo-eng\textsubscript{Mea}  & 48                                       & \multicolumn{1}{c}{---}                     & 50.34                              \\ \bottomrule
\end{tabular}
\caption{Case study on using machine translated input and GPT-4 on samples mispredicted by ChatGPT. }\label{tab:case-study}
\end{table}
%In comparison to the original evaluation setting of ChatGPT, all the results exhibit statistical significance (with $\alpha<0.01$), except for the results of Hate-tel\textsubscript{Mar} due to the very small number of samples. 

%% file: table_fig/fewshot_res.tex
% \usepackage{booktabs}
% \usepackage{multirow}
\begin{table*}[]
\scriptsize
\centering
\setlength\tabcolsep{3pt}
\begin{tabular}{@{}llcccccc|cccccc|cccccc@{}}
\toprule
\multicolumn{2}{c}{\multirow{2}{*}{\textbf{Tasks}}} & \multicolumn{6}{c}{\textbf{Zero-shot}}                                                                                                                                                                                                                        & \multicolumn{6}{c}{\textbf{Three-shot}}                                                                                                                                                                                                                       & \multicolumn{6}{c}{\textbf{Five-shot}}                                                                                                                                                                                                                        \\ \cmidrule(l){3-8} \cmidrule(l){9-14} \cmidrule(l){15-20} 
\multicolumn{2}{c}{}                                & \multicolumn{1}{c}{\textbf{BM}} & \multicolumn{1}{c}{\textbf{\begin{tabular}[c]{@{}c@{}}BMZ\\ P3\end{tabular}}} & \multicolumn{1}{c}{\textbf{mT5}} & \multicolumn{1}{c}{\textbf{mT0}} & \multicolumn{1}{c}{\textbf{LLa.}} & \multicolumn{1}{c}{\textbf{Vic.}} & \multicolumn{1}{c}{\textbf{BM}} & \multicolumn{1}{c}{\textbf{\begin{tabular}[c]{@{}c@{}}BMZ\\ P3\end{tabular}}} & \multicolumn{1}{c}{\textbf{mT5}} & \multicolumn{1}{c}{\textbf{mT0}} & \multicolumn{1}{c}{\textbf{LLa.}} & \multicolumn{1}{c}{\textbf{Vic.}} & \multicolumn{1}{c}{\textbf{BM}} & \multicolumn{1}{c}{\textbf{\begin{tabular}[c]{@{}c@{}}BMZ\\ P3\end{tabular}}} & \multicolumn{1}{c}{\textbf{mT5}} & \multicolumn{1}{c}{\textbf{mT0}} & \multicolumn{1}{c}{\textbf{LLa.}} & \multicolumn{1}{c}{\textbf{Vic.}} \\ \midrule
\multirow{7}{*}{\rotatebox[origin=c]{90}{\textbf{Antisocial}}}      & Aggressive      & 51.06                           & \textcolor{red}{18.72}                                                                         & 53.67                            & \textcolor{red}{15.82}                            & \textcolor{red}{18.31}                             & \textcolor{red}{25.07}                             & 46.43                           & 43.93                                                                         & \textcolor{red}{40.73}                            & \textcolor{red}{41.88}                            & 43.70                             & 44.53                             & 47.92                           & 47.63                                                                         & \textcolor{red}{33.80}                            & \textcolor{red}{37.52}                            & 50.66                             & \textbf{55.27}                    \\
                                  & Dangerours      & 46.87                           & 46.87                                                                         & \textbf{49.31}                   & 46.87                            & 46.87                             & 46.87                             & 46.87                           & 46.87                                                                         & \textcolor{red}{45.68}                            & 46.87                            & 46.87                             & 46.87                             & 46.87                           & 46.87                                                                         & 48.91                            & 46.87                            & 46.87                             & 46.87                             \\
                                  & Hate            & \textcolor{red}{39.83}                           & \textcolor{red}{38.52}                                                                         & \textcolor{red}{23.29}                            & \textcolor{red}{37.33}                            & \textcolor{red}{37.80}                             & \textcolor{red}{41.59}                             & \textcolor{red}{38.83}                           & \textcolor{red}{38.30}                                                                         & \textcolor{red}{39.43}                            & \textcolor{red}{37.82}                            & \textcolor{red}{43.51}                             & \textbf{49.17}                    & \textcolor{red}{37.95}                           & \textcolor{red}{37.14}                                                                         & \textcolor{red}{39.53}                            & \textcolor{red}{37.70}                            & \textcolor{red}{41.87}                             & 48.37                             \\
                                  & Offense         & 41.06                           & \textcolor{red}{38.59}                                                                         & \textcolor{red}{24.99}                            & \textcolor{red}{39.90}                            & 39.85                             & 48.70                             & 43.94                           & 40.25                                                                         & \textcolor{red}{21.99}                            & 41.53                            & 46.36                             & \textbf{54.49}                    & 42.42                           & 40.59                                                                         & \textcolor{red}{34.10}                            & 41.67                            & 43.83                             & 51.72                             \\
                                  & H/O-Group       & \textcolor{red}{13.63}                           & \textbf{21.23}                                                                & \textcolor{red}{7.02}                             & 16.25                            & \textcolor{red}{12.35}                             & \textcolor{red}{9.26}                              & \textcolor{red}{11.43}                           & \textcolor{red}{13.04}                                                                         & \textcolor{red}{7.92}                             & 15.98                            & \textcolor{red}{11.81}                             & \textcolor{red}{14.19}                             & \textcolor{red}{9.68}                            & \textcolor{red}{11.76}                                                                         & \textcolor{red}{7.23}                             & \textcolor{red}{14.50}                            & \textcolor{red}{12.58}                             & 16.27                             \\
                                  & H/O-Target      & \textcolor{red}{18.73}                           & \textcolor{red}{16.89}                                                                         & \textcolor{red}{6.69}                             & 20.58                            & \textcolor{red}{19.32}                             & \textcolor{red}{17.01}                             & \textcolor{red}{17.09}                           & \textcolor{red}{18.41}                                                                         & \textcolor{red}{10.48}                            & \textcolor{red}{16.55}                            & 20.56                             & \textbf{24.84}                    & \textcolor{red}{17.44}                           & \textcolor{red}{17.45}                                                                         & \textcolor{red}{9.32}                             & \textcolor{red}{16.56}                            & 20.09                             & 23.60                             \\\cdashline{2-20}
                                  & \textbf{AS}              & \textcolor{red}{33.70}                           & \textcolor{red}{31.97}                                                                         & \textcolor{red}{20.14}                            & \textcolor{red}{32.02}                            & \textcolor{red}{31.68}                             & \textcolor{red}{34.50}                             & \textcolor{red}{33.14}                           & \textcolor{red}{32.55}                                                                         & \textcolor{red}{27.19}                            & \textcolor{red}{32.36}                            & 36.42                             & \textbf{41.69}                    & \textcolor{red}{32.43}                           & \textcolor{red}{31.88}                                                                         & \textcolor{red}{29.17}                            & \textcolor{red}{32.09}                            & 35.35                             & 40.99                             \\ \hline
\multicolumn{2}{l}{\textbf{Emotion}}                         & \textcolor{red}{9.71}                            & \textcolor{red}{15.07}                                                                         & \textcolor{red}{7.75}                             & 27.87                            & \textcolor{red}{15.14}                             & 18.12                             & 17.08                           & \textcolor{red}{12.17}                                                                         & \textcolor{red}{10.66}                            & 23.12                            & 32.12                             & 40.28                             & 18.48                           & \textcolor{red}{12.35}                                                                         & \textcolor{red}{10.07}                            & 25.57                            & 34.20                             & \textbf{41.79}                    \\\hline
\multicolumn{2}{l}{\textbf{Humor}}                           & \textcolor{red}{41.78}                           & \textcolor{red}{33.04}                                                                         & \textcolor{red}{43.60}                            & \textcolor{red}{33.12}                            & \textcolor{red}{39.78}                             & \textcolor{red}{46.19}                             & \textcolor{red}{33.67}                           & \textcolor{red}{33.12}                                                                         & \textcolor{red}{44.70}                            & \textcolor{red}{38.19}                            & 55.20                             & 57.15                             & \textcolor{red}{34.06}                           & \textcolor{red}{33.12}                                                                         & \textcolor{red}{40.20}                            & \textcolor{red}{37.08}                            & 53.86                             & \textbf{58.75}                    \\\hline
\multirow{4}{*}{\rotatebox[origin=c]{90}{\textbf{I\&S}}}            & Irony           & \textcolor{red}{36.63}                           & 44.46                                                                         & \textcolor{red}{36.52}                            & \textcolor{red}{34.69}                            & \textcolor{red}{40.78}                             & \textbf{47.48}                    & \textcolor{red}{42.21}                           & \textcolor{red}{41.58}                                                                         & 42.61                            & \textcolor{red}{35.18}                            & \textcolor{red}{36.76}                             & \textcolor{red}{39.78}                             & 44.34                           & 44.14                                                                         & \textcolor{red}{39.67}                            & \textcolor{red}{34.82}                            & \textcolor{red}{38.40}                             & \textcolor{red}{41.61}                             \\
                                  & Sarcasm         & \textcolor{red}{43.00}                           & \textcolor{red}{41.68}                                                                         & \textcolor{red}{36.09}                            & \textcolor{red}{41.62}                            & \textcolor{red}{41.17}                             & 47.67                             & 46.14                           & \textcolor{red}{42.91}                                                                         & 46.72                            & 48.42                            & 49.75                             & \textbf{52.55}                    & \textcolor{red}{45.43}                           & 43.05                                                                         & 45.75                            & \textcolor{red}{39.88}                            & 49.03                             & 52.51                             \\
                                  & Irony-Type      & \textcolor{red}{18.83}                           & \textcolor{red}{18.83}                                                                         & \textcolor{red}{18.83}                            & \textcolor{red}{18.83}                            & \textcolor{red}{18.83}                             & \textcolor{red}{18.83}                             & \textcolor{red}{18.83}                           & \textcolor{red}{18.83}                                                                         & \textcolor{red}{18.83}                            & \textcolor{red}{18.83}                            & \textcolor{red}{18.83}                             & \textcolor{red}{18.83}                             & \textcolor{red}{18.83}                           & \textcolor{red}{18.83}                                                                         & \textcolor{red}{18.83}                            & \textcolor{red}{18.83}                            & \textcolor{red}{18.83}                             & \textbf{18.83}                    \\\cdashline{2-20}
                                  & \textbf{I\&S}             & \textcolor{red}{38.92}                           & \textcolor{red}{41.79}                                                                         & \textcolor{red}{35.42}                            & \textcolor{red}{37.36}                            & \textcolor{red}{39.87}                             & \textbf{46.14}                    & 43.01                           & \textcolor{red}{41.11}                                                                         & 43.48                            & \textcolor{red}{40.98}                            & \textcolor{red}{42.36}                             & 45.12                             & 43.61                           & \textcolor{red}{42.33}                                                                         & \textcolor{red}{41.67}                            & \textcolor{red}{36.55}                            & \textcolor{red}{42.74}                             & 45.92                             \\\hline
\multicolumn{2}{l}{\textbf{Sentiment}}                       & \textcolor{red}{26.67}                           & \textbf{43.03}                                                                & \textcolor{red}{20.77}                            & \textcolor{red}{34.65}                            & \textcolor{red}{27.55}                             & \textcolor{red}{25.02}                             & 34.81                           & 35.79                                                                         & \textcolor{red}{24.76}                            & \textcolor{red}{31.37}                            & 37.73                             & \textcolor{red}{34.53}                             & \textcolor{red}{33.15}                           & 37.71                                                                         & \textcolor{red}{23.17}                            & \textcolor{red}{29.25}                            & 40.88                             & 39.37                             \\\hline
\multicolumn{2}{l}{\textbf{Subjectivity}}                    & 44.12                           & \textcolor{red}{30.73}                                                                         & \textcolor{red}{37.35}                            & 41.64                            & \textcolor{red}{42.30}                             & \textcolor{red}{38.73}                             & \textcolor{red}{36.50}                           & \textcolor{red}{30.66}                                                                         & \textcolor{red}{37.11}                            & \textcolor{red}{34.36}                            & 44.20                             & 54.77                             & \textcolor{red}{33.70}                           & \textcolor{red}{30.72}                                                                         & \textcolor{red}{39.36}                            & \textcolor{red}{31.42}                            & 46.53                             & \textbf{56.15}                    \\\Xhline{3\arrayrulewidth}
\multicolumn{2}{l}{\textbf{SPARROW}}                         & \textcolor{red}{27.94}                           & 35.70                                                                         & \textcolor{red}{21.45}                            & 33.63                            & \textcolor{red}{28.75}                             & \textcolor{red}{29.36}                             & \textcolor{red}{32.76}                           & \textcolor{red}{31.91}                                                                         & \textcolor{red}{26.12}                            & \textcolor{red}{31.71}                            & 37.82                             & 39.44                             & \textcolor{red}{32.03}                           & \textcolor{red}{32.84}                                                                         & \textcolor{red}{25.47}                            & \textcolor{red}{30.44}                            & 39.48                             & \textbf{41.97}                    \\ \bottomrule
\end{tabular}
\caption{Evaluating open-source LLMs on SPARROW with few-shot in-context learning. The best performance in each setting is \textbf{Bold}. The \textcolor{red}{red font} denotes a performance lower than the random baseline. }\label{tab:few-shot}
\end{table*}

%% file: conclusion.tex
\section{Conclusion}
In order to understand the abilities of ChatGPT and other instruction tuned LLMs on capturing sociopragmatic meaning, we introduced a massively multilingual evaluation benchmark, dubbed SPARROW. The benchmark involves $169$ datasets covering $64$ languages from $12$ language families and $16$ scripts. Evaluating ChatGPT on SPARROW, we find it struggles with different languages. We also reveal that task-specific models finetuned on SM (much smaller than ChatGPT) consistently outperform larger models by a significant margin even on English.

% We also developed strong baseline models on SPARROW, as a way to facilitate future meaningful comparisons. %Along with SPARROW benchmark, we evaluated seven SoTA M-PLMs. % and introduced a model that intermediately fine-tuned XLM-R on emoji-based SLP for SM understanding. 

%% file: limitation_ethic.tex
\section{Limitations}
%As a initial work for evaluation multilingual SM understanding, o
\paragraph{\textbf{Benchmark Construction.}} Our SPARROW benchmark only includes text classification tasks related to SM. Despite our best efforts, we acknowledge that our benchmark has not covered existing SM datasets exhaustively. We will continue expanding this benchmark and welcome future datasets or metric contributions to it. We also plan to extend~\sparrow~to more types of tasks related to SM, such as span-based sentiment analysis~\cite{xu-etal-2020-aspect}, affective language generation~\cite{goswamy-etal-2020-adapting}, and conversational sentiment analysis~\cite{ojamaa-2015-sentiment}. We only include text-based SM tasks. Another improvement direction is to extend this benchmark to more tasks that involve more modalities, such affective image captioning~\cite{mohamed-2022-okay} and multi-modal emotion recognition~\cite{firdaus-etal-2020-meisd}. 

\paragraph{\textbf{Model Selection.}} Due to computation constraints, we cannot evaluate on model sizes $>7$B. However, we hope~\sparrow~will be used in the future to evaluate larger-sized models. Again, due to budget constraints, we only conduct a relatively small case study on GPT-4 and do not evaluate more diverse commercial instruction tuned models that are more expensive (e.g., \texttt{text-davinci-003} by OpenAI). %We leave this as future work.

\paragraph{\textbf{Experiments.}}
While we customize prompts employed for each task, we do not tailor prompts specifically for each model. We acknowledge that the performance of models may be influenced by different prompt variants. In future work, we will test diverse prompt variations for more robust results. We only experiment with machine translated prompts in our analyses and acknowledge that the performance drop may stem from the poor quality of machine translation. We will investigate the utility of human translated prompts in a future study. In this paper, we only evaluate LLMs on zero-shot learning.  The adoption of few-shot in-context learning may enhance performance, which we also leave to future work.

\section*{Ethics Statement and Broad Impacts}

% \hl{This section can have some positive things about our work, as these can be part of the wider impact. For example, we mitigate data decay, we also allow meaningful comparisons through a scoring board across different languages and tasks, and so on. I encourage you to enhance this section and also the previous one. }

\paragraph{\textbf{Data Collection and Releasing.}} All the $169$ datasets are produced by previous research. Since there are large numbers of datasets and languages in SPARROW, it is hard to manually verify the quality of all the datasets. As a quality assurance measure, we only include in SPARROW datasets that are introduced in peer-reviewed published research. To facilitate access to information about each dataset, we link to each published paper describing each of these datasets inTables~\ref{tab:antisocial_data},~\ref{tab:emotion_data},~\ref{tab:humor_data},~\ref{tab:irony_data},~\ref{tab:sentiment_data}, and~\ref{tab:subj_data}. 

Following privacy protection policies, we anonymize all~\sparrow~data as described in Section~\ref{subsec:preprocess}. With reference to accessibility of the original individual dataset, SPARROW data can be categorized into three releasing strategies: \textbf{(1)} In the case of datasets requiring approval by the original authors, we require future researchers to obtain approval first and will share our splits once approval has been obtained. We indicate these nine datasets in our data description tables. \textbf{(2)} For the $25$ datasets (see Table~\ref{tab:data_decay} in Appendix) that are shared via tweet IDs, we share our obtained data for research use. By doing so, we expect to mitigate the issue of data decay and allow fair comparisons. \textbf{(3)} We will share the other $135$ publicly accessible datasets upon request. We will also require a justification for responsible use of the datasets. %, and will share them for research purposes only. 
Each dataset will be shared in our Train, Dev, and Test splits along with a dataset card to indicate the original publication of the dataset. 

\paragraph{\textbf{Intended Use.}} The intended use of SPARROW benchmark is to construct a scoring board to facilitate model comparisons as well as enhance fairness and reproducibility across different languages and tasks. We also aim to mitigate data decay issues in social media research. SPARROW could help researchers investigate model's capacity on SM tasks across languages. SPARROW may also be used to investigate model transferability across a wide range of tasks and diverse languages in different settings (such as zero- or few-shot settings and prompting). 

\paragraph{\textbf{Potential Misuse and Bias.}}
We notice that some annotations in the datasets of SPARROW (e.g., for hate speech task~\cite{waseem-2016-hateful}) can carry annotation and temporal biases. We recommend that any dataset in SPARROW not be used for research or in applications without careful consideration of internal biases of the datasets and potential biases of the resulting systems. 
We also suggest that users of SPARROW not only focus on the overall SPARROW score but also a model performance on each task and dataset. The SPARROW score is an unweighted average score over all the dataset-specific metrics, which may lose the fine-grained information and be dominated by the largest task cluster (i.e., sentiment analysis) or languages (e.g., languages from Indo-European language family).

\section*{Acknowledgements}\label{sec:acknow}
We acknowledge support from Canada Research Chairs (CRC), the Natural Sciences and Engineering Research Council of Canada (NSERC; RGPIN-2018-04267), the Social Sciences and Humanities Research Council of Canada (SSHRC; 435-2018-0576; 895-2020-1004; 895-2021-1008), Canadian Foundation for Innovation (CFI; 37771), Digital Research Alliance of Canada,\footnote{\href{https://alliancecan.ca}{https://alliancecan.ca}} and UBC ARC-Sockeye.\footnote{\href{https://arc.ubc.ca/ubc-arc-sockeye}{https://arc.ubc.ca/ubc-arc-sockeye}}
% \section*{Acknowledgements}\label{sec:acknow}
% We gratefully acknowledge support from the Natural Sciences and Engineering Research Council of Canada (NSERC; RGPIN-2018-04267), the Social Sciences and Humanities Research Council of Canada (SSHRC; 435-2018-0576; 895-2020-1004; 895-2021-1008), Canadian Foundation for Innovation (CFI; 37771), Digital Research Alliance of Canada (the Alliance),\footnote{\href{https://ccdb.alliancecan.ca}{https://www.computecanada.ca}} and UBC ARC-Sockeye.\footnote{\href{https://arc.ubc.ca/ubc-arc-sockeye}{https://arc.ubc.ca/ubc-arc-sockeye}} Any opinions, conclusions or recommendations expressed in this material are those of the author(s) and do not necessarily reflect the views of NSERC, SSHRC, CFI, the Alliance, or UBC ARC-Sockeye.

%% file: appendix.tex
\textbf{\Large Appendices}

\input{table_fig/language_sum}
 
\section{Benchmark}
Table~\ref{tab:lang_sum} summarizes language distribution of datasets in SPARROW and taxonomy of these language according to Ethnologue~\cite{gordon2005ethnologue} and Glottolog~\cite{hammarstrom2021glottolog}. Tables~\ref{tab:antisocial_data},~\ref{tab:emotion_data},~\ref{tab:humor_data},~\ref{tab:irony_data},~\ref{tab:sentiment_data}, and~\ref{tab:subj_data} describe the datasets in tasks of antisocial language detection, emotion recognition, humor detection, irony and sarcasm detection, sentiment analysis, and subjectivity analysis, respectively.  
\input{table_fig/data_decay}

We empirically characterize the issue of data inaccessibility by re-collecting tweets content via tweet IDs. Table~\ref{tab:data_decay} shows the data decay issue of $25$ datasets.

\input{table_fig/antisocial}
\input{table_fig/emotion}
\input{table_fig/humor}
\input{table_fig/irony}
\input{table_fig/sentiment}
\input{table_fig/subject}

% \input{table_fig/data_decay}

% \section{Data for SLP Training}\label{sec:data_slp}
% To obtain multilingual surrogate labeled training data, we collect $2$B tweets produced from 2014 to 2021. Tweets are normalized by converting user mentions and hyperlinks to `USER' and `URL', respectively. We select tweet that contains only a unique type of emoji (regardless of the number of emojis) and also has more than five actual words (without counting the special tokens such as hashtag, emoji, user mention, and hyperlink). We then extract the emoji as a label of the tweet and remove the emoji from the tweet. We exclude emojis occurring less than $200$ times. As a result, we acquire a set of $1,067$ emojis in $594$M tweets from $66$ languages. Due to the imbalance between languages, we set the maximal size of each language as $10$M and random sample $10$M tweets for languages that excess this upper bound, which gives us $111$M tweets. 
\input{append_models}

\section{Experiments}
\subsection{Hyperparameters}\label{sec:hyper}
To be computation friendly, we only tune the peak learning rate of each model in a set of $\{1e-4, 5e-5, 3e-5, 1e-5\}$ and randomly select $45$ datasets for hyper-parameter tuning. We fine-tune a PLM with an arbitrary batch size of $32$, sequence length of $128$ tokens, and $20$ epochs with patience of five epochs based on the model performance on Dev set. We fine-tune each dataset three time with different seeds and identify the best model based on Dev set performance. The best learning rate for each model is identified based on the average score of Dev set of the $45$ datasets. The best peak learning rate is $3e-5$ for mBERT, XLM-T, and Bernice and $1e-5$ for other models.

\subsection{Prompts}
The prompts we use in our experiments are summarized in Table~\ref{tab:prompt_new}.

\input{table_fig/new_prompt}

\subsection{Results}\label{sec:append_results}
Table~\ref{tab:finetune_more_results} shows aggregated performance of finetuned models on Dev and Test-S. We report the average of dataset-specific metrics and standard deviation in a task and a category. 
We also report the Test-S performance of tasks of antisocial language detection, emotion recognition, humor detection, irony and sarcasm detection, sentiment analysis, and subjectivity analysis in Tables~\ref{tab:anti-social_res}, \ref{tab:emotion_res}, \ref{tab:humor_res}, \ref{tab:irony_res}, \ref{tab:sentiment_res}, and \ref{tab:subjectivity_res}, respectively. 

We provide a concise study to probe the sensitivity of open-source LLMs to prompts and present the results in Table~\ref{tab:prompt_sensitive}.

\input{table_fig/finetune_more_results}
\input{table_fig/antisocial_res}
\input{table_fig/emotion_res}
\input{table_fig/humor_res}
\input{table_fig/irony_res}
\input{table_fig/sentiment_res}
\input{table_fig/subject_res}

\input{table_fig/language_wise_results}
\input{table_fig/prompt_sensitive}

%% file: table_fig/language_sum.tex
% \usepackage{booktabs}
% \usepackage{multirow}
% \usepackage[normalem]{ulem}
\begin{table}[ht]
\tiny
\centering
\begin{tabular}{@{}llccr@{}}
\toprule
\multicolumn{1}{c}{\textbf{Lang. Family}} & \multicolumn{1}{c}{\textbf{Lang.}} & \multicolumn{1}{c}{\textbf{Code}} & \textbf{\# dataset} & \multicolumn{1}{c}{\textbf{Script}} \\ \midrule
\multirow{7}{*}{Afro-Asiatic}            & Amharic                           & amh                               & 1                   & Ethiopic                            \\
                                & Arabic                            & ara                               & 15                  & Arabic                              \\
                                         
                                         & Darija                            & ary                               & 1                   & Arabic                              \\
                                         & Dziria                            & arq                               & 1                   & Arabic                              \\
                                         & Hausa                             & hau                               & 1                   & Latin                               \\
                                         & Hebrew                            & heb                               & 1                   & Hebrew                              \\
                                         & Maltese                           & mlt                               & 1                   & Latin                               \\\midrule
\multirow{7}{*}{Atlantic-Congo}          & Bambara                           & bam                               & 1                   & Latin                               \\
                                         & Igbo                              & ibo                               & 1                   & Latin                               \\
                                         & Kinyarwanda                       & kin                               & 1                   & Latin                               \\
                                         & Swahili                           & swh                               & 1                   & Latin                               \\
                                         & Twi                               & twi                               & 1                   & Latin                               \\
                                         & Tsonga                            & tso                               & 1                   & Latin                               \\
                                         & Yoruba                            & yor                               & 2                   & Latin                               \\\midrule
Austroasiatic                            & Vietnamese                        & vie                               & 1                   & Latin                               \\\midrule
\multirow{13}{*}{Austronesian}           & Acehnese                          & ace                               & 1                   & Latin                               \\
                                         & Balinese                          & ban                               & 1                   & Latin                               \\
                                         & Banjarese                         & bjn                               & 1                   & Latin                               \\
                                         & Buginese                          & bug                               & 1                   & Latin                               \\
                                         & Filipino                          & fil                               & 1                   & Latin                               \\
                                         & Indonesian                        & ind                               & 3                   & Latin                               \\
                                         & Javanese                          & jav                               & 1                   & Latin                               \\
                                         & Madurese                          & mad                               & 1                   & Latin                               \\
                                         % & Malay                           & msa                               & 1                   & Latin                          \\
                                         & Minangkabau                       & min                               & 1                   & Latin                               \\
                                         & Ngaju                             & nij                               & 1                   & Latin                               \\
                                         & Sundanese                         & sun                               & 1                   & Latin                               \\
                                         & Toba batak                        & bbc                               & 1                   & Latin                               \\\midrule
\multirow{4}{*}{Dravidian}                                & Kannada                           & kan                               & 2                   & Kannada, Latin                      \\
                                         & Malayalam                         & mal                               & 2                   & Malayalam, Latin                    \\
                                         & Tamil                             & tam                               & 2                   & Tamil, Latin                        \\
                                         & Telugu                            & tel                               & 2                   & Telugu                              \\\midrule
\multirow{26}{*}{Indo-European}          & Albanian                          & sqi                               & 1                   & Latin                               \\
                                         & Bosnian                           & bos                               & 1                   & Latin                               \\
                                         & Bulgarian                         & bul                               & 1                   & Cyrillic                            \\
                                         & Bengali                           & ben                               & 4                   & Bengali, Latin                      \\
                                         & Croatian                          & hrv                               & 1                   & Latin                               \\
                                         & Czech                             & ces                               & 2                   & Latin                               \\
                                         & Danish                            & dan                               & 1                   & Latin                               \\
                                         & English                           & eng                               & 27                  & Latin                               \\
                                         & French                            & fra                               & 6                   & Latin                               \\
                                         & German                            & deu                               & 3                   & Latin                               \\
                                         & Greek                             & ell                               & 1                   & Greek                               \\
                                         & Hindi                             & hin                               & 5                   & Devanagari, Latin                               \\
                                         & Italian                           & ita                               & 11                   & Latin                               \\
                                         & Marathi                           & mar                               & 1                   & Devanagari                          \\
                                         & Nigerian Pidgin                   & pcm                               & 2                   & Latin                               \\
                                         & Norwegian                         & nor                               & 1                   & Latin                               \\
                                         & Persian                           & fas                               & 3                   & Arabic                              \\
                                         & Portuguese                        & por                               & 4                   & Latin                               \\
                                         & Polish                            & pol                               & 4                   & Latin                               \\
                                         & Romanian                          & ron                               & 2                   & Latin                               \\
                                         & Russian                           & rus                               & 3                   & Cyrillic                            \\
                                         & Spanish                           & spa                               & 9                   & Latin                               \\
                                         & Serbian                           & srp                               & 1                   & Cyrillic                            \\
                                         & Slovak                            & slk                               & 1                   & Latin                               \\
                                         & Slovenian                         & slv                               & 2                   & Latin                               \\
                                         & Swedish                           & swe                               & 1                   & Latin                               \\\midrule
Japonic                                  & Japanese                          & jpn                               & 1                   & Han, Hir., Kat.                     \\\midrule
Koreanic                                 & Korean                            & kor                               & 5                   & Hangul                              \\\midrule
Sino-Tibetan                             & Chinese                           & zho                               & 6                   & Han                                 \\\midrule 
Tai-Kadai                                & Thai                              & tha                               & 1                   & Thai                                \\ \midrule
Turkic                                   & Turkish                           & tur                               & 2                   & Latin                               \\ \midrule
\multirow{2}{*}{Uralic}                  & Finnish                           & fin                               & 3                   & Latin                               \\ 
                                         & Hungarian                         & hun                               & 1                   & Latin                               \\ \bottomrule
\end{tabular}
\caption{Summary of languages covered in SPARROW. \textbf{Lang.:} Language. Language code is marked by ISO 639-3 code. Language information is retrieved from Ethnologue~\cite{gordon2005ethnologue} and Glottolog~\cite{hammarstrom2021glottolog}. The column \textbf{\# dataset} shows the number of datasets covered by SPARROW per language. \textbf{Hir.:} Hiragana, \textbf{Kat.:} Katakana} 
\label{tab:lang_sum}

\end{table}

%https://glottolog.org/glottolog/language

%% file: table_fig/data_decay.tex
\begin{table}[h]
\setlength\tabcolsep{3pt}
\centering
\tiny
\begin{tabular}{@{}llcrrc@{}}
\toprule
\multicolumn{1}{c}{\textbf{Dataset}}         & \multicolumn{1}{c}{\textbf{Study}}                & \textbf{Year} & \textbf{Original} & \textbf{Retrieval} & \textbf{Decay \%} \\ \midrule
Sarc-eng\textsubscript{Ril} & \citet{riloff2013sarcasm}        & 2013          & 3K             & 1K             & 0.41                \\
Sarc-ces\textsubscript{Pta} & \citet{ptavcek2014sarcasm}       & 2013          & 7K             & 4K             & 0.29                \\
Sarc-eng\textsubscript{Pta} & \citet{ptavcek2014sarcasm}       & 2013          & 100K           & 89K            & 0.11                \\
Sarc-eng\textsubscript{Bam} & \citet{bamman2015contextualized} & 2015          & 19K            & 14K            & 0.24                \\
Sent-bul\textsubscript{Moz} & \citet{mozetivc2016multilingual} & 2016          & 67K            & 27K            & 0.59                \\
Sent-bos\textsubscript{Moz} & \citet{mozetivc2016multilingual} & 2016          & 44K            & 20K            & 0.54                \\
Sent-deu\textsubscript{Moz} & \citet{mozetivc2016multilingual} & 2016          & 109K           & 52K            & 0.52                \\
Sent-eng\textsubscript{Moz} & \citet{mozetivc2016multilingual} & 2016          & 103K           & 43K            & 0.58                \\
Sent-spa\textsubscript{Moz} & \citet{mozetivc2016multilingual} & 2016          & 275K           & 153K           & 0.44                \\
Sent-hrv\textsubscript{Moz} & \citet{mozetivc2016multilingual} & 2016          & 97K            & 66K            & 0.32                \\
Sent-hun\textsubscript{Moz} & \citet{mozetivc2016multilingual} & 2016          & 109K           & 40K            & 0.63                \\
Sent-pol\textsubscript{Moz} & \citet{mozetivc2016multilingual} & 2016          & 223K           & 109K           & 0.51                \\
Sent-por\textsubscript{Moz} & \citet{mozetivc2016multilingual} & 2016          & 157K           & 49K            & 0.69                \\
Sent-rus\textsubscript{Moz} & \citet{mozetivc2016multilingual} & 2016          & 107K           & 41K            & 0.62                \\
Sent-slk\textsubscript{Moz} & \citet{mozetivc2016multilingual} & 2016          & 70K            & 38K            & 0.46                \\
Sent-slv\textsubscript{Moz} & \citet{mozetivc2016multilingual} & 2016          & 133K           & 74K            & 0.44                \\
Sent-sqi\textsubscript{Moz} & \citet{mozetivc2016multilingual} & 2016          & 53K            & 36K            & 0.31                \\
Sent-srp\textsubscript{Moz} & \citet{mozetivc2016multilingual} & 2016          & 73K            & 27K            & 0.63                \\
Sent-swe\textsubscript{Moz} & \citet{mozetivc2016multilingual} & 2016          & 58K            & 34K            & 0.42                \\
Hate-eng\textsubscript{Was} & \citet{waseem-2016-hateful}      & 2016          & 16K            & 10K            & 0.36                \\
Sent-por\textsubscript{Bru} & \citet{brum2018expansao}          & 2017          & 157K            & 56K            & 0.64                \\ 
Sent-eng\textsubscript{Ros} & \citet{rosenthal-2017-semeval}   & 2017          & 50K            & 42K            & 0.15                \\
Iron-hin\textsubscript{Vij} & \citet{vijay-2018-dataset-irony} & 2018          & 3K              & 2K              & 0.10                \\
Sexi-fre\textsubscript{Chi} & \citet{chiril2020annotated}      & 2018          & 12K             & 9K              & 0.22                \\
Humo-hin\textsubscript{Agg} & \citet{aggarwal2020sarcasm}      & 2018          & 7K              & 5K              & 0.30                \\
\bottomrule
\end{tabular}
\caption{Data decay issue in social media data. These $25$ datasets are distribute by tweet IDs. We retrieve these tweet on Nov. 2020 - Jan. 2022 and find that $42\%$ samples are inaccessible. } \label{tab:data_decay}
\end{table}

%% file: table_fig/antisocial.tex
\begin{table*}[h]
\centering 
\tiny
\begin{tabular}{@{}llcp{0.1\linewidth}lcp{0.2\linewidth}rc@{}}
\toprule
\textbf{Dataset}	 & \textbf{Study}	 & \textbf{Lang.}	 & \textbf{Source / Domain}	 & \textbf{Year}	 & \textbf{\#Lb}	 & \textbf{Labels}	 & \textbf{Data Slipt}	 & \textbf{Metric}\\ \midrule
Aggr-hin\textsubscript{Kum} 	 & \citet{kumar-2018-aggression} 	 & hin 	 & Twitter, Facebook 	 & 2018 	 & 2 	 & \{Aggressive, Not\} 	 & 9,306/1,163/1,164 	 & M-F1	\\ \hline
Dang-ara\textsubscript{Als} 	 & \citet{alshehri-etal-2020-understanding} 	 & ara 	 & Twitter 	 & 2020 	 & 2 	 & \{Dangerous, Not\} 	 & 3,474/615/663 	 & M-F1	\\ \hline
Hate-eng\textsubscript{Was} 	 & \citet{waseem-2016-hateful} 	 & eng 	 & Twitter 	 & 2016 	 & 3 	 & \{Not, Racism, Sexism\} 	 & 8,683/1,086/1,085 	 & W-F1	\\ 
Hate-eng\textsubscript{Dav} 	 & \citet{davidson-2017-hateoffensive} 	 & eng 	 & Twitter 	 & 2017 	 & 3 	 & \{Hate, Not, Offensive\} 	 & 19,826/2,478/2,479 	 & W-F1	\\ 
Hate-ara\textsubscript{Ala} 	 & \citet{alakrot2018dataset} 	 & ara 	 & YouTube comment 	 & 2017 	 & 2 	 & \{Hate, Not\} 	 & 9,014/1,127/1,127 	 & M-F1	\\ 
Hate-ita\textsubscript{Bos}\textsuperscript{$\star$} 	 & \citet{bosco2018overview} 	 & ita 	 & Twitter 	 & 2018 	 & 2 	 & \{Hate, Not\} 	 & 2,700/300/1,000 	 & M-F1	\\ 
Hate-fil\textsubscript{Cab} 	 & \citet{cabasag2019hate} 	 & fil 	 & Twitter 	 & 2016 	 & 2 	 & \{Hate, Not\} 	 & 10,000/4,232/4,232 	 & M-F1	\\ 
Hate-ara\textsubscript{Mul} 	 & \citet{mulki2019hsab} 	 & ara 	 & Twitter 	 & 2019 	 & 3 	 & \{Abusive, Hate, Not\} 	 & 4,208/468/1,170 	 & M-F1	\\ 
Hate-eng\textsubscript{Bas} 	 & \citet{basile-2019-semeval} 	 & eng 	 & Twitter 	 & 2019 	 & 2 	 & \{Hate, Not\} 	 & 9,000/1,000/3,000 	 & M-F1	\\ 
Hate-spa\textsubscript{Bas} 	 & \citet{basile-2019-semeval} 	 & spa 	 & Twitter 	 & 2019 	 & 2 	 & \{Hate, Not\} 	 & 4,500/500/1,600 	 & M-F1	\\ 
Hate-por\textsubscript{For} 	 & \citet{fortuna-2019-hierarchically} 	 & por 	 & Twitter 	 & 2019 	 & 2 	 & \{Hate, Not\} 	 & 4,536/567/567 	 & M-F1	\\ 
Hate-pol\textsubscript{Pta} 	 & \citet{ptaszynski2019results} 	 & pol 	 & Twitter 	 & 2019 	 & 2 	 & \{Hate, Not\} 	 & 9,037/1,004/1,000 	 & M-F1	\\ 
Hate-kor\textsubscript{Moo} 	 & \citet{moon-etal-2020-beep} 	 & kor 	 & News comment 	 & 2020 	 & 3 	 & \{Hate, Not, Offensive\} 	 & 7,106/790/471 	 & M-F1	\\ 
Hate-ara\textsubscript{Mub} 	 & \citet{mubarak-etal-2020-overview} 	 & ara 	 & Twitter 	 & 2020 	 & 2 	 & \{Hate, Not\} 	 & 6,839/1,000/2,000 	 & M-F1	\\ 
Hate-zho\textsubscript{Den} 	 & \citet{deng-2022-cold} 	 & zho 	 & Weibo 	 & 2022 	 & 2 	 & \{Hate, Not\} 	 & 25,726/6,431/5,323 	 & M-F1	\\ 
Hate-kor\textsubscript{Jeo} 	 & \citet{jeong-2022-kold} 	 & kor 	 & News and YouTube comment 	 & 2022 	 & 2 	 & \{Hate, Not\} 	 & 32,343/4,043/4,043 	 & M-F1	\\ 
Hate-tel\textsubscript{Mar} 	 & \citet{mounika-2022-resource} 	 & tel 	 & Misc 	 & 2022 	 & 2 	 & \{Hate, Not\} 	 & 24,599/3,510/7,033 	 & M-F1	\\ 
Sexi-fra\textsubscript{Chi} 	 & \citet{chiril2020annotated} 	 & fra 	 & Twitter 	 & 2018 	 & 2 	 & \{Not, Sexism\} 	 & 7,670/959/959 	 & M-F1	\\  \midrule
Offe-eng\textsubscript{Zam} 	 & \citet{zampieri-2019-predicting} 	 & eng 	 & Twitter 	 & 2019 	 & 2 	 & \{Not, Offensive\} 	 & 11,916/1,324/860 	 & M-F1	\\ 
Offe-ara\textsubscript{Zam} 	 & \citet{zampieri-etal-2020-semeval} 	 & ara 	 & Twitter 	 & 2019 	 & 2 	 & \{Not, Offensive\} 	 & 7,055/784/1,827 	 & M-F1	\\ 
Offe-dan\textsubscript{Zam} 	 & \citet{zampieri-etal-2020-semeval} 	 & dan 	 & Misc 	 & 2019 	 & 2 	 & \{Not, Offensive\} 	 & 2,664/296/329 	 & M-F1	\\ 
Offe-ell\textsubscript{Zam} 	 & \citet{zampieri-etal-2020-semeval} 	 & ell 	 & Twitter 	 & 2019 	 & 2 	 & \{Not, Offensive\} 	 & 7,869/874/1,544 	 & M-F1	\\ 
Offe-tur\textsubscript{Zam} 	 & \citet{zampieri-etal-2020-semeval} 	 & tur 	 & Twitter 	 & 2019 	 & 2 	 & \{Not, Offensive\} 	 & 28,149/3,128/3,515 	 & M-F1	\\ 
Offe-ara\textsubscript{Mub} 	 & \citet{mubarak-etal-2020-overview} 	 & ara 	 & Twitter 	 & 2020 	 & 2 	 & \{Not, Offensive\} 	 & 6,839/1,000/2,000 	 & M-F1	\\ 
Offe-slv\textsubscript{Nov} 	 & \citet{kralj2021slovenian} 	 & slv 	 & Twitter 	 & 2020 	 & 4 	 & \{Appropriate, Inappropriate, Not, Offensive\} 	 & 65,021/8,127/8,128 	 & M-F1	\\ \midrule
Offe-G-eng\textsubscript{Zam} 	 & \citet{zampieri-2019-predicting} 	 & eng 	 & Twitter 	 & 2019 	 & 3 	 & \{Group, Individual, Others\} 	 & 3,485/391/213 	 & M-F1	\\ 
Hate-G-ara\textsubscript{Ous} 	 & \citet{ousidhoum2019multilingual} 	 & ara 	 & Twitter 	 & 2019 	 & 13 	 & \{African\_descent, Arabs, Asians, Christian, Gay, Immigrants, Indian/hindu, Individual, Jews, Muslims, Others, Refugees, Women\} 	 & 2,682/334/335 	 & M-F1	\\ 
Hate-G-fra\textsubscript{Ous} 	 & \citet{ousidhoum2019multilingual} 	 & fra 	 & Twitter 	 & 2019 	 & 16 	 & \{African\_descent, Arabs, Asians, Christian, Gay, Gispanics, Immigrants, Indian/hindu, Individual, Jews, Left\_wing\_people, Muslims, Others, Refugees, Special\_needs, Women\} 	 & 3,211/401/402 	 & M-F1	\\  \midrule
Offe-T-eng\textsubscript{Zam} 	 & \citet{zampieri-2019-predicting} 	 & eng 	 & Twitter 	 & 2019 	 & 2 	 & \{Targeted, Untargeted\} 	 & 3,963/437/240 	 & M-F1	\\ 
Hate-T-ara\textsubscript{Ous} 	 & \citet{ousidhoum2019multilingual} 	 & ara 	 & Twitter 	 & 2019 	 & 4 	 & \{Gender, Origin, Others, Religion\} 	 & 2,682/334/336 	 & M-F1	\\ 
Hate-T-fra\textsubscript{Ous} 	 & \citet{ousidhoum2019multilingual} 	 & fra 	 & Twitter 	 & 2019 	 & 6 	 & \{Disability, Gender, Origin, Others, Religion, Sexual\_Orientation\} 	 & 3,211/401/402 	 & M-F1	\\ 
Hate-T-ben\textsubscript{Kar} 	 & \citet{karim2020deephateexplainer} 	 & ben 	 & Misc 	 & 2020 	 & 4 	 & \{Geopolitical, Personal, Political, Religion\} 	 & 4,558/570/570 	 & M-F1	\\ 
Offe-T-kan\textsubscript{Cha} 	 & \citet{chakravarthi-2022-dravidian} 	 & kan 	 & YouTube comment 	 & 2019 	 & 5 	 & \{Group, Individual, Not, Others, Untargeted\} 	 & 4,694/586/593 	 & M-F1	\\ 
Offe-T-mal\textsubscript{Cha} 	 & \citet{chakravarthi-2022-dravidian} 	 & mal 	 & YouTube comment 	 & 2019 	 & 4 	 & \{Group, Individual, Not, Untargeted\} 	 & 14,723/1,836/1,844 	 & M-F1	\\ 
Offe-T-tam\textsubscript{Cha} 	 & \citet{chakravarthi-2022-dravidian} 	 & tam 	 & YouTube comment 	 & 2019 	 & 5 	 & \{Group, Individual, Not, Others, Untargeted\} 	 & 33,685/4,216/4,232 	 & M-F1	\\ 
Hate-T-kor\textsubscript{Jeo} 	 & \citet{jeong-2022-kold} 	 & kor 	 & News and YouTube comment 	 & 2022 	 & 4 	 & \{Group, Individual, Other, Untargeted\} 	 & 16,239/2,049/2,022 	 & M-F1	\\ 
\bottomrule 
\end{tabular}
\caption{Description of $36$ antisocial language detection datasets. \textbf{Lang.:} Language is marked by ISO 639-3, \textbf{\#Lb:} the label size of a dataset. \textbf{M-F1:} Macro-F1, \textbf{W-F1:} Weighted-F1. \textsuperscript{$\star$} indicates that data sharing needs approval from the original authors. }\label{tab:antisocial_data}
\end{table*}

%% file: table_fig/emotion.tex
\begin{table*}[h]
\centering 
\tiny
\begin{tabular}{@{}llcp{0.1\linewidth}lcp{0.2\linewidth}rc@{}}
 \toprule
\textbf{Dataset}	 & \textbf{Study}	 & \textbf{Lang.}	 & \textbf{Source / Domain}	 & \textbf{Year}	 & \textbf{\#Lb}	 & \textbf{Labels}	 & \textbf{Data Slipt}	 & \textbf{Metric}\\ \midrule
Emot-eng\textsubscript{Wal} 	 & \citet{wallbott1986universal} 	 & eng 	 & Questionnaire 	 & 1986 	 & 7 	 & \{Anger, Disgust, Fear, Guilt, Joy, Sadness, Shame\} 	 & 6,132/767/767 	 & M-F1	\\ 
Emot-zho\textsubscript{Lee} 	 & \citet{wang-2015-emotion} 	 & zho 	 & Weibo 	 & 2015 	 & 5 	 & \{Anger, Fear, Happy, Sadness, Surprise\} 	 & 3,122/347/418 	 & Accuracy	\\ 
Emot-fin\textsubscript{Kaj} 	 & \citet{kajava2018cross} 	 & fin 	 & Subtitle 	 & 2016 	 & 8 	 & \{Anger, Anticipation, Disgust, Fear, Joy, Sadness, Surprise, Trust\} 	 & 5,197/577/653 	 & M-F1	\\ 
Emot-fra\textsubscript{Kaj} 	 & \citet{kajava2018cross} 	 & fra 	 & Subtitle 	 & 2016 	 & 8 	 & \{Anger, Anticipation, Disgust, Fear, Joy, Sadness, Surprise, Trust\} 	 & 5,198/577/653 	 & M-F1	\\ 
Emot-ita\textsubscript{Kaj} 	 & \citet{kajava2018cross} 	 & ita 	 & Subtitle 	 & 2016 	 & 8 	 & \{Anger, Anticipation, Disgust, Fear, Joy, Sadness, Surprise, Trust\} 	 & 5,197/577/653 	 & M-F1	\\ 
% Emot-msa\textsubscript{Hus} 	 & \citet{Malay-Dataset} 	 & msa 	 & Misc 	 & 2017 	 & 6 	 & \{Anger, Fear, Happy, Love, Sadness, Surprise\} 	 & 128,020/16002/16003 	 & M-F1	\\ 
Emot-ara\textsubscript{Abd} 	 & \citet{mageed-2020-aranet} 	 & ara 	 & Twitter 	 & 2016 	 & 8 	 & \{Anger, Anticipation, Disgust, Fear, Joy, Sadness, Surprise, Trust\} 	 & 50,000/910/941 	 & M-F1	\\ 
Emot-eng\textsubscript{Moh} 	 & \citet{mohammad-2018-semeval} 	 & eng 	 & Twitter 	 & 2018 	 & 4 	 & \{Anger, Joy, Optimism, Sadness\} 	 & 3,257/374/1,421 	 & M-F1	\\ 
Emot-ara\textsubscript{Moh} 	 & \citet{mohammad-2018-semeval} 	 & ara 	 & Twitter 	 & 2018 	 & 4 	 & \{Anger, Joy, Fear, Sadness\} 	 & 2,284/490/1,188 	 & M-F1	\\ 
Emot-spa\textsubscript{Moh} 	 & \citet{mohammad-2018-semeval} 	 & spa 	 & Twitter 	 & 2018 	 & 4 	 & \{Anger, Joy, Fear, Sadness\} 	 & 2,708/479/1,696 	 & M-F1	\\ 
Emot-ind\textsubscript{Sap} 	 & \citet{emotion-2018-saputri} 	 & ind 	 & Twitter 	 & 2019 	 & 5 	 & \{Anger, Fear, Happy, Love, Sadness\} 	 & 3,520/440/441 	 & M-F1	\\ 
Emot-tur\textsubscript{Guv} 	 & \citet{guven2020comparison} 	 & tur 	 & Twitter 	 & 2020 	 & 5 	 & \{Anger, Fear, Happy, Sadness, Surprise\} 	 & 3,200/400/400 	 & Accuracy	\\ 
% Emot-spa\textsubscript{Moh} 	 & \citet{mohammad-2018-semeval} 	 & spa 	 & Twitter 	 & 2018 	 & 4 	 & \{Anger, Fear, Joy, Sadness\} 	 & 4,541/793/2,616 	 & M-F1	\\ 
Emot-ind\textsubscript{Wil} 	 & \citet{wong-2020-indonlu} 	 & ind 	 & Twitter 	 & 2018 	 & 5 	 & \{Anger, Fear, Happy, Love, Sadness\} 	 & 3,169/352/440 	 & M-F1	\\ 
Emot-vie\textsubscript{Ho} 	 & \citet{ho2019emotion} 	 & vie 	 & Facebook 	 & 2019 	 & 7 	 & \{Anger, Disgust, Fear, Joy, Others, Sadness, Surprise\} 	 & 5,548/686/693 	 & W-F1	\\ 
Emot-eng\textsubscript{Pla} 	 & \citet{plaza-del-arco-etal-2020-emoevent} 	 & eng 	 & Twitter 	 & 2020 	 & 7 	 & \{Anger, Disgust, Fear, Joy, Others, Sadness, Surprise\} 	 & 5,842/730/731 	 & M-F1	\\ 
Emot-spa\textsubscript{Pla} 	 & \citet{plaza-del-arco-etal-2020-emoevent} 	 & spa 	 & Twitter 	 & 2020 	 & 7 	 & \{Anger, Disgust, Fear, Joy, Others, Sadness, Surprise\} 	 & 6,727/841/841 	 & M-F1	\\ 
Emot-fin\textsubscript{Ohm} 	 & \citet{ohman2020xed} 	 & fin 	 & Subtitle 	 & 2020 	 & 8 	 & \{Anger, Anticipation, Disgust, Fear, Joy, Sadness, Surprise, Trust\} 	 & 8,864/1,118/1,086 	 & M-F1	\\ 
Emot-eng\textsubscript{Dem} 	 & \citet{demszky2020goemotion} 	 & eng 	 & Reddit 	 & 2020 	 & 27 	 & \{Admiration, Amusement, Anger, Annoyance, Approval, Caring, Confusion, Curiosity, Desire, Disappointment, Disapproval, Disgust, Embarrassment, Excitement, Fear, Gratitude, Grief, Joy, Love, Nervousness, Optimism, Pride, Realization, Relief, Remorse, Sadness5, Surprise6\} 	 & 23,485/2,956/2,984 	 & M-F1	\\ 
Emot-ita\textsubscript{Bia} 	 & \citet{bianchi2021feel} 	 & ita 	 & Twitter 	 & 2021 	 & 4 	 & \{Anger, Fear, Joy, Sadness\} 	 & 1,629/204/204 	 & M-F1	\\ 
Emot-ron\textsubscript{Cio} 	 & \citet{red-2021-ciobotaru} 	 & ron 	 & Twitter 	 & 2020 	 & 4 	 & \{Anger, Fear, Joy, Sadness\} 	 & 2,600/318/324 	 & M-F1	\\ 
Emot-hin\textsubscript{Deb} 	 & \citet{emohind-2021-debaditya} 	 & hin 	 & Machine Translation 	 & 2021 	 & 27 	 & \{Admiration, Amusement, Anger, Annoyance, Approval, Caring, Confusion, Curiosity, Desire, Disappointment, Disapproval, Disgust, Embarrassment, Excitement, Fear, Gratitude, Grief, Joy, Love, Nervousness, Optimism, Pride, Realization, Relief, Remorse, Sadness, Surprise\} 	 & 23,485/2,956/2,984 	 & M-F1	\\ 
Emot-por\textsubscript{Cor} 	 & \citet{weak-2021-diogo} 	 & por 	 & Twitter 	 & 2021 	 & 28 	 & \{Admiration, Amusement, Anger, Annoyance, Approval, Compassion, Confusion, Curiosity, Desire, Disappointment, Disapproval, Disgust, Embarrassment, Envy, Excitement, Fear, Gratitude, Grief, Joy, Longing, Love, Nervousness, Optimism, Pride, Relief, Remorse, Sadness, Surprise\} 	 & 24,919/2,769/12,966 	 & M-F1	\\ 
Emot-fas\textsubscript{Sab} 	 & \citet{emopars-2021-sabri} 	 & fas 	 & Twitter 	 & 2021 	 & 6 	 & \{Anger, Fear, Happy, Hatred, Sadness, Wonder\} 	 & 4,180/523/523 	 & M-F1	\\ 
Emot-rus\textsubscript{Sbo} 	 & \citet{sboev2021data} 	 & rus 	 & Misc 	 & 2021 	 & 5 	 & \{Anger, Fear, Joy, Sadness, Surprise\} 	 & 3,951/427/1,128 	 & M-F1	\\ 
Emot-ben\textsubscript{Iqb} 	 & \citet{iqbal2022bemoc} 	 & ben 	 & Misc 	 & 2022 	 & 6 	 & \{Anger, Disgust, Fear, Joy, Sadness, Surprise\} 	 & 5,600/700/700 	 & M-F1	\\ 
Emot-fra\textsubscript{Bia}\textsuperscript{$\star$} 	 & \citet{bianchi-2022-xlm} 	 & fra 	 & Machine Translation 	 & 2018 	 & 4 	 & \{Anger, Fear, Joy, Sadness\} 	 & 3,798/476/476 	 & M-F1	\\ 
Emot-deu\textsubscript{Bia}\textsuperscript{$\star$} 	 & \citet{bianchi-2022-xlm} 	 & deu 	 & Machine Translation 	 & 2018 	 & 4 	 & \{Anger, Fear, Joy, Sadness\} 	 & 3,798/476/476 	 & M-F1	\\ 
\bottomrule 
\end{tabular}
\caption{Description of $26$ emotion recognition datasets. \textbf{Lang.:} Language is marked by ISO 639-3, \textbf{\#Lb:} the label size of a dataset. \textbf{M-F1:} Macro-F1, \textbf{W-F1:} Weighted-F1.} \label{tab:emotion_data}
\end{table*}

%% file: table_fig/humor.tex
\begin{table*}[h]
\centering 
\tiny
\begin{tabular}{@{}llcp{0.2\linewidth}lcp{0.1\linewidth}rc@{}}
 \toprule
\textbf{Dataset}	 & \textbf{Study}	 & \textbf{Lang}	 & \textbf{Source / Domain}	 & \textbf{Year}	 & \textbf{\#Lb}	 & \textbf{Labels}	 & \textbf{Data Slipt}	 & \textbf{Metric}\\ \midrule
Humo-hin\textsubscript{Agg} 	 & \citet{aggarwal2020sarcasm} 	 & hin 	 & Twitter 	 & 2018 	 & 2 	 & \{Humor, Not\} 	 & 4,187/524/523 	 & Accuracy	\\ 
Humo-rus\textsubscript{Bli} 	 & \citet{blinov-2019-large} 	 & rus 	 & Misc 	 & 2018 	 & 2 	 & \{Humor, Not\} 	 & 251,416/61,794/1,877 	 & M-F1	\\ 
Humo-spa\textsubscript{Chi} 	 & \citet{chiruzzo2021overview} 	 & spa 	 & Twitter 	 & 2019 	 & 2 	 & \{Humor, Not\} 	 & 24,000/6,000/6,000 	 & M-F1	\\ 
Humo-eng\textsubscript{Mea} 	 & \citet{meaney2021hahackathon} 	 & eng 	 & Twitter 	 & 2021 	 & 2 	 & \{Humor, Not\} 	 & 8,000/1,000/1,000 	 & M-F1	\\ 
\bottomrule 
\end{tabular}
\caption{Description of four humor detection datasets. \textbf{Lang:} Language is marked by ISO 639-3, \textbf{\#Lb:} the label size of a dataset. \textbf{M-F1:} Macro-F1.}\label{tab:humor_data}
\end{table*}

%% file: table_fig/irony.tex
\begin{table*}[h]
\centering 
\tiny
\begin{tabular}{@{}llcp{0.1\linewidth}lcp{0.2\linewidth}rc@{}}
 \toprule
\textbf{Dataset}	 & \textbf{Study}	 & \textbf{Lang.}	 & \textbf{Source / Domain}	 & \textbf{Year}	 & \textbf{\#Lb}	 & \textbf{Labels}	 & \textbf{Data Slipt}	 & \textbf{Metric}\\ \midrule
Iron-ita\textsubscript{Bas} 	 & \citet{basile2014overview} 	 & ita 	 & Twitter 	 & 2014 	 & 2 	 & \{Irony, Not\} 	 & 4,062/453/1,936 	 & M-F1	\\ 
Iron-spa\textsubscript{Bar} 	 & \citet{barbieri2016overview} 	 & spa 	 & Twitter 	 & 2014 	 & 2 	 & \{Irony, Not\} 	 & 6,669/741/1,997 	 & M-F1	\\ 
Iron-eng\textsubscript{Hee} 	 & \citet{van-hee2018semeval} 	 & eng 	 & Twitter 	 & 2018 	 & 2 	 & \{Irony, Not\} 	 & 3,450/384/784 	 & F1-irony	\\ 
Iron-ita\textsubscript{Cig} 	 & \citet{cignarella2018overview} 	 & ita 	 & Twitter 	 & 2018 	 & 2 	 & \{Irony, Not\} 	 & 3,579/398/872 	 & M-F1	\\ 
Iron-hin\textsubscript{Vij} 	 & \citet{vijay-2018-dataset-irony} 	 & hin 	 & Twitter 	 & 2018 	 & 2 	 & \{Irony, Not\} 	 & 2,217/277/277 	 & M-F1	\\ 
Iron-ara\textsubscript{Gha} 	 & \citet{idat2019ghanem} 	 & ara 	 & Twitter 	 & 2019 	 & 2 	 & \{Irony, Not\} 	 & 3,622/402/1,006 	 & M-F1	\\ 
Iron-spa\textsubscript{Ort} 	 & \citet{ortega2019overview} 	 & spa 	 & Twitter 	 & 2019 	 & 2 	 & \{Irony, Not\} 	 & 2,160/240/600 	 & M-F1	\\ 
Iron-fas\textsubscript{Gol}\textsuperscript{$\star$} 	 & \citet{golazizian2020mirasirony} 	 & fas 	 & Twitter 	 & 2019 	 & 2 	 & \{Irony, Not\} 	 & 2,352/295/294 	 & Accuracy	\\ 
Iron-zho\textsubscript{Xia}\textsuperscript{$\star$} 	 & \citet{xiang2020ciron} 	 & zho 	 & Weibo 	 & 2020 	 & 5 	 & \{Insufficient\_Evidence, Irony, Not, Unlikely\_Ironic, Weakly\_Irony\} 	 & 7,014/876/876 	 & M-F1	\\  \midrule
Sarc-eng\textsubscript{Wal} 	 & \citet{walker2012corpus} 	 & eng 	 & Debate Forum 	 & 2012 	 & 2 	 & \{Not, Sarcasm\} 	 & 900/100/995 	 & M-F1	\\ 
Sarc-eng\textsubscript{Ril} 	 & \citet{riloff2013sarcasm} 	 & eng 	 & Twitter 	 & 2013 	 & 2 	 & \{Not, Sarcasm\} 	 & 1,413/177/177 	 & F1-sarcasm	\\ 
Sarc-ces\textsubscript{Pta} 	 & \citet{ptavcek2014sarcasm} 	 & ces 	 & Twitter 	 & 2013 	 & 2 	 & \{Not, Sarcasm\} 	 & 3,977/497/497 	 & M-F1	\\ 
Sarc-eng\textsubscript{Pta} 	 & \citet{ptavcek2014sarcasm} 	 & eng 	 & Twitter 	 & 2013 	 & 2 	 & \{Not, Sarcasm\} 	 & 71,433/8,929/8,930 	 & M-F1	\\ 
Sarc-eng\textsubscript{Bam} 	 & \citet{bamman2015contextualized} 	 & eng 	 & Twitter 	 & 2015 	 & 2 	 & \{Not, Sarcasm\} 	 & 11,864/1,483/1,484 	 & Accuracy	\\ 
Sarc-eng\textsubscript{Raj} 	 & \citet{rajadesingan2015sarcasm} 	 & eng 	 & Twitter 	 & 2015 	 & 2 	 & \{Not, Sarcasm\} 	 & 41,261/5,158/5,158 	 & Accuracy	\\ 
Sarc-eng\textsubscript{Ora} 	 & \citet{oraby2016creating} 	 & eng 	 & Debate Forum 	 & 2016 	 & 2 	 & \{Not, Sarcasm\} 	 & 900/100/2,260 	 & M-F1	\\ 
Sarc-zho\textsubscript{Gon}\textsuperscript{$\star$} 	 & \citet{gong-2020-design} 	 & zho 	 & News comment 	 & 2019 	 & 2 	 & \{Not, Sarcasm\} 	 & 3,978/497/497 	 & M-F1	\\ 
Sarc-ara\textsubscript{Abu} 	 & \citet{abu-farha-magdy-2020-arabic} 	 & ara 	 & Twitter 	 & 2020 	 & 2 	 & \{Not, Sarcasm\} 	 & 7,593/844/2,110 	 & M-F1	\\ 
Sarc-ara\textsubscript{Far} 	 & \citet{abufarha-etal-2021-arsarcasm-v2} 	 & ara 	 & Twitter 	 & 2020 	 & 2 	 & \{Not, Sarcasm\} 	 & 11,293/1,255/3,000 	 & M-F1	\\ \midrule
Iron-T-eng\textsubscript{Hee} 	 & \citet{van-hee2018semeval} 	 & eng 	 & Twitter 	 & 2018 	 & 4 	 & \{Ironic\_by\_clash, Not, Other\_irony, Situational\_irony\} 	 & 3,450/384/784 	 & M-F1	\\ 
\bottomrule 
\end{tabular}
\caption{Description of $20$ irony and sarcasm detection datasets. \textbf{Lang.:} Language is marked by ISO 639-3, \textbf{\#Lb:} the label size of a dataset. \textbf{M-F1:} Macro-F1.  \textsuperscript{$\star$} indicates that data sharing needs an approval from the original authors. } \label{tab:irony_data}
\end{table*}

%% file: table_fig/sentiment.tex
\begin{table*}[h]
\centering 
\tiny
\begin{tabular}{lp{0.15\linewidth}cp{0.12\linewidth}lcp{0.16\linewidth}rc}
 \toprule
\textbf{Dataset}	 & \textbf{Study}	 & \textbf{Lang.}	 & \textbf{Source / Domain}	 & \textbf{Year}	 & \textbf{\#Lb}	 & \textbf{Labels}	 & \textbf{Data Slipt}	 & \textbf{Metric}\\ \midrule
Sent-eng\textsubscript{Pan} 	 & \citet{pang2005seeing} 	 & eng 	 & Moview review 	 & 2005 	 & 2 	 & \{Negative, Positive\} 	 & 8,529/1,066/1,067 	 & Accuracy	\\ 
Sent-zho\textsubscript{Tan} 	 & \citet{tan-2008-empirical} 	 & zho 	 & Misc 	 & 2008 	 & 2 	 & \{Negative, Positive\} 	 & 9,600/1,200/1,200 	 & M-F1	\\ 
Sent-T-eng\textsubscript{The} 	 & \citet{thelwall2012sentiment} 	 & eng 	 & Twitter 	 & 2012 	 & 2 	 & \{Negative, Positive\} 	 & 900/100/1,113 	 & Accuracy	\\ 
Sent-Y-eng\textsubscript{The} 	 & \citet{thelwall2012sentiment} 	 & eng 	 & YouTube comment 	 & 2012 	 & 2 	 & \{Negative, Positive\} 	 & 900/100/1,142 	 & Accuracy	\\ 
Sent-5-eng\textsubscript{Soc} 	 & \citet{socher-2013-recursive} 	 & eng 	 & Moview review 	 & 2013 	 & 5 	 & \{Negative, Neutral, Positive, Very\_Negative, Very\_Positive\} 	 & 8,544/1,101/2,210 	 & Accuracy	\\ 
Sent-kor\textsubscript{Jan}\textsuperscript{$\star$} 	 & \citet{jang-2013-kosac} 	 & kor 	 & News article 	 & 2013 	 & 4 	 & \{Complex, Negative, Neutral, Positive\} 	 & 4,187/523/524 	 & M-F1	\\ 
Sent-eng\textsubscript{Soc} 	 & \citet{socher-2013-recursive} 	 & eng 	 & Moview review 	 & 2013 	 & 2 	 & \{Negative, Positive\} 	 & 6,920/872/1,821 	 & Accuracy	\\ 
Sent-ita\textsubscript{Bas} 	 & \citet{basile2014overview} 	 & ita 	 & Twitter 	 & 2014 	 & 2 	 & \{Negative, Positive\} 	 & 2,376/265/1,207 	 & M-F1	\\ 
Sent-ita\textsubscript{Bas} 	 & \citet{barbieri2016overview} 	 & ita 	 & Twitter 	 & 2016 	 & 2 	 & \{Negative, Positive\} 	 & 3,738/416/1,018 	 & M-F1	\\ 
Sent-mlt\textsubscript{Din} 	 & \citet{dingli-2016-sentiment} 	 & mlt 	 & Moview review 	 & 2016 	 & 2 	 & \{Negative, Positive\} 	 & 596/85/171 	 & M-F1	\\ 
Sent-bul\textsubscript{Moz} 	 & \citet{mozetivc2016multilingual} 	 & bul 	 & Twitter 	 & 2016 	 & 3 	 & \{Negative, Neutral, Positive\} 	 & 22,184/2,773/2,773 	 & M-F1	\\ 
Sent-bos\textsubscript{Moz} 	 & \citet{mozetivc2016multilingual} 	 & bos 	 & Twitter 	 & 2016 	 & 3 	 & \{Negative, Neutral, Positive\} 	 & 16,335/2,042/2,042 	 & M-F1	\\ 
Sent-deu\textsubscript{Moz} 	 & \citet{mozetivc2016multilingual} 	 & deu 	 & Twitter 	 & 2016 	 & 3 	 & \{Negative, Neutral, Positive\} 	 & 42,010/5,251/5,252 	 & M-F1	\\ 
Sent-eng\textsubscript{Moz} 	 & \citet{mozetivc2016multilingual} 	 & eng 	 & Twitter 	 & 2016 	 & 3 	 & \{Negative, Neutral, Positive\} 	 & 34,538/4,317/4,318 	 & M-F1	\\ 
Sent-spa\textsubscript{Moz} 	 & \citet{mozetivc2016multilingual} 	 & spa 	 & Twitter 	 & 2016 	 & 3 	 & \{Negative, Neutral, Positive\} 	 & 122,410/15,301/15,302 	 & M-F1	\\ 
Sent-hrv\textsubscript{Moz} 	 & \citet{mozetivc2016multilingual} 	 & hrv 	 & Twitter 	 & 2016 	 & 3 	 & \{Negative, Neutral, Positive\} 	 & 52,971/6,621/6,622 	 & M-F1	\\ 
Sent-hun\textsubscript{Moz} 	 & \citet{mozetivc2016multilingual} 	 & hun 	 & Twitter 	 & 2016 	 & 3 	 & \{Negative, Neutral, Positive\} 	 & 32,717/4,089/4,090 	 & M-F1	\\ 
Sent-pol\textsubscript{Moz} 	 & \citet{mozetivc2016multilingual} 	 & pol 	 & Twitter 	 & 2016 	 & 3 	 & \{Negative, Neutral, Positive\} 	 & 87,941/10,993/10,992 	 & M-F1	\\ 
Sent-por\textsubscript{Moz} 	 & \citet{mozetivc2016multilingual} 	 & por 	 & Twitter 	 & 2016 	 & 3 	 & \{Negative, Neutral, Positive\} 	 & 39,525/4,941/4,940 	 & M-F1	\\ 
Sent-rus\textsubscript{Moz} 	 & \citet{mozetivc2016multilingual} 	 & rus 	 & Twitter 	 & 2016 	 & 3 	 & \{Negative, Neutral, Positive\} 	 & 32,941/4,117/4,118 	 & M-F1	\\ 
Sent-slk\textsubscript{Moz} 	 & \citet{mozetivc2016multilingual} 	 & slk 	 & Twitter 	 & 2016 	 & 3 	 & \{Negative, Neutral, Positive\} 	 & 30,694/3,837/3,837 	 & M-F1	\\ 
Sent-slv\textsubscript{Moz} 	 & \citet{mozetivc2016multilingual} 	 & slv 	 & Twitter 	 & 2016 	 & 3 	 & \{Negative, Neutral, Positive\} 	 & 59,924/7,491/7,490 	 & M-F1	\\ 
Sent-sqi\textsubscript{Moz} 	 & \citet{mozetivc2016multilingual} 	 & sqi 	 & Twitter 	 & 2016 	 & 3 	 & \{Negative, Neutral, Positive\} 	 & 29,375/3,672/3,672 	 & M-F1	\\ 
Sent-srp\textsubscript{Moz} 	 & \citet{mozetivc2016multilingual} 	 & srp 	 & Twitter 	 & 2016 	 & 3 	 & \{Negative, Neutral, Positive\} 	 & 22,124/2,765/2,766 	 & M-F1	\\ 
Sent-swe\textsubscript{Moz} 	 & \citet{mozetivc2016multilingual} 	 & swe 	 & Twitter 	 & 2016 	 & 3 	 & \{Negative, Neutral, Positive\} 	 & 27,277/3,409/3,410 	 & M-F1	\\ 
Sent-deu\textsubscript{Rei} 	 & \citet{rei2016multilingual} 	 & deu 	 & Twitter 	 & 2016 	 & 3 	 & \{Negative, Neutral, Positive\} 	 & 2,701/337/338 	 & M-F1	\\ 
Sent-spa\textsubscript{Rei} 	 & \citet{rei2016multilingual} 	 & spa 	 & Twitter 	 & 2016 	 & 3 	 & \{Negative, Neutral, Positive\} 	 & 6,099/763/762 	 & M-F1	\\ 
Sent-ita\textsubscript{Rei} 	 & \citet{rei2016multilingual} 	 & ita 	 & Twitter 	 & 2016 	 & 3 	 & \{Negative, Neutral, Positive\} 	 & 6,818/853/852 	 & M-F1	\\ 
Sent-eng\textsubscript{Ros} 	 & \citet{rosenthal-2017-semeval} 	 & eng 	 & Twitter 	 & 2017 	 & 3 	 & \{Negative, Neutral, Positive\} 	 & 42,756/4,751/12,284 	 & M-Recall	\\ 
Sent-ben\textsubscript{Pat}\textsuperscript{$\star$} 	 & \citet{patra-2018-sentiment} 	 & ben 	 & Twitter 	 & 2015 	 & 3 	 & \{Negative, Neutral, Positive\} 	 & 2,250/250/3,038 	 & M-F1	\\ 
Sent-hin\textsubscript{Pat}\textsuperscript{$\star$} 	 & \citet{patra-2018-sentiment} 	 & hin 	 & Twitter 	 & 2015 	 & 3 	 & \{Negative, Neutral, Positive\} 	 & 11,642/1,293/5,525 	 & M-F1	\\ 
Sent-heb\textsubscript{Amr} 	 & \citet{amram-etal-2018-representations} 	 & heb 	 & Facebook 	 & 2018 	 & 2 	 & \{Negative, Positive\} 	 & 8,951/995/2,488 	 & Accuracy	\\ 
Sent-por\textsubscript{Bru} 	 & \citet{brum-2018-building} 	 & por 	 & Twitter 	 & 2017 	 & 3 	 & \{Negative, Neutral, Positive\} 	 & 45,127/5,585/5,637 	 & M-F1	\\ 
Sent-fin\textsubscript{Kaj} 	 & \citet{kajava2018cross} 	 & fin 	 & Subtitle 	 & 2016 	 & 2 	 & \{Negative, Positive\} 	 & 5,197/577/653 	 & M-F1	\\ 
Sent-fra\textsubscript{Kaj} 	 & \citet{kajava2018cross} 	 & fra 	 & Subtitle 	 & 2016 	 & 2 	 & \{Negative, Positive\} 	 & 5,198/577/653 	 & M-F1	\\ 
Sent-ita\textsubscript{Kaj} 	 & \citet{kajava2018cross} 	 & ita 	 & Subtitle 	 & 2016 	 & 2 	 & \{Negative, Positive\} 	 & 5,197/577/653 	 & M-F1	\\ 
Sent-nor\textsubscript{Vel} 	 & \citet{velldal-etal-2018-norec} 	 & nor 	 & Online review 	 & 2018 	 & 6 	 & \{Negative1, Negative2, Negative3, Positive4, Positive5, Positive6\} 	 & 34,903/4,360/4,351 	 & M-F1	\\ 
Sent-pol\textsubscript{Koc} 	 & \citet{kocon-etal-2019-multi} 	 & pol 	 & Customer review 	 & 2019 	 & 4 	 & \{Complex, Negative, Neutral, Positive\} 	 & 5,170/574/1,217 	 & M-F1	\\ 
Sent-tha\textsubscript{Sur} 	 & \citet{suriyawongkul-2019-pythainlp} 	 & tha 	 & Facebook 	 & 2019 	 & 3 	 & \{Negative, Neutral, Positive\} 	 & 21,152/2,362/2,614 	 & M-F1	\\ 
Sent-zho\textsubscript{Wan} 	 & \citet{wan-2020-s2ap} 	 & zho 	 & Weibo 	 & 2019 	 & 2 	 & \{Negative, Positive\} 	 & 95,990/11,999/11,999 	 & M-F1	\\ 
Sent-fas\textsubscript{Ash}\textsuperscript{$\star$} 	 & \citet{ashrafi2020mirasopinion} 	 & fas 	 & Customer review 	 & 2020 	 & 3 	 & \{Negative, Neutral, Positive\} 	 & 75,094/9,387/9,387 	 & M-F1	\\ 
Sent-ron\textsubscript{Dum} 	 & \citet{dumitrescu-2020-romanian} 	 & ron 	 & Customer review 	 & 2020 	 & 2 	 & \{Negative, Positive\} 	 & 16,146/1,795/11,005 	 & M-F1	\\ 
Sent-pcm\textsubscript{Oye} 	 & \citet{oyewusi-2020-semantic} 	 & pcm 	 & Twitter 	 & 2020 	 & 3 	 & \{Negative, Neutral, Positive\} 	 & 11,200/1,400/1,400 	 & M-F1	\\ 
Sent-pol\textsubscript{Ryb} 	 & \citet{rybak-etal-2020-klej} 	 & pol 	 & Customer review 	 & 2020 	 & 5 	 & \{Negative, Neutral, Positive, Very\_Negative, Very\_Positive\} 	 & 8,619/958/1,002 	 & M-F1	\\ 
Sent-ind\textsubscript{Wil} 	 & \citet{wong-2020-indonlu} 	 & ind 	 & Misc 	 & 2019 	 & 3 	 & \{Negative, Neutral, Positive\} 	 & 9,900/1,100/1,260 	 & M-F1	\\ 
Sent-ara\textsubscript{Abd} 	 & \citet{arabert-2021-abdulmageed} 	 & ara 	 & Twitter 	 & 2021 	 & 3 	 & \{Negative, Neutral, Positive\} 	 & 49,301/4,443/4,933 	 & M-F1	\\ 
Sent-bam\textsubscript{Dia} 	 & \citet{diallo-2021-bambara} 	 & bam 	 & Misc 	 & 2021 	 & 3 	 & \{Negative, Neutral, Positive\} 	 & 2,436/305/305 	 & M-F1	\\ 
Sent-ben\textsubscript{Isl} 	 & \citet{islam-etal-2021-sentnob-dataset} 	 & ben 	 & Twitter 	 & 2021 	 & 3 	 & \{Negative, Neutral, Positive\} 	 & 12,575/1,567/1,586 	 & M-F1	\\ 
Sent-mar\textsubscript{Kul} 	 & \citet{kulkarni-2021-l3cubemahasent} 	 & mar 	 & Twitter 	 & 2020 	 & 3 	 & \{Negative, Neutral, Positive\} 	 & 12,114/1,500/2,250 	 & Accuracy	\\ 
Sent-kan\textsubscript{Cha} 	 & \citet{chakravarthi-2022-dravidian} 	 & kan 	 & YouTube comment 	 & 2019 	 & 2 	 & \{Negative, Positive\} 	 & 3,995/505/502 	 & M-F1	\\ 
Sent-mal\textsubscript{Cha} 	 & \citet{chakravarthi-2022-dravidian} 	 & mal 	 & YouTube comment 	 & 2019 	 & 2 	 & \{Negative, Positive\} 	 & 8,410/1,044/1,039 	 & M-F1	\\ 
Sent-tam\textsubscript{Cha} 	 & \citet{chakravarthi-2022-dravidian} 	 & tam 	 & YouTube comment 	 & 2019 	 & 2 	 & \{Negative, Positive\} 	 & 24,063/2,966/3,047 	 & M-F1	\\ 
Sent-ara\textsubscript{Muh} 	 & \citet{mageed:2022:nadi} 	 & ara 	 & Twitter 	 & 2021 	 & 3 	 & \{Negative, Neutral, Positive\} 	 & 1,500/500/3,000 	 & Accuracy	\\ 
Sent-amh\textsubscript{Muh} 	 & \citet{yimam-etal-2020-exploring} 	 & amh 	 & Twitter 	 & 2020 	 & 3 	 & \{Negative, Neutral, Positive\} 	 & 5,984/1,497/1,999 	 & W-F1	\\ 
Sent-ary\textsubscript{Muh} 	 & \citet{muhammadSemEval2023} 	 & ary 	 & Twitter 	 & 2021 	 & 3 	 & \{Negative, Neutral, Positive\} 	 & 5,583/494/2,961 	 & W-F1	\\ 
Sent-arq\textsubscript{Muh} 	 & \citet{muhammadSemEval2023} 	 & arq 	 & Twitter 	 & 2021 	 & 3 	 & \{Negative, Neutral, Positive\} 	 & 1,651/414/958 	 & W-F1	\\ 
Sent-hau\textsubscript{Muh} 	 & \citet{muhammad-2022-naijasenti} 	 & hau 	 & Twitter 	 & 2021 	 & 3 	 & \{Negative, Neutral, Positive\} 	 & 14,172/2,677/5,303 	 & W-F1	\\ 
Sent-ibo\textsubscript{Muh} 	 & \citet{muhammad-2022-naijasenti} 	 & ibo 	 & Twitter 	 & 2021 	 & 3 	 & \{Negative, Neutral, Positive\} 	 & 10,192/1,841/3,682 	 & W-F1	\\ 
Sent-pcm\textsubscript{Muh} 	 & \citet{muhammad-2022-naijasenti} 	 & pcm 	 & Twitter 	 & 2021 	 & 3 	 & \{Negative, Neutral, Positive\} 	 & 5,121/1,281/4,154 	 & W-F1	\\ 
Sent-kin\textsubscript{Muh} 	 & \citet{muhammad-2022-naijasenti} 	 & kin 	 & Twitter 	 & 2021 	 & 3 	 & \{Negative, Neutral, Positive\} 	 & 3,302/827/1,026 	 & W-F1	\\ 
Sent-swh\textsubscript{Muh} 	 & \citet{muhammad-2022-naijasenti} 	 & swh 	 & Twitter 	 & 2021 	 & 3 	 & \{Negative, Neutral, Positive\} 	 & 1,810/453/748 	 & W-F1	\\ 
Sent-tso\textsubscript{Muh} 	 & \citet{muhammad-2022-naijasenti} 	 & tso 	 & Twitter 	 & 2021 	 & 3 	 & \{Negative, Neutral, Positive\} 	 & 804/203/254 	 & W-F1	\\ 
Sent-twi\textsubscript{Muh} 	 & \citet{muhammad-2022-naijasenti} 	 & twi 	 & Twitter 	 & 2021 	 & 3 	 & \{Negative, Neutral, Positive\} 	 & 3,481/388/949 	 & W-F1	\\ 
Sent-yor\textsubscript{Muh} 	 & \citet{muhammad-2022-naijasenti} 	 & yor 	 & Twitter 	 & 2021 	 & 3 	 & \{Negative, Neutral, Positive\} 	 & 8,522/2,090/4,515 	 & W-F1	\\ 
Sent-yor\textsubscript{Sho} 	 & \citet{shode-2021-africanlp} 	 & yor 	 & Misc 	 & 2021 	 & 2 	 & \{Negative, Positive\} 	 & 800/200/500 	 & M-F1	\\ 
Sent-jpn\textsubscript{Suz} 	 & \citet{suzuki-etal-2022-japanese} 	 & jpn 	 & SNS post 	 & 2021 	 & 5 	 & \{Negative, Neutral, Positive, Very\_Negative, Very\_Positive\} 	 & 30,000/2,500/2,500 	 & Accuracy	\\ 
Sent-ace\textsubscript{Win} 	 & \citet{nusax-2020-genta} 	 & ace 	 & Trans. of online com. 	 & 2022 	 & 3 	 & \{Negative, Neutral, Positive\} 	 & 500/100/400 	 & M-F1	\\ 
Sent-ban\textsubscript{Win} 	 & \citet{nusax-2020-genta} 	 & ban 	 & Trans. of online com. 	 & 2022 	 & 3 	 & \{Negative, Neutral, Positive\} 	 & 500/100/400 	 & M-F1	\\ 
Sent-bbc\textsubscript{Win} 	 & \citet{nusax-2020-genta} 	 & bbc 	 & Trans. of online com. 	 & 2022 	 & 3 	 & \{Negative, Neutral, Positive\} 	 & 500/100/400 	 & M-F1	\\ 
Sent-bjn\textsubscript{Win} 	 & \citet{nusax-2020-genta} 	 & bjn 	 & Trans. of online com. 	 & 2022 	 & 3 	 & \{Negative, Neutral, Positive\} 	 & 500/100/400 	 & M-F1	\\ 
Sent-bug\textsubscript{Win} 	 & \citet{nusax-2020-genta} 	 & bug 	 & Trans. of online com. 	 & 2022 	 & 3 	 & \{Negative, Neutral, Positive\} 	 & 500/100/400 	 & M-F1	\\ 
Sent-jav\textsubscript{Win} 	 & \citet{nusax-2020-genta} 	 & jav 	 & Trans. of online com. 	 & 2022 	 & 3 	 & \{Negative, Neutral, Positive\} 	 & 500/100/400 	 & M-F1	\\ 
Sent-mad\textsubscript{Win} 	 & \citet{nusax-2020-genta} 	 & mad 	 & Trans. of online com. 	 & 2022 	 & 3 	 & \{Negative, Neutral, Positive\} 	 & 500/100/400 	 & M-F1	\\ 
Sent-min\textsubscript{Win} 	 & \citet{nusax-2020-genta} 	 & min 	 & Trans. of online com. 	 & 2022 	 & 3 	 & \{Negative, Neutral, Positive\} 	 & 500/100/400 	 & M-F1	\\ 
Sent-nij\textsubscript{Win} 	 & \citet{nusax-2020-genta} 	 & nij 	 & Trans. of online com. 	 & 2022 	 & 3 	 & \{Negative, Neutral, Positive\} 	 & 500/100/400 	 & M-F1	\\ 
Sent-sun\textsubscript{Win} 	 & \citet{nusax-2020-genta} 	 & sun 	 & Trans. of online com. 	 & 2022 	 & 3 	 & \{Negative, Neutral, Positive\} 	 & 500/100/400 	 & M-F1	\\ 
Sent-tel\textsubscript{Mar} 	 & \citet{mounika-2022-resource} 	 & tel 	 & Misc 	 & 2022 	 & 3 	 & \{Negative, Neutral, Positive\} 	 & 24,599/3,510/7,033 	 & M-F1	\\ 
\bottomrule 
\end{tabular}
\caption{Description of $77$ sentiment analysis datasets. \textbf{Lang.:} Language is marked by ISO 639-3, \textbf{\#Lb:} the label size of a dataset. \textbf{M-F1:} Macro-F1, \textbf{M-Recall:} Macro-Recall, \textbf{Trans. of online com.}: Trans. of online com. \textsuperscript{$\star$} indicates that data sharing needs approval from the original authors. }\label{tab:sentiment_data}
\end{table*}

%% file: table_fig/subject.tex
\begin{table*}[h]
\centering 
\tiny
\begin{tabular}{llcp{0.13\linewidth}lcp{0.15\linewidth}rc}
 \toprule
\textbf{Dataset}	 & \textbf{Study}	 & \textbf{Lang.}	 & \textbf{Source / Domain}	 & \textbf{Year}	 & \textbf{\#Lb}	 & \textbf{Labels}	 & \textbf{Data Slipt}	 & \textbf{Metric}\\ \midrule
Subj-eng\textsubscript{Pan} 	 & \citet{pang-2004-sentimental} 	 & eng 	 & Moview review 	 & 2004 	 & 2 	 & \{Objective, Subjective\} 	 & 8,100/900/1,000 	 & Accuracy	\\ 
Subj-kor\textsubscript{Jan}\textsuperscript{$\star$} 	 & \citet{jang-2013-kosac} 	 & kor 	 & News article 	 & 2013 	 & 7 	 & \{Agreement, Argument, Emotion, Intention, Judgment, Others, Speculation\} 	 & 4,284/535/536 	 & M-F1	\\ 
Subj-ita\textsubscript{Bas} 	 & \citet{basile2014overview} 	 & ita 	 & Twitter 	 & 2014 	 & 2 	 & \{Objective, Subjective\} 	 & 4,061/452/1,935 	 & M-F1	\\ 
Subj-ita\textsubscript{Bas} 	 & \citet{barbieri2016overview} 	 & ita 	 & Twitter 	 & 2016 	 & 2 	 & \{Objective, Subjective\} 	 & 6,669/741/1,943 	 & M-F1	\\ 
Subj-spa\textsubscript{Bar} 	 & \citet{barbieri2016overview} 	 & spa 	 & Twitter 	 & 2014 	 & 2 	 & \{Objective, Subjective\} 	 & 6,669/741/1,998 	 & M-F1	\\ 
Subj-ces\textsubscript{Pri} 	 & \citet{priban-2022-czech} 	 & ces 	 & Moview review 	 & 2021 	 & 2 	 & \{Objective, Subjective\} 	 & 7,500/500/2,000 	 & Accuracy	\\ 
\bottomrule 
\end{tabular}
\caption{Description of four subjectivity analysis datasets. \textbf{Lang.:} Language is marked by ISO 639-3, \textbf{\#Lb:} the label size of a dataset. \textbf{M-F1:} Macro-F1. \textsuperscript{$\star$} indicates that data sharing needs an approval from the original authors. }\label{tab:subj_data}
\end{table*}

%% file: append_models.tex
\section{Models}\label{append:models}
\subsection{Finetuning on Encoder-only LLMs}
%\subsection{SoTA M-PLMs}
%Research on NLP has been revolutionized by transformer-encoder-based PLMs such as BERT~\cite{devlin-2019-bert}.  M-PLMs~\cite{devlin-2019-bert,conneau2020unsupervised} can learn powerful cross-lingual representations and transfer learned knowledge across languages. PLMs are initially trained on a large corpus with self-supervised objectives (e.g., masked language modelling [MLM] and next sentence predication [NSP]). After that, a PLM is fine-tuned on downstream tasks with a labeled dataset. %To evaluate the transferability of PLMs, previous work introduces evaluation benchmarks~\cite{superglue-2019-wang, wang-etal-2018-glue}. However, previous studies have not systematically investigated model capacity on SM tasks. To fill this gap, 
We evaluate the following Transformer-encoder-based multilingual PLMs on SPARROW. We finetune each PLMs on the full training set and update all the parameters of the model during the training. 

\textbf{(1) Multilingual-BERT} (mBERT)~\cite{devlin-2019-bert} is trained on a Wikipedia corpus including $104$ languages with masked language modelling (MLM) and next sentence prediction objectives. It contains $110$M parameters. mBERT tokenizes text by using WordPiece with a vocabulary size of $172$K. 

\textbf{(2) XLM-RoBERTa\textsubscript{Base}} (XLM-R)~\cite{conneau2020unsupervised} is trained on CommonCrawl data involving $100$ languages with MLM objective. It uses a SentencePiece tokenizer with a vocabulary size of $250$K and contains $270$M parameters. 

% \textbf{(3) XLM-Twitter} (XLM-T)~\cite{barbieri-etal-2022-xlm} takes XLM-R\textsubscript{B} model and continues pretraining it on $198$M tweets with MLM objective. 

% \textbf{(4) TwHIN-BERT}~\cite{zhang-2022-twhinbert} is trained on $7$B tweets from $100$ languages with MLM objective and a graph-based contrastive social objective. TwHIN-BERT contains $280$M parameters and a SentencePiece tokenizer with a vocabulary size of $250$K.

\textbf{(3) Bernice}~\cite{alexandra-2022-bernice} is trained with $2.5$B tweets in $66$ languages and MLM objective. Bernice consists of $270$M parameters and a tweet-specific SentencePiece tokenizer including a vocabulary size of $250$K. %We finetune each model on the full Train set of each task, identify the best model based on the model's Dev performance, and report the Test performance of the best model. 

\textbf{(4) InfoDCL}~\cite{zhang2023contrastive} further trains XLM-R with $100$M tweets in $66$ languages with two contrastive learning, MLM, and distant label prediction objectives. InfoDCL shows that it effectively learns language representations for understanding SM.

\subsection{Zero-shot Setting on LLMs}
We also investigate the zero-shot performance on a wide range of LLMs: 

\textbf{(1) BLOOM}~\cite{scao-2022-bloom} is a Transformer decoder-only model trained on the ROOTS corpus consisting of $46$ natural and $13$ programming languages. BLOOM uses a multilingual vocabulary with $250$K tokens and is trained with auto-regressive language modelling objectives.

\textbf{(2) Multilingual T5 (mT5)}~\cite{xue-2021-mt5} is Transformer encoder-decoder model trained on CommonCrawl data involving $101$ languages and contains a vocabulary with $250$K tokens. It trained with sequence-to-sequence MLM objective.

\textbf{(3) LLaMA}~\cite{touvron-2023-llama} is a Transformer decoder-only model pretrained on 1.4T tokens where the majority are English and a small amount of data in 20 other languages. We utilize LLaMA with $7$B parameters and a vocabulary with $30$K tokens.

\textbf{(4) BLOOMZ}~\cite{muennighoff-2022-crosslingual} is also an instruction finetune model. It further finetunes BLOOM on xP3 corpus that contains $13$ type of tasks in $46$ languages with English prompt. We benchmark SPARROW on the BLOOM-based models with a size of $7.1$B parameters. 

\textbf{(5) BLOOMZ-P3}~\cite{muennighoff-2022-crosslingual} is an instruction finetuned model. It is initialized by BLOOM and further finetunes on English-only P3 corpus~\cite{sanh-2022-multitask} containing $2,073$ natural language prompts for eight types of NLP tasks. 

\textbf{(6) BLOOM-Bactrian}~\cite{li-2023-bactrian} tune BLOOM on a $3.4$M instruction-following dataset in $52$ languages with low-rank adaptation modules. \citet{li-2023-bactrian} translate the English $67$K instructions from Alpaca and Dolly datasets into $51$ languages and utilize ChatGPT API to generate responses in the corresponding language. 

\textbf{(7) mT0}~\cite{muennighoff-2022-crosslingual} is instruction fine-tuned mT5 model with xP3 corpus. We evaluate the mT5-based models with XL size (with $3.7$B parameters).

\textbf{(8) Alpaca}~\cite{alpaca-2023-taori} further tune LLaMA on a $52$K instruction-following dataset that is generated by \texttt{gpt-3.5-turbo} of OpenAI API. The dataset includes diverse English instruction-following tasks, e.g., question answering and programming.

\textbf{(9) Vicuna}~\cite{vicuna2023-chiang} further tune LLaMA on $70$K diverse user-shared conversations with ChatGPT in English.

\textbf{(10) ChatGPT} is a conversation-based LLM trained GTP-3~\cite{brown-2020-gpt3} through reinforcement learning with human feedback~\cite{ouyang-2022-instructgpt, christiano-2017-deep}. We expolit \texttt{gpt-3.5-turbo-0301} via OpenAI API.\footnote{\url{https://openai.com/}} 
% For all PLMs, we utilize their public released checkpoints from Huggingface Models.\footnote{\url{https://huggingface.co/models}}

% \subsection{Intermediate Fine-Tuning for SM}
% While previous studies have trained m-PLMs with billions of social media data, these m-PLMs are not intentionally trained for capturing SM. \citet{felbo2017using, zhang-2022-improving} show that intermediate fine-tuning of a PLM on surrogate label prediction (SLP), e.g., emoji prediction objective, results in rich representations for many SM tasks. This approach also needs with significantly less data than training a domain-specific model from scratch. We thus adopt intermediate fine-tune of XLM-R on SLP. To this end, we prepare an emoji-based surrogate labeled dataset that contains a set of $1,067$ emojis in $111$M tweets from $66$ languages\footnote{Language is labeled by Twitter API. More details about this dataset are in Section~\ref{sec:data_slp} in Appendix.}. We fine-tune XLM-R on the SLP objective for ten epochs with a batch size of $4,096$ and a constant learning rate of $5e-5$. We train this model on four NVIDIA V100 $32$G GPUs, and each epoch takes $19$ hours. We refer to the resulting model as SLP. 

%% file: table_fig/new_prompt.tex
\begin{table*}[h]
\centering 
\tiny
\begin{tabular}{lll}
 \toprule
& \textbf{Dataset}	 & \textbf{Prompt} \\ 
\noalign{\smallskip}\hline\noalign{\smallskip}
\multirow{27}{*}{\rotatebox[origin=c]{90}{\textbf{Anti-social}}} & Dang-ara\textsubscript{Als} 	 & \makecell[l]{\textcolor{blue}{\{Content\}} \\ Question: Is the language of this sentence \textcolor{red}{\{labels\}}? \\ Answer: } \\
\noalign{\smallskip}\cdashline{2-3}\noalign{\smallskip}
& Aggr-hin\textsubscript{Kum}, Hate-eng\textsubscript{Dav}, Hate-eng\textsubscript{Was}, Hate-ara\textsubscript{Ala}, Hate-ita\textsubscript{Bos}\textsuperscript, Hate-fil\textsubscript{Cab} 	 & \multirow{4}{*}{\makecell[l]{\textcolor{blue}{\{Content\}} \\ Question: Is the language of this text \textcolor{red}{\{labels\}}? \\ Answer: }}	\\
& Hate-ara\textsubscript{Mul}, Hate-eng\textsubscript{Bas}, Hate-spa\textsubscript{Bas}, Hate-por\textsubscript{For}, Hate-pol\textsubscript{Pta}, Hate-kor\textsubscript{Moo} 	 & \\
& Hate-ara\textsubscript{Mub}, Hate-zho\textsubscript{Den}, Hate-kor\textsubscript{Jeo}, Hate-tel\textsubscript{Mar}, Sexi-fra\textsubscript{Chi}, Offe-eng\textsubscript{Zam} 	 & \\ 
& Offe-ara\textsubscript{Zam}, Offe-dan\textsubscript{Zam}, Offe-ell\textsubscript{Zam}, Offe-tur\textsubscript{Zam}, Offe-ara\textsubscript{Mub}, Offe-slv\textsubscript{Nov} 	 & \\ \noalign{\smallskip}\cdashline{2-3}\noalign{\smallskip}
& Offe-T-kan\textsubscript{Cha}, Offe-G-eng\textsubscript{Zam}, Hate-T-kor\textsubscript{Jeo} 	 & \makecell[l]{\textcolor{blue}{\{Content\}} \\ Question: Does this offensive text target \textcolor{red}{\{labels\}}? \\ Answers: } \\
\noalign{\smallskip}\cdashline{2-3}\noalign{\smallskip}
& Hate-G-ara\textsubscript{Ous}, Hate-G-fra\textsubscript{Ous} 	 & \makecell[l]{\textcolor{blue}{\{Content\}} \\ Question: Does this hate speech target \textcolor{red}{\{labels\}}? \\ Answer: } \\
\noalign{\smallskip}\cdashline{2-3}\noalign{\smallskip}
& Offe-T-eng\textsubscript{Zam} 	 & \makecell[l]{\textcolor{blue}{\{Content\}} \\ Question: Is this offensive text \textcolor{red}{\{labels\}} insult? \\ Answer: } \\
\noalign{\smallskip}\cdashline{2-3}\noalign{\smallskip}
& Hate-T-ara\textsubscript{Ous}, Hate-T-fra\textsubscript{Ous} 	 & \makecell[l]{\textcolor{blue}{\{Content\}} \\ Question: Does this hate speech text insult against people based on \\ their attribute of \textcolor{red}{\{labels\}}? \\ Answer: }
\\ \noalign{\smallskip}\cdashline{2-3}\noalign{\smallskip}
& Hate-T-ben\textsubscript{Kar} 	 & \makecell[l]{\textcolor{blue}{\{Content\}} \\ Question: Does this text express \textcolor{red}{\{labels\}}? \\ Answer: } \\
\noalign{\smallskip}\cdashline{2-3}\noalign{\smallskip}
& Offe-T-mal\textsubscript{Cha}, Offe-T-tam\textsubscript{Cha} 	 & \makecell[l]{\textcolor{blue}{\{Content\}} \\ Question: Is this sentence hate speech or not? If yes, does this sentence \\ target individual, group or not? \\ Answer: } \\ 
\noalign{\smallskip}\hline\noalign{\smallskip}
\multirow{4}{*}{\rotatebox[origin=c]{90}{\textbf{Emotion}}} & Emot-eng\textsubscript{Wal}, Emot-zho\textsubscript{Lee}, Emot-fin\textsubscript{Kaj}, Emot-fra\textsubscript{Kaj}, Emot-ita\textsubscript{Kaj}, Emot-msa\textsubscript{Hus}, Emot-ara\textsubscript{Abd} 	 & 	\multirow{4}{*}{\makecell[l]{\textcolor{blue}{\{Content\}} \\ Question: Is the emotion of this sentence \textcolor{red}{\{labels\}}? \\ Answer: }} \\ 
& Emot-eng\textsubscript{Moh}, Emot-ara\textsubscript{Moh}, Emot-spa\textsubscript{Moh}, Emot-ind\textsubscript{Sap}, Emot-tur\textsubscript{Guv}, Emot-spa\textsubscript{Moh}, Emot-ind\textsubscript{Wil} 	 & 	\\ 
& Emot-vie\textsubscript{Ho}, Emot-eng\textsubscript{Pla}, Emot-spa\textsubscript{Pla}, Emot-fin\textsubscript{Ohm}, Emot-eng\textsubscript{Dem}, Emot-ita\textsubscript{Bia}, Emot-ron\textsubscript{Cio} 	 & 	\\
& Emot-hin\textsubscript{Deb}, Emot-por\textsubscript{Cor}, Emot-fas\textsubscript{Sab}, Emot-rus\textsubscript{Sbo}, Emot-ben\textsubscript{Iqb}, Emot-fra\textsubscript{Bia}, Emot-deu\textsubscript{Bia} 	 & 	\\
\noalign{\smallskip}\hline\noalign{\smallskip}
\rotatebox[origin=c]{90}{\textbf{Humor}} & Humo-hin\textsubscript{Agg}, Humo-rus\textsubscript{Bli}, Humo-spa\textsubscript{Chi}, Humo-eng\textsubscript{Mea} & \makecell[l]{\textcolor{blue}{\{Content\}} \\ Question: Is this sentence \textcolor{red}{\{labels\}}? \\ Answer: } \\
\noalign{\smallskip}\hline\noalign{\smallskip}
\multirow{6}{*}{\rotatebox[origin=c]{90}{\textbf{Irony}}} & Iron-ita\textsubscript{Bas}, Iron-spa\textsubscript{Bar}, Iron-eng\textsubscript{Hee}, Iron-ita\textsubscript{Cig}, Iron-hin\textsubscript{Vij}, Iron-ara\textsubscript{Gha}, Iron-spa\textsubscript{Ort} & \multirow{3}{*}{\makecell[l]{\textcolor{blue}{\{Content\}} \\ Question: Is this sentence \textcolor{red}{\{labels\}}? \\ Answer: }} \\
& Iron-fas\textsubscript{Gol}\textsuperscript, Iron-zho\textsubscript{Xia}\textsuperscript, Sarc-eng\textsubscript{Wal}, Sarc-eng\textsubscript{Ril}, Sarc-ces\textsubscript{Pta}, Sarc-eng\textsubscript{Pta}, Sarc-eng\textsubscript{Bam} \\
& Sarc-eng\textsubscript{Raj}, Sarc-eng\textsubscript{Ora}, Sarc-zho\textsubscript{Gon}\textsuperscript, Sarc-ara\textsubscript{Abu}, Sarc-ara\textsubscript{Far} \\
\noalign{\smallskip}\cdashline{2-3}\noalign{\smallskip}
& Iron-T-eng\textsubscript{Hee} 	 & \makecell[l]{\textcolor{blue}{\{Content\}} \\ Question: Is the type of this text \textcolor{red}{\{labels\}}? \\ Answer: } \\
\noalign{\smallskip}\hline\noalign{\smallskip}
\multirow{15}{*}{\rotatebox[origin=c]{90}{\textbf{Sentiment Analysis}}} & Sent-eng\textsubscript{Pan}, Sent-zho\textsubscript{Tan}, Sent-kor\textsubscript{Jan}, Sent-eng\textsubscript{Soc}, Sent-ita\textsubscript{Bas}, Sent-ben\textsubscript{Pat}, Sent-hin\textsubscript{Pat} & \multirow{10}{*}{\makecell[l]{\textcolor{blue}{\{Content\}} \\ Question: Is the sentiment of this sentence \textcolor{red}{\{labels\}}? \\ Answer: }}	\\
& Sent-ita\textsubscript{Bas}, Sent-mlt\textsubscript{Din}, Sent-bul\textsubscript{Moz}, Sent-bos\textsubscript{Moz}, Sent-deu\textsubscript{Moz}, Sent-eng\textsubscript{Moz}, Sent-spa\textsubscript{Moz} & \\
& Sent-hrv\textsubscript{Moz}, Sent-hun\textsubscript{Moz}, Sent-pol\textsubscript{Moz}, Sent-por\textsubscript{Moz}, Sent-rus\textsubscript{Moz}, Sent-slk\textsubscript{Moz}, Sent-slv\textsubscript{Moz} & \\
& Sent-sqi\textsubscript{Moz}, Sent-srp\textsubscript{Moz}, Sent-swe\textsubscript{Moz}, Sent-deu\textsubscript{Rei}, Sent-spa\textsubscript{Rei}, Sent-ita\textsubscript{Rei}, Sent-eng\textsubscript{Ros} & \\
& Sent-heb\textsubscript{Amr}, Sent-por\textsubscript{Bru}, Sent-fin\textsubscript{Kaj}, Sent-fra\textsubscript{Kaj}, Sent-ita\textsubscript{Kaj}, Sent-pol\textsubscript{Koc}, Sent-tha\textsubscript{Sur} & \\
& Sent-zho\textsubscript{Wan}, Sent-fas\textsubscript{Ash}, Sent-ron\textsubscript{Dum}, Sent-pcm\textsubscript{Oye}, Sent-pol\textsubscript{Ryb}, Sent-ind\textsubscript{Wil}, Sent-ara\textsubscript{Abd} & \\
& Sent-bam\textsubscript{Dia}, Sent-ben\textsubscript{Isl}, Sent-mar\textsubscript{Kul}, Sent-kan\textsubscript{Cha}, Sent-mal\textsubscript{Cha}, Sent-tam\textsubscript{Cha}, Sent-ara\textsubscript{Muh} & \\
& Sent-amh\textsubscript{Muh}, Sent-ary\textsubscript{Muh}, Sent-arq\textsubscript{Muh}, Sent-hau\textsubscript{Muh}, Sent-ibo\textsubscript{Muh}, Sent-pcm\textsubscript{Muh}, Sent-kin\textsubscript{Muh} & \\
& Sent-swh\textsubscript{Muh}, Sent-tso\textsubscript{Muh}, Sent-twi\textsubscript{Muh}, Sent-yor\textsubscript{Muh}, Sent-yor\textsubscript{Sho}, Sent-jpn\textsubscript{Suz}, Sent-ace\textsubscript{Win} & \\
& Sent-ban\textsubscript{Win}, Sent-bbc\textsubscript{Win}, Sent-bjn\textsubscript{Win}, Sent-bug\textsubscript{Win}, Sent-jav\textsubscript{Win}, Sent-mad\textsubscript{Win}, Sent-min\textsubscript{Win} & \\
& Sent-nij\textsubscript{Win}, Sent-sun\textsubscript{Win}, Sent-tel\textsubscript{Mar}, Sent-T-eng\textsubscript{The}, Sent-Y-eng\textsubscript{The}, Sent-5-eng\textsubscript{Soc} & \\
\noalign{\smallskip}\cdashline{2-3}\noalign{\smallskip}
& Sent-nor\textsubscript{Vel} 	 & 	\makecell[l]{\textcolor{blue}{\{Content\}} \\ Question: Is this text rated as \textcolor{red}{\{labels\}}? Higher is better. \\ Answer: } \\
\noalign{\smallskip}\hline\noalign{\smallskip}
\multirow{4}{*}{\rotatebox[origin=c]{90}{\textbf{Subjective}}} & Subj-kor\textsubscript{Jan} & \makecell[l]{\textcolor{blue}{\{Content\}} \\ Question: Does this sentence express \{labele\}? \\ Answer: } \\
\noalign{\smallskip}\cdashline{2-3}\noalign{\smallskip}
& Subj-eng\textsubscript{Pan}, Subj-ita\textsubscript{Bas}, Subj-ita\textsubscript{Bas}, Subj-spa\textsubscript{Bar}, Subj-ces\textsubscript{Pri} & \makecell[l]{\textcolor{blue}{\{Content\}} \\ Question: Is this sentence \textcolor{red}{\{labels\}}? \\ Answer: } \\
\bottomrule 
\end{tabular}
\caption{Prompts use for zero-shot evaluation with lm-evaluation-harness.}\label{tab:prompt_new}
\end{table*}

%% file: table_fig/finetune_more_results.tex
\begin{table*}[]
\tiny
\centering
\begin{tabular}{llcccc|cccc}
\toprule
                                  &            & \multicolumn{4}{c}{\textbf{Dev Set}}                       & \multicolumn{4}{c}{\textbf{Test-S Set}}                      \\ \cmidrule(r){3-6} \cmidrule(r){7-10} 
                                  &            & mBERT      & XLM-R      & Bernice    & InfoDCL    & mBERT      & XLM-R      & Bernice    & InfoDCL    \\ \midrule
\multirow{7}{*}{Antisocial}       & Aggressive & 73.39$\pm$0.30 & 73.92$\pm$0.50 & 76.79$\pm$0.52 & 76.09$\pm$0.38 & 72.71$\pm$1.92 & 74.64$\pm$0.14 & 75.45$\pm$0.73 & 73.96$\pm$0.91 \\
                                  & Dangerours & 69.76$\pm$1.56 & 69.53$\pm$1.04 & 74.92$\pm$1.18 & 73.42$\pm$0.80 & 62.36$\pm$1.08 & 63.57$\pm$1.15 & 67.13$\pm$0.56 & 65.23$\pm$1.60 \\
                                  & Hate       & 77.73$\pm$0.76 & 79.40$\pm$0.85 & 81.16$\pm$1.15 & 80.62$\pm$0.60 & 72.97$\pm$1.40 & 74.37$\pm$1.48 & 76.76$\pm$2.43 & 75.85$\pm$0.90 \\
                                  & Offense    & 78.96$\pm$1.04 & 80.21$\pm$0.92 & 82.15$\pm$0.68 & 81.55$\pm$0.46 & 77.53$\pm$1.27 & 75.88$\pm$2.43 & 78.45$\pm$1.89 & 78.88$\pm$2.63 \\
                                  & H/O-Group  & 51.57$\pm$2.35 & 41.24$\pm$3.16 & 48.30$\pm$1.90 & 46.05$\pm$2.25 & 46.18$\pm$4.10 & 42.39$\pm$3.30 & 51.15$\pm$2.01 & 50.24$\pm$4.19 \\
                                  & H/O-Target & 54.02$\pm$3.60 & 59.23$\pm$1.71 & 60.83$\pm$1.26 & 60.14$\pm$1.21 & 53.16$\pm$4.49 & 57.67$\pm$1.79 & 60.96$\pm$2.26 & 60.79$\pm$1.38 \\ \cdashline{2-10}
                                  & AS         & 70.18$\pm$1.59 & 71.47$\pm$1.24 & 73.80$\pm$1.13 & 73.04$\pm$0.85 & 66.92$\pm$2.29 & 67.99$\pm$1.84 & 71.14$\pm$2.15 & 70.61$\pm$1.64 \\ \hline 
\multicolumn{2}{l}{Emotion}                    & 62.30$\pm$0.90 & 67.43$\pm$0.68 & 68.56$\pm$0.85 & 69.34$\pm$0.53 & 61.42$\pm$1.51 & 66.87$\pm$0.99 & 68.13$\pm$1.28 & 69.27$\pm$1.03 \\ \hline
\multicolumn{2}{l}{Humor}                      & 85.15$\pm$0.32 & 85.83$\pm$0.57 & 86.72$\pm$0.63 & 86.74$\pm$0.42 & 84.35$\pm$1.23 & 85.19$\pm$1.62 & 86.75$\pm$1.13 & 87.05$\pm$0.82 \\ \hline
\multirow{4}{*}{I\&S} & Irony      & 69.16$\pm$1.29 & 69.94$\pm$1.02 & 72.19$\pm$1.24 & 71.12$\pm$0.72 & 64.24$\pm$1.16 & 65.53$\pm$1.57 & 66.88$\pm$1.23 & 68.38$\pm$1.00 \\
                                  & Sarcasm    & 74.65$\pm$1.14 & 75.64$\pm$1.92 & 77.82$\pm$1.56 & 77.21$\pm$1.05 & 72.41$\pm$1.38 & 73.40$\pm$2.42 & 74.78$\pm$1.69 & 74.94$\pm$1.13 \\
                                  & Irony-Type & 53.51$\pm$2.11 & 52.18$\pm$2.15 & 58.55$\pm$3.17 & 57.72$\pm$2.92 & 47.35$\pm$1.89 & 46.43$\pm$0.63 & 56.04$\pm$1.87 & 57.58$\pm$1.42 \\\cdashline{2-10}
                                  & I\&S       & 71.13$\pm$1.26 & 71.90$\pm$1.52 & 74.32$\pm$1.50 & 73.49$\pm$1.00 & 67.48$\pm$1.31 & 68.51$\pm$1.95 & 70.29$\pm$1.49 & 71.12$\pm$1.09 \\ \hline
\multicolumn{2}{l}{Sentiment}                  & 69.29$\pm$1.14 & 71.34$\pm$0.78 & 72.95$\pm$0.88 & 73.81$\pm$0.70 & 66.34$\pm$1.92 & 69.58$\pm$1.41 & 70.44$\pm$1.61 & 71.64$\pm$1.31 \\ \hline 
\multicolumn{2}{l}{Subjectivity}               & 75.18$\pm$0.63 & 77.28$\pm$0.74 & 76.97$\pm$0.69 & 77.78$\pm$0.87 & 72.54$\pm$1.46 & 74.45$\pm$1.18 & 74.80$\pm$1.08 & 75.73$\pm$1.33 \\ \midrule
\multicolumn{2}{l}{SM}                         & 69.29$\pm$1.17 & 71.50$\pm$0.93 & 73.16$\pm$0.98 & 73.46$\pm$0.72 & 66.60$\pm$1.83 & 69.38$\pm$1.51 & 70.85$\pm$1.63 & 71.60$\pm$1.30 \\ \bottomrule
\end{tabular}
\caption{Performance of finetuned models on Dev and Test-S set. We finetune each model on each dataset for three runs with different random seeds and calculate the mean and standard deviation of dataset-specific metrics over the three runs. We report the average of dataset-specific metrics and standard deviation in a task and a category. \textbf{I\&S:} irony and sarcasm.}\label{tab:finetune_more_results}
\end{table*}

%% file: table_fig/antisocial_res.tex
\begin{table*}[]
\centering 
\tiny
\setlength\tabcolsep{2pt}
\resizebox{\textwidth}{!}{
\begin{tabular}{@{}llc|cccc|ccccc:ccc:ccc|cc|cl@{}}
 \toprule
\textbf{Dataset}     & \textbf{Metric}   & \textbf{Random}   & \textbf{mBERT}    & \textbf{XLM-R}    & \textbf{Bernice}  & \textbf{InfoDCL}  & \textbf{BLOOM}    & \textbf{BLOOMZ}   & \multicolumn{1}{c}{\textbf{\begin{tabular}[c]{@{}c@{}}BLOOMZ\\ (MT)\end{tabular}}}  & \multicolumn{1}{c}{\textbf{\begin{tabular}[c]{@{}c@{}}BLOOMZ-\\ P3\end{tabular}}}    & \multicolumn{1}{c}{\textbf{\begin{tabular}[c]{@{}c@{}}BLOOMZ-\\ Bactrian\end{tabular}}}  & \textbf{mT5}  & \textbf{mT0}  & \multicolumn{1}{c}{\textbf{\begin{tabular}[c]{@{}c@{}}mT0\\ (MT)\end{tabular}}}     & \textbf{LLaMA}    & \textbf{Alpaca}   & \textbf{Vicuna}   & \textbf{ChatGPT}  & \multicolumn{1}{c}{\textbf{\begin{tabular}[c]{@{}c@{}}ChatGPT\\ (MT)\end{tabular}}}     & \textbf{SoTA}     & \textbf{SoTA study}\\ \midrule
Aggr-hin\textsubscript{Kum}      & M-F1      & 43.14     & 72.71     & 74.64     & \textbf{75.45}    & 73.96     & 51.06     & 15.82     & 15.82     & 18.72     & 16.37     & 53.67     & 15.82     & 22.00     & 18.31     & 49.29     & 25.07     & 63.53     & 54.36     & \underline{70.00}     & \citet{kumar-2018-aggression}    \\ 
Dang-ara\textsubscript{Als}      & M-F1      & 42.06     & 62.36     & 63.57     & \textbf{67.13}    & 65.23     & 46.87     & 46.87     & 50.84     & 46.87     & 46.87     & 49.31     & 46.87     & 46.87     & 46.87     & 46.87     & 46.87     & 37.93     & 33.68     & \underline{59.60}     & \citet{alshehri-etal-2020-understanding} \\ 
Hate-eng\textsubscript{Was}      & W-F1      & 41.57     & 87.74     & \textbf{89.33}    & 88.79     & 88.93     & 58.26     & 65.96     & 65.96     & 60.07     & 60.17     & 18.66     & 59.75     & 59.75     & 60.42     & 37.33     & 61.66     & 79.98     & 79.98     & \underline{73.62}     & \citet{waseem-2016-hateful}  \\ 
Hate-eng\textsubscript{Dav}      & W-F1      & 38.97     & 91.28     & \textbf{92.96}    & 91.99     & 91.12     & 7.99      & 10.78     & 10.78     & 25.43     & 9.32      & 13.36     & 8.33      & 8.33      & 8.33      & 73.10     & 55.89     & 69.58     & 69.58     & \underline{90.00}     & \citet{davidson-2017-hateoffensive}  \\ 
Hate-ara\textsubscript{Ala}      & M-F1      & 43.70     & 82.07     & 81.23     & 83.99     & \textbf{85.95}    & 54.28     & 43.63     & 43.63     & 43.63     & 52.08     & 24.11     & 43.63     & 43.63     & 43.63     & 43.50     & 43.63     & 63.96     & 52.85     & -     & ---  \\ 
Hate-ita\textsubscript{Bos}      & M-F1      & 47.00     & 75.67     & 76.96     & 80.19     & \textbf{81.17}    & 44.63     & 40.26     & 40.26     & 40.26     & 45.69     & 25.55     & 40.26     & 40.26     & 40.26     & 42.34     & 40.26     & 77.98     & 57.62     & \underline{79.93}     & \citet{bosco2018overview}    \\ 
Hate-fil\textsubscript{Cab}      & M-F1      & 52.37     & 74.38     & 78.40     & \textbf{79.50}    & 79.01     & 46.79     & 34.47     & 45.46     & 34.47     & 46.80     & 32.93     & 34.47     & 57.76     & 34.47     & 47.00     & 34.47     & 69.13     & 66.67     & \underline{71.12}     & \citet{cabasag2019hate}  \\ 
Hate-ara\textsubscript{Mul}      & M-F1      & 29.56     & 69.38     & 67.12     & \textbf{76.53}    & 71.74     & 15.62     & 25.46     & 25.46     & 37.90     & 19.22     & 14.50     & 25.46     & 25.46     & 25.46     & 17.16     & 30.99     & 61.13     & 32.01     & \underline{89.60}     & \citet{mulki2019hsab}    \\ 
Hate-eng\textsubscript{Bas}      & M-F1      & 52.61     & 50.24     & 51.87     & 54.17     & 53.25     & 51.11     & 36.22     & 36.22     & 36.22     & 52.86     & 29.97     & 36.22     & 36.22     & 36.65     & 53.15     & 37.96     & \textbf{63.69}    & 63.69     & \underline{65.10}     & \citet{basile-2019-semeval}  \\ 
Hate-spa\textsubscript{Bas}      & M-F1      & 44.59     & 74.18     & 76.26     & \textbf{78.20}    & 76.96     & 46.61     & 37.50     & 29.21     & 37.50     & 48.68     & 31.06     & 37.50     & 37.50     & 38.95     & 59.16     & 37.50     & 64.93     & 54.43     & \underline{73.00}     & \citet{basile-2019-semeval}  \\ 
Hate-por\textsubscript{For}      & M-F1      & 47.84     & 70.08     & 70.02     & \textbf{74.40}    & 73.22     & 48.01     & 39.69     & 39.90     & 39.69     & 51.53     & 29.73     & 39.69     & 39.69     & 40.09     & 55.27     & 39.69     & 68.45     & 63.34     & \underline{72.00}     & \citet{fortuna-2019-hierarchically}  \\ 
Hate-pol\textsubscript{Pta}      & M-F1      & 44.15     & 69.69     & 70.26     & 71.68     & 71.23     & 47.77     & 46.47     & 46.47     & 46.47     & 48.05     & 12.41     & 46.47     & 46.47     & 47.46     & 53.11     & 46.47     & \textbf{77.02}    & 64.96     & \underline{50.30}     & \citet{rybak-etal-2020-klej} \\ 
Hate-kor\textsubscript{Moo}      & M-F1      & 36.74     & 57.10     & 63.17     & \textbf{64.80}    & 63.09     & 18.03     & 16.90     & 21.94     & 20.95     & 16.90     & 15.88     & 16.90     & 20.30     & 16.90     & 27.59     & 16.90     & 39.79     & 46.90     & \underline{63.30}     & \citet{moon-etal-2020-beep}  \\ 
Hate-ara\textsubscript{Mub}      & M-F1      & 37.84     & 73.92     & 79.67     & 81.16     & \textbf{82.00}    & 35.69     & 48.67     & 48.67     & 48.67     & 34.08     & 8.36      & 48.67     & 48.67     & 48.67     & 48.51     & 48.67     & 60.72     & 52.04     & \underline{84.79}     & \citet{arabert-2021-abdulmageed} \\ 
Hate-zho\textsubscript{Den}      & M-F1      & 48.72     & 83.29     & 83.36     & 84.39     & \textbf{84.86}    & 69.31     & 36.22     & 30.17     & 36.22     & 60.96     & 30.56     & 36.22     & 38.13     & 36.73     & 36.22     & 36.22     & 72.34     & 74.24     & \underline{81.00}     & \citet{deng-2022-cold}   \\ 
Hate-kor\textsubscript{Jeo}      & M-F1      & 51.96     & 79.03     & 78.50     & \textbf{80.19}    & 79.32     & 35.55     & 33.69     & 33.69     & 33.69     & 34.75     & 35.77     & 33.69     & 34.05     & 33.87     & 40.06     & 33.69     & 63.61     & 57.09     & \underline{77.20}     & \citet{jeong-2022-kold}  \\ 
Hate-tel\textsubscript{Mar}      & M-F1      & 34.32     & 49.90     & 49.78     & \textbf{58.22}    & 49.75     & 32.67     & 49.90     & 49.90     & 49.90     & 48.19     & 0.60      & 49.90     & 49.90     & 49.90     & 49.90     & 49.90     & 49.90     & 34.23     & \underline{60.00}     & \citet{mounika-2022-resource}    \\ 
Sexi-fra\textsubscript{Chi}      & M-F1      & 46.05     & 79.60     & 81.01     & 79.96     & \textbf{81.99}    & 24.92     & 65.25     & 36.45     & 25.26     & 46.40     & 49.17     & 40.05     & 38.65     & 42.95     & 25.60     & 51.53     & 74.81     & 70.29     & \underline{76.20}     & \citet{chiril2020annotated}  \\ 
Offe-eng\textsubscript{Zam}      & M-F1      & 44.20     & 75.07     & 75.10     & 77.75     & \textbf{78.67}    & 42.06     & 42.06     & 42.06     & 42.62     & 42.06     & 23.97     & 42.06     & 42.06     & 42.06     & 25.57     & 59.67     & 67.90     & 67.90     & \underline{82.90}     & \citet{zampieri-2019-predicting} \\ 
Offe-ara\textsubscript{Zam}      & M-F1      & 45.64     & 86.27     & 86.88     & \textbf{91.55}    & 89.53     & 44.07     & 44.07     & 17.49     & 44.07     & 44.07     & 27.25     & 44.07     & 17.49     & 44.07     & 17.49     & 58.47     & 82.01     & 67.52     & \underline{90.17}     & \citet{zampieri-etal-2020-semeval}   \\ 
Offe-dan\textsubscript{Zam}      & M-F1      & 41.14     & 77.53     & 76.11     & 78.03     & \textbf{82.09}    & 46.68     & 46.68     & 11.08     & 46.68     & 46.68     & 20.35     & 46.68     & 11.08     & 46.68     & 11.46     & 51.84     & 66.91     & 66.79     & 81.19     & \citet{zampieri-etal-2020-semeval}   \\ 
Offe-ell\textsubscript{Zam}      & M-F1      & 41.24     & \textbf{80.64}    & 76.91     & 79.63     & 79.13     & 45.47     & 45.47     & 14.24     & 46.71     & 45.47     & 33.99     & 45.47     & 14.24     & 45.47     & 14.24     & 48.21     & 60.94     & 34.98     & \underline{85.22}     & \citet{zampieri-etal-2020-semeval}   \\ 
Offe-tur\textsubscript{Zam}      & M-F1      & 41.11     & 73.11     & 76.90     & \textbf{77.07}    & 76.08     & 44.38     & 44.38     & 19.92     & 44.38     & 44.38     & 38.31     & 44.38     & 44.38     & 44.38     & 16.81     & 51.12     & 75.03     & 45.68     & \underline{82.58}     & \citet{zampieri-etal-2020-semeval}   \\ 
Offe-ara\textsubscript{Mub}      & M-F1      & 46.28     & 86.85     & 86.44     & \textbf{93.16}    & 91.40     & 43.69     & 43.69     & 18.30     & 43.69     & 43.63     & 22.86     & 43.69     & 18.30     & 43.69     & 18.59     & 58.96     & 84.48     & 69.36     & \underline{90.50}     & \citet{mubarak-etal-2020-overview}  \\ 
Offe-slv\textsubscript{Nov}      & M-F1      & 16.75     & \textbf{63.23}    & 52.83     & 51.98     & 55.29     & 21.02     & 16.60     & 18.88     & 1.98      & 16.68     & 8.19      & 12.93     & 0.20      & 12.57     & 13.57     & 12.59     & 33.86     & 16.67     & -     & ---  \\ 
Offe-G-eng\textsubscript{Zam}    & M-F1      & 31.22     & 54.49     & 55.44     & \textbf{61.93}    & 61.45     & 28.72     & 36.00     & 36.00     & 44.00     & 25.66     & 19.73     & 26.95     & 26.95     & 31.46     & 30.73     & 20.68     & 51.52     & 51.52     & 75.50     & \citet{zampieri-2019-predicting} \\ 
Hate-G-ara\textsubscript{Ous}    & M-F1      & 6.89      & 46.71     & 37.11     & \textbf{52.38}    & 51.08     & 3.87      & 7.21      & 0.07      & 8.41      & 7.55      & 1.32      & 14.83     & 16.67     & 0.00      & 0.27      & 1.43      & 35.05     & 10.90     & \underline{40.00}     & \citet{ousidhoum2019multilingual}    \\ 
Hate-G-fra\textsubscript{Ous}    & M-F1      & 6.36      & 37.35     & 34.61     & \textbf{39.13}    & 38.19     & 8.29      & 8.57      & 6.61      & 11.28     & 11.23     & 0.00      & 6.98      & 7.42      & 5.61      & 11.39     & 5.66      & 32.41     & 17.79     & \underline{37.00}     & \citet{ousidhoum2019multilingual}    \\ 
Offe-T-eng\textsubscript{Zam}    & M-F1      & 39.77     & 63.61     & 64.04     & \textbf{78.16}    & 72.80     & 47.02     & 47.02     & 47.02     & 47.02     & 47.63     & 10.62     & 22.75     & 22.75     & 58.67     & 39.89     & 37.48     & 45.97     & 45.97     & 66.00     & \citet{zampieri-2019-predicting} \\ 
Hate-T-ara\textsubscript{Ous}    & M-F1      & 21.42     & 48.32     & 52.94     & 52.96     & \textbf{53.84}    & 10.15     & 19.64     & 8.65      & 15.58     & 14.25     & 16.76     & 30.19     & 11.07     & 6.48      & 7.64      & 6.48      & 44.53     & 40.74     & \underline{63.00}     & \citet{ousidhoum2019multilingual}    \\ 
Hate-T-fra\textsubscript{Ous}    & M-F1      & 14.85     & 45.88     & 43.39     & 43.30     & \textbf{48.18}    & 4.11      & 8.07      & 19.58     & 7.92      & 2.11      & 2.08      & 18.84     & 0.22      & 2.08      & 2.26      & 3.15      & 39.98     & 41.14     & \underline{43.00}     & \citet{ousidhoum2019multilingual}    \\ 
Hate-T-ben\textsubscript{Kar}    & M-F1      & 22.89     & 83.56     & 85.31     & 85.84     & \textbf{86.30}    & 11.12     & 15.52     & 22.29     & 7.53      & 12.98     & 12.11     & 23.15     & 42.74     & 18.48     & 12.76     & 23.42     & 49.03     & 46.60     & \underline{87.00}     & \citet{karim2020deephateexplainer}   \\ 
Offe-T-kan\textsubscript{Cha}    & M-F1      & 14.73     & 42.13     & 38.78     & \textbf{46.47}    & 40.65     & 16.69     & 16.69     & 9.50      & 16.61     & 16.69     & 4.27      & 8.09      & 6.51      & 20.01     & 15.44     & 14.16     & 21.19     & 9.63      & \underline{43.00}     & \citet{chakravarthi-2022-dravidian}  \\ 
Offe-T-mal\textsubscript{Cha}    & M-F1      & 12.71     & 42.12     & 74.22     & 76.51     & \textbf{81.53}    & 24.04     & 24.04     & 24.04     & 20.81     & 24.04     & 1.17      & 26.80     & 24.04     & 24.04     & 24.04     & 24.04     & 19.32     & 4.00      & \underline{72.00}     & \citet{chakravarthi-2022-dravidian}  \\ 
Offe-T-tam\textsubscript{Cha}    & M-F1      & 14.28     & 38.70     & 36.25     & \textbf{39.98}    & 39.76     & 17.32     & 17.35     & 4.39      & 15.76     & 17.35     & 2.26      & 17.90     & 17.82     & 17.22     & 17.34     & 17.35     & 25.72     & 11.61     & \underline{44.00}     & \citet{chakravarthi-2022-dravidian}  \\ 
Hate-T-kor\textsubscript{Jeo}    & M-F1      & 22.51     & 60.98     & \textbf{66.43}    & 64.47     & 63.25     & 19.37     & 3.87      & 14.46     & 3.87      & 15.14     & 4.28      & 16.91     & 18.80     & 7.56      & 15.27     & 9.96      & 41.35     & 29.67     & \underline{62.70}     & \citet{jeong-2022-kold}  \\ \midrule
Average      & ---   & 35.20     & 66.92     & 67.99     & \textbf{71.14}     & 70.61     & 33.70     & 32.80     & 27.93     & 31.97     & 33.79     & 20.14     & 32.02     & 28.79     & 31.68     & 30.55     & 34.50     & 56.55     & 47.40     & ---   & ---  \\ 
\bottomrule 
\end{tabular}
}
\caption{Full Test-S results on Antisocial task. \textbf{SoTA:} Previous SoTA performance on each respective dataset. \textbf{Underscore} indicates that we have different data splits to the SoTA model. Best model of each dataset is in \textbf{bold}. }\label{tab:anti-social_res}
\end{table*}

%% file: table_fig/emotion_res.tex
\begin{table*}[]
\centering 
\tiny
\setlength\tabcolsep{2pt}
\resizebox{\textwidth}{!}{
\begin{tabular}{@{}llc|cccc|ccccc:ccc:ccc|cc|cl@{}}
 \toprule
\textbf{Dataset}     & \textbf{Metric}   & \textbf{Random}   & \textbf{mBERT}    & \textbf{XLM-R}    & \textbf{Bernice}  & \textbf{InfoDCL}  & \textbf{BLOOM}    & \textbf{BLOOMZ}   & \multicolumn{1}{c}{\textbf{\begin{tabular}[c]{@{}c@{}}BLOOMZ\\ (MT)\end{tabular}}}  & \multicolumn{1}{c}{\textbf{\begin{tabular}[c]{@{}c@{}}BLOOMZ-\\ P3\end{tabular}}}    & \multicolumn{1}{c}{\textbf{\begin{tabular}[c]{@{}c@{}}BLOOMZ-\\ Bactrian\end{tabular}}}  & \textbf{mT5}  & \textbf{mT0}  & \multicolumn{1}{c}{\textbf{\begin{tabular}[c]{@{}c@{}}mT0\\ (MT)\end{tabular}}}     & \textbf{LLaMA}    & \textbf{Alpaca}   & \textbf{Vicuna}   & \textbf{ChatGPT}  & \multicolumn{1}{c}{\textbf{\begin{tabular}[c]{@{}c@{}}ChatGPT\\ (MT)\end{tabular}}}     & \textbf{SoTA}     & \textbf{SoTA study}\\ \midrule
Emot-eng\textsubscript{Wal}      & M-F1      & 12.00     & 65.35     & 69.24     & 69.51     & \textbf{70.10}    & 16.86     & 21.13     & 21.13     & 11.05     & 24.58     & 4.54      & 27.01     & 27.01     & 47.94     & 60.08     & 44.88     & 68.06     & 68.06     & \underline{57.00}     & \citet{suresh2021not}    \\ 
Emot-zho\textsubscript{Lee}      & Acc.      & 16.27     & 65.87     & 70.81     & 67.30     & \textbf{72.97}    & 14.11     & 52.63     & 56.94     & 51.44     & 18.42     & 23.92     & 55.50     & 57.42     & 16.03     & 49.76     & 19.14     & 68.18     & 65.55     & 53.90     & \citet{wang-2015-emotion}    \\ 
Emot-fin\textsubscript{Kaj}      & M-F1      & 13.32     & 46.15     & 58.61     & 49.54     & \textbf{58.87}    & 4.88      & 3.45      & 3.53      & 3.45      & 8.46      & 3.61      & 18.70     & 8.89      & 3.30      & 3.86      & 5.45      & 52.79     & 46.58     & -     & ---  \\ 
Emot-fra\textsubscript{Kaj}      & M-F1      & 12.85     & 49.99     & 58.23     & \textbf{60.10}    & 60.04     & 4.24      & 12.51     & 12.04     & 4.39      & 8.45      & 3.75      & 20.70     & 22.32     & 8.82      & 17.40     & 14.23     & 58.29     & 45.88     & -     & ---  \\ 
Emot-ita\textsubscript{Kaj}      & M-F1      & 13.28     & 54.61     & 59.35     & 60.68     & \textbf{60.84}    & 6.21      & 8.78      & 4.34      & 5.87      & 12.75     & 5.62      & 20.58     & 14.28     & 8.60      & 14.08     & 11.84     & 60.53     & 38.16     & -     & ---  \\ 
Emot-ara\textsubscript{Abd}      & M-F1      & 13.16     & 57.77     & 59.70     & \textbf{66.24}    & 65.69     & 5.08      & 9.42      & 10.93     & 6.65      & 8.92      & 5.17      & 13.98     & 5.32      & 3.19      & 5.33      & 9.49      & 32.09     & 27.15     & \underline{60.32}     & \citet{mageed-2020-aranet}   \\ 
Emot-eng\textsubscript{Moh}      & M-F1      & 24.41     & 70.90     & 78.30     & 80.86     & \textbf{81.37}    & 16.48     & 9.63      & 9.63      & 29.70     & 31.00     & 15.51     & 47.57     & 47.57     & 15.77     & 59.71     & 25.15     & 74.44     & 74.44     & \underline{78.50}     & \citet{barbieri-2020-tweeteval}  \\ 
Emot-ara\textsubscript{Moh}      & M-F1      & 22.98     & 72.93     & 83.19     & \textbf{84.10}    & 83.29     & 19.12     & 23.15     & 34.83     & 21.07     & 22.68     & 12.70     & 35.32     & 8.75      & 26.18     & 41.16     & 25.85     & 80.57     & 71.69     & \underline{91.00}     & \citet{bianchi-2022-xlm} \\ 
Emot-spa\textsubscript{Moh}      & M-F1      & 25.58     & 78.53     & 82.27     & 83.33     & \textbf{85.49}    & 15.56     & 25.14     & 8.30      & 19.50     & 33.17     & 12.53     & 42.72     & 23.43     & 25.37     & 54.06     & 29.18     & 75.89     & 74.64     & \underline{91.00}     & \citet{bianchi-2022-xlm} \\ 
Emot-ind\textsubscript{Sap}      & M-F1      & 19.69     & 63.58     & 76.93     & 78.37     & \textbf{81.27}    & 8.85      & 36.78     & 7.69      & 35.53     & 15.55     & 6.73      & 29.03     & 34.63     & 26.94     & 36.37     & 24.04     & 77.51     & 60.20     & \underline{68.00}     & \citet{emotion-2018-saputri} \\ 
Emot-tur\textsubscript{Guv}      & Acc.      & 18.00     & 98.25     & 98.50     & 98.92     & \textbf{99.33}    & 22.50     & 20.25     & 22.00     & 20.25     & 19.50     & 21.75     & 51.75     & 49.00     & 22.75     & 36.50     & 19.75     & 89.50     & 84.00     & 87.00     & \citet{guven2020comparison}  \\ 
Emot-ind\textsubscript{Wil}      & M-F1      & 18.34     & 63.10     & 73.39     & \textbf{76.12}    & 75.18     & 8.00      & 33.62     & 8.00      & 33.93     & 16.30     & 5.47      & 30.50     & 30.10     & 29.79     & 36.78     & 25.26     & 74.87     & 58.25     & \underline{79.47}     & \citet{wong-2020-indonlu}    \\ 
Emot-vie\textsubscript{Ho}   & W-F1      & 16.12     & 54.63     & 63.50     & 63.12     & \textbf{64.58}    & 2.19      & 14.90     & 5.52      & 12.22     & 10.20     & 1.76      & 27.81     & 24.03     & 2.04      & 16.34     & 9.52      & 54.69     & 32.96     & \underline{66.34}     & \citet{ho2019emotion}    \\ 
Emot-eng\textsubscript{Pla}      & M-F1      & 11.23     & 39.09     & 46.49     & \textbf{48.51}    & 47.45     & 5.30      & 9.74      & 9.74      & 7.80      & 10.12     & 2.93      & 19.29     & 19.29     & 7.96      & 25.59     & 20.60     & 40.73     & 40.73     & \underline{32.00}     & \citet{plaza-del-arco-etal-2020-emoevent}    \\ 
Emot-spa\textsubscript{Pla}      & M-F1      & 11.47     & 50.89     & 52.28     & \textbf{57.95}    & 54.25     & 5.64      & 12.63     & 10.31     & 7.77      & 3.52      & 3.36      & 13.54     & 13.45     & 9.41      & 27.48     & 22.31     & 40.51     & 44.00     & \underline{44.00}     & \citet{plaza-del-arco-etal-2020-emoevent}    \\ 
Emot-fin\textsubscript{Ohm}      & M-F1      & 13.60     & 41.74     & 49.82     & 45.07     & 50.36     & 5.96      & 3.78      & 2.39      & 2.84      & 7.67      & 4.39      & 14.47     & 7.32      & 3.73      & 3.73      & 6.04      & \textbf{52.94}    & 42.55     & -     & ---  \\ 
Emot-eng\textsubscript{Dem}      & M-F1      & 3.18      & 53.75     & 52.58     & 57.50     & \textbf{58.83}    & 1.96      & 5.40      & 5.40      & 1.90      & 3.14      & 0.31      & 11.36     & 11.36     & 4.03      & 15.91     & 10.65     & 32.61     & 32.61     & \underline{64.80}     & \citet{suresh2021not}    \\ 
Emot-ita\textsubscript{Bia}      & M-F1      & 22.54     & 66.47     & 67.76     & \textbf{76.78}    & 75.58     & 16.21     & 27.07     & 15.42     & 21.44     & 22.65     & 16.80     & 38.69     & 13.15     & 27.47     & 50.42     & 24.10     & 73.66     & 73.19     & 71.00     & \citet{bianchi2021feel}  \\ 
Emot-ron\textsubscript{Cio}      & M-F1      & 23.99     & 87.63     & 91.24     & 89.26     & \textbf{91.43}    & 11.37     & 19.59     & 9.90      & 14.75     & 30.85     & 10.25     & 41.07     & 33.29     & 26.95     & 52.76     & 30.81     & 84.74     & 60.97     & 78.00     & \citet{red-2021-ciobotaru}   \\ 
Emot-hin\textsubscript{Deb}      & M-F1      & 3.10      & 45.42     & 50.17     & 48.52     & \textbf{52.46}    & 1.17      & 4.70      & 3.03      & 4.51      & 3.67      & 0.37      & 9.07      & 12.67     & 2.44      & 6.70      & 2.88      & 31.31     & 19.91     & -     & ---  \\ 
Emot-por\textsubscript{Cor}      & M-F1      & 3.44      & 72.09     & 73.93     & \textbf{76.70}    & 75.16     & 0.68      & 2.62      & 3.58      & 1.12      & 2.99      & 0.18      & 2.55      & 0.65      & 1.05      & 6.70      & 2.54      & 8.72      & 8.98      & \underline{64.00}     & \citet{weak-2021-diogo}  \\ 
Emot-fas\textsubscript{Sab}      & M-F1      & 20.27     & 21.49     & 21.32     & 21.15     & 26.21     & 5.23      & 11.15     & 3.35      & 12.00     & 5.24      & 6.59      & 14.03     & 7.79      & 5.13      & 12.26     & 4.47      & \textbf{28.59}    & 27.80     & -     & ---  \\ 
Emot-rus\textsubscript{Sbo}      & M-F1      & 15.84     & 76.49     & 83.17     & 81.96     & \textbf{83.75}    & 8.61      & 15.05     & 12.36     & 11.02     & 11.74     & 3.57      & 30.27     & 21.97     & 8.30      & 51.06     & 8.59      & 79.62     & 76.53     & \underline{78.00}     & \citet{sboev2021data}    \\ 
Emot-ben\textsubscript{Iqb}      & M-F1      & 15.97     & 53.79     & 59.57     & \textbf{63.86}    & 63.69     & 10.43     & 13.37     & 15.01     & 11.39     & 9.36      & 5.19      & 17.90     & 30.11     & 2.98      & 9.52      & 2.98      & 53.23     & 44.21     & -     & ---  \\ 
Emot-fra\textsubscript{Bia}      & M-F1      & 20.09     & 75.07     & 79.98     & \textbf{84.15}    & 83.56     & 20.36     & 29.15     & 50.61     & 20.00     & 37.94     & 12.18     & 47.99     & 54.75     & 30.42     & 63.48     & 39.38     & 79.02     & 78.19     & \underline{88.00}     & \citet{bianchi-2022-xlm} \\ 
Emot-deu\textsubscript{Bia}      & M-F1      & 21.56     & 71.35     & 78.21     & \textbf{81.75}    & 79.18     & 15.48     & 21.00     & 14.22     & 20.26     & 15.96     & 12.25     & 43.21     & 51.05     & 26.96     & 69.86     & 32.10     & 76.07     & 24.99     & \underline{87.00}     & \citet{bianchi-2022-xlm} \\ \midrule
Average      & ---   & 15.86     & 61.42     & 66.87     & 68.13     & \textbf{69.27}     & 9.71      & 17.18     & 13.85     & 15.07     & 15.19     & 7.75      & 27.87     & 24.21     & 15.14     & 31.80     & 18.12     & 59.58     & 50.85     & ---   & ---  \\ 
\bottomrule 
\end{tabular}
}
\caption{Full Test-S results on emotion recognition. \textbf{SoTA:} Previous SoTA performance on each respective dataset. \textbf{Underscore} indicates that we have different data splits to the SoTA model. Best model of each dataset is in \textbf{bold}. }\label{tab:emotion_res}
\end{table*}

%% file: table_fig/humor_res.tex
\begin{table*}[]
\centering 
\tiny
\setlength\tabcolsep{2pt}
\resizebox{\textwidth}{!}{
\begin{tabular}{@{}llc|cccc|ccccc:ccc:ccc|cc|cl@{}}
 \toprule
\textbf{Dataset}     & \textbf{Metric}   & \textbf{Random}   & \textbf{mBERT}    & \textbf{XLM-R}    & \textbf{Bernice}  & \textbf{InfoDCL}  & \textbf{BLOOM}    & \textbf{BLOOMZ}   & \multicolumn{1}{c}{\textbf{\begin{tabular}[c]{@{}c@{}}BLOOMZ\\ (MT)\end{tabular}}}  & \multicolumn{1}{c}{\textbf{\begin{tabular}[c]{@{}c@{}}BLOOMZ-\\ P3\end{tabular}}}    & \multicolumn{1}{c}{\textbf{\begin{tabular}[c]{@{}c@{}}BLOOMZ-\\ Bactrian\end{tabular}}}  & \textbf{mT5}  & \textbf{mT0}  & \multicolumn{1}{c}{\textbf{\begin{tabular}[c]{@{}c@{}}mT0\\ (MT)\end{tabular}}}     & \textbf{LLaMA}    & \textbf{Alpaca}   & \textbf{Vicuna}   & \textbf{ChatGPT}  & \multicolumn{1}{c}{\textbf{\begin{tabular}[c]{@{}c@{}}ChatGPT\\ (MT)\end{tabular}}}     & \textbf{SoTA}     & \textbf{SoTA study}\\ \midrule
Humo-hin\textsubscript{Agg}      & Acc.      & 51.40     & 78.27     & 78.87     & 78.27     & \textbf{80.40}    & 58.60     & 39.60     & 39.80     & 39.60     & 40.80     & 52.20     & 39.60     & 42.40     & 56.80     & 60.40     & 37.20     & 56.00     & 61.00     & \underline{69.30}     & \citet{aggarwal2020sarcasm}  \\ 
Humo-rus\textsubscript{Bli}      & M-F1      & 47.38     & 86.47     & 87.60     & 88.13     & \textbf{88.80}    & 34.12     & 34.12     & 34.12     & 34.12     & 36.38     & 45.09     & 34.12     & 34.12     & 35.36     & 32.52     & 54.60     & 72.75     & 71.44     & \underline{89.00}     & \citet{blinov-2019-large}    \\ 
Humo-spa\textsubscript{Chi}      & M-F1      & 54.58     & 85.46     & 84.45     & 87.20     & \textbf{87.66}    & 48.25     & 32.07     & 32.07     & 31.97     & 39.39     & 34.28     & 32.07     & 32.07     & 38.56     & 34.55     & 50.10     & 68.28     & 68.80     & \underline{88.50}     & \citet{chiruzzo2021overview} \\ 
Humo-eng\textsubscript{Mea}      & M-F1      & 45.25     & 87.19     & 89.86     & \textbf{93.41}    & 91.34     & 26.16     & 26.69     & 26.69     & 26.47     & 27.06     & 42.83     & 26.69     & 26.69     & 28.39     & 39.39     & 42.86     & 89.56     & 89.56     & \underline{98.54}     & \citet{meaney2021hahackathon}    \\ \midrule
Average      & ---   & 49.65     & 84.35     & 85.19     & 86.75     & \textbf{87.05}     & 41.78     & 33.12     & 33.17     & 33.04     & 35.91     & 43.60     & 33.12     & 33.82     & 39.78     & 41.72     & 46.19     & 71.65     & 72.70     & ---   & ---  \\ 
\bottomrule 
\end{tabular}
}
\caption{Full Test-S results on humor detection. \textbf{SoTA:} Previous SoTA performance on each respective dataset. \textbf{Underscore} indicates that we have different data splits to the SoTA model.}\label{tab:humor_res}
\end{table*}

%% file: table_fig/irony_res.tex
\begin{table*}[]
\centering 
\tiny
\setlength\tabcolsep{2pt}
\resizebox{\textwidth}{!}{
\begin{tabular}{@{}llc|cccc|ccccc:ccc:ccc|cc|cl@{}}
 \toprule
\textbf{Dataset}     & \textbf{Metric}   & \textbf{Random}   & \textbf{mBERT}    & \textbf{XLM-R}    & \textbf{Bernice}  & \textbf{InfoDCL}  & \textbf{BLOOM}    & \textbf{BLOOMZ}   & \multicolumn{1}{c}{\textbf{\begin{tabular}[c]{@{}c@{}}BLOOMZ\\ (MT)\end{tabular}}}  & \multicolumn{1}{c}{\textbf{\begin{tabular}[c]{@{}c@{}}BLOOMZ-\\ P3\end{tabular}}}    & \multicolumn{1}{c}{\textbf{\begin{tabular}[c]{@{}c@{}}BLOOMZ-\\ Bactrian\end{tabular}}}  & \textbf{mT5}  & \textbf{mT0}  & \multicolumn{1}{c}{\textbf{\begin{tabular}[c]{@{}c@{}}mT0\\ (MT)\end{tabular}}}     & \textbf{LLaMA}    & \textbf{Alpaca}   & \textbf{Vicuna}   & \textbf{ChatGPT}  & \multicolumn{1}{c}{\textbf{\begin{tabular}[c]{@{}c@{}}ChatGPT\\ (MT)\end{tabular}}}     & \textbf{SoTA}     & \textbf{SoTA study}\\ \midrule
Iron-ita\textsubscript{Bas}      & M-F1      & 41.87     & 63.38     & 61.54     & 63.16     & \textbf{65.67}    & 46.92     & 46.92     & 50.38     & 50.97     & 46.92     & 49.59     & 46.92     & 46.92     & 49.02     & 13.27     & 55.49     & 59.91     & 55.16     & \underline{59.59}     & \citet{basile2014overview}   \\ 
Iron-spa\textsubscript{Bar}      & M-F1      & 40.70     & 57.90     & 58.01     & 61.66     & \textbf{67.52}    & 46.47     & 46.47     & 48.65     & 56.77     & 46.47     & 48.89     & 46.47     & 46.47     & 49.34     & 13.88     & 52.78     & 66.10     & 63.97     & \underline{54.12}     & \citet{barbieri2016overview} \\ 
Iron-eng\textsubscript{Hee}      & F1-iro.   & 45.00     & 59.99     & 63.43     & 67.43     & \textbf{68.25}    & 0.00      & 0.00      & 0.00      & 25.41     & 0.00      & 1.03      & 0.00      & 0.00      & 5.29      & 55.06     & 41.19     & 59.00     & 59.00     & \underline{70.50}     & \citet{van-hee2018semeval}   \\ 
Iron-ita\textsubscript{Cig}      & M-F1      & 51.80     & 70.37     & 72.66     & \textbf{77.44}    & 75.92     & 34.67     & 34.21     & 45.82     & 48.21     & 34.21     & 38.78     & 34.21     & 34.21     & 48.22     & 33.24     & 56.22     & 73.32     & 74.20     & \underline{73.10}     & \citet{cignarella2018overview}   \\ 
Iron-hin\textsubscript{Vij}      & M-F1      & 46.88     & 70.90     & \textbf{73.25}    & 72.90     & 71.16     & 56.66     & 43.99     & 57.75     & 51.30     & 55.80     & 46.28     & 42.53     & 46.63     & 52.02     & 23.61     & 57.73     & 52.89     & 57.92     & \underline{77.00}     & \citet{vijay-2018-dataset-irony} \\ 
Iron-ara\textsubscript{Gha}      & M-F1      & 48.39     & 82.19     & \textbf{83.95}    & 82.45     & 83.08     & 40.10     & 34.24     & 32.61     & 51.43     & 32.61     & 33.80     & 32.61     & 32.61     & 39.17     & 34.40     & 35.94     & 68.78     & 67.40     & \underline{84.40}     & \citet{mageed-2020-aranet}   \\ 
Iron-spa\textsubscript{Ort}      & M-F1      & 46.84     & 67.42     & 71.18     & 72.79     & \textbf{73.88}    & 39.47     & 39.47     & 53.40     & 54.46     & 39.47     & 40.22     & 39.47     & 39.47     & 60.64     & 29.47     & 56.77     & 62.92     & 61.69     & \underline{71.67}     & \citet{ortega2019overview}   \\ 
Iron-fas\textsubscript{Gol}      & Acc.      & 48.30     & 74.04     & 74.60     & 73.58     & \textbf{76.53}    & 51.70     & 56.46     & 56.46     & 51.70     & 56.46     & 56.46     & 56.46     & 56.46     & 49.66     & 43.88     & 57.82     & 62.59     & 56.80     & \underline{83.10}     & \citet{golazizian2020mirasirony} \\ 
Iron-zho\textsubscript{Xia}      & M-F1      & 11.71     & 31.96     & 31.19     & 30.52     & \textbf{33.36}    & 13.65     & 14.56     & 3.09      & 9.89      & 13.65     & 13.65     & 13.55     & 3.17      & 13.61     & 0.63      & 13.40     & 18.58     & 10.06     & \underline{57.20}     & \citet{xiang2020ciron}   \\ 
Sarc-eng\textsubscript{Wal}      & M-F1      & 53.33     & 63.32     & 65.36     & 67.45     & 63.65     & 45.21     & 32.80     & 32.80     & 33.23     & 41.54     & 45.01     & 32.80     & 32.80     & 34.55     & 50.47     & 33.38     & \textbf{70.79}    & 70.79     & \underline{69.00}     & \citet{felbo2017using}   \\ 
Sarc-eng\textsubscript{Ril}      & F1-sar.   & 27.00     & 46.76     & 52.50     & 54.89     & \textbf{57.46}    & 10.53     & 0.00      & 0.00      & 0.00      & 0.00      & 36.36     & 0.00      & 0.00      & 0.00      & 38.97     & 14.81     & 50.00     & 50.00     & \underline{51.00}     & \citet{riloff2013sarcasm}    \\ 
Sarc-ces\textsubscript{Pta}      & M-F1      & 35.92     & 66.12     & 60.17     & 65.97     & \textbf{67.89}    & 46.90     & 49.03     & 3.68      & 49.03     & 49.75     & 33.94     & 49.03     & 3.68      & 34.46     & 6.22      & 55.00     & 51.20     & 52.48     & \underline{58.20}     & \citet{ptavcek2014sarcasm}   \\ 
Sarc-eng\textsubscript{Pta}      & M-F1      & 49.85     & 94.28     & \textbf{95.76}    & 94.99     & 95.56     & 41.77     & 38.42     & 38.42     & 39.49     & 37.05     & 45.25     & 38.42     & 38.42     & 40.45     & 33.01     & 48.17     & 74.30     & 74.30     & \underline{92.37}     & \citet{ptavcek2014sarcasm}   \\ 
Sarc-eng\textsubscript{Bam}      & Acc.      & 52.20     & 79.73     & 80.40     & \textbf{82.40}    & 82.27     & 48.60     & 52.00     & 52.00     & 51.80     & 50.60     & 49.00     & 52.00     & 52.00     & 49.20     & 52.00     & 52.20     & 64.60     & 64.60     & \underline{85.10}     & \citet{bamman2015contextualized} \\ 
Sarc-eng\textsubscript{Raj}      & Acc.      & 48.80     & 94.20     & 95.33     & \textbf{96.27}    & 95.67     & 77.60     & 91.20     & 91.20     & 90.80     & 87.80     & 19.60     & 91.20     & 91.20     & 87.00     & 15.00     & 83.60     & 74.60     & 74.60     & \underline{92.94}     & \citet{rajadesingan2015sarcasm}  \\ 
Sarc-eng\textsubscript{Ora}      & M-F1      & 49.00     & 72.73     & \textbf{75.87}    & 74.69     & 75.64     & 41.48     & 32.89     & 32.89     & 32.80     & 42.11     & 41.83     & 32.89     & 32.89     & 33.71     & 47.08     & 36.08     & 73.78     & 73.78     & \underline{75.00}     & \citet{felbo2017using}   \\ 
Sarc-zho\textsubscript{Gon}      & M-F1      & 48.25     & 71.01     & 70.63     & \textbf{72.75}    & 70.53     & 49.04     & 33.29     & 33.20     & 33.20     & 56.10     & 38.89     & 33.29     & 33.29     & 36.25     & 41.41     & 49.36     & 53.50     & 51.05     & \underline{73.68}     & \citet{gong-2020-design} \\ 
Sarc-ara\textsubscript{Abu}      & M-F1      & 44.26     & 69.06     & 69.57     & 69.57     & 71.87     & 27.16     & 44.57     & 16.39     & 44.57     & 45.15     & 20.66     & 44.57     & 16.39     & 53.38     & 17.21     & 53.74     & 74.38     & \textbf{75.47}    & \underline{76.30}     & \citet{arabert-2021-abdulmageed} \\ 
Sarc-ara\textsubscript{Far}      & M-F1      & 46.22     & 66.87     & 68.45     & 68.80     & \textbf{68.81}    & 41.67     & 42.00     & 21.63     & 41.93     & 53.31     & 30.34     & 42.00     & 21.63     & 42.65     & 23.47     & 50.37     & 66.24     & 68.43     & \underline{73.10}     & \citet{abufarha-etal-2021-arsarcasm-v2}  \\ 
Iron-T-eng\textsubscript{Hee}    & M-F1      & 22.36     & 47.35     & 46.43     & 56.04     & \textbf{57.58}    & 18.83     & 18.83     & 18.83     & 18.83     & 18.83     & 18.83     & 18.83     & 18.83     & 18.83     & 18.83     & 18.83     & 30.81     & 30.81     & \underline{50.70}     & \citet{van-hee2018semeval}   \\ \midrule
Average      & ---   & 42.93     & 67.48     & 68.51     & 70.29     & \textbf{71.12}     & 38.92     & 37.57     & 34.46     & 41.79     & 40.39     & 35.42     & 37.36     & 32.35     & 39.87     & 29.56     & 46.14     & 60.41     & 59.63     & ---   & ---  \\ 
\bottomrule 
\end{tabular}
}
\caption{Full Test-S results on irony \& sarcasm detection. \textbf{SoTA:} Previous SoTA performance on each respective dataset. \textbf{Underscore} indicates that we have different data splits to the SoTA model. Best model of each dataset is in \textbf{bold}. }\label{tab:irony_res}
\end{table*}

%% file: table_fig/sentiment_res.tex
\begin{table*}[]
\centering 
\tiny
\setlength\tabcolsep{2pt}
\resizebox{\textwidth}{!}{
\begin{tabular}{@{}llc|cccc|ccccc:ccc:ccc|cc|cl@{}}
 \toprule
\textbf{Dataset}     & \textbf{Metric}   & \textbf{Random}   & \textbf{mBERT}    & \textbf{XLM-R}    & \textbf{Bernice}  & \textbf{InfoDCL}  & \textbf{BLOOM}    & \textbf{BLOOMZ}   & \multicolumn{1}{c}{\textbf{\begin{tabular}[c]{@{}c@{}}BLOOMZ\\ (MT)\end{tabular}}}  & \multicolumn{1}{c}{\textbf{\begin{tabular}[c]{@{}c@{}}BLOOMZ-\\ P3\end{tabular}}}    & \multicolumn{1}{c}{\textbf{\begin{tabular}[c]{@{}c@{}}BLOOMZ-\\ Bactrian\end{tabular}}}  & \textbf{mT5}  & \textbf{mT0}  & \multicolumn{1}{c}{\textbf{\begin{tabular}[c]{@{}c@{}}mT0\\ (MT)\end{tabular}}}     & \textbf{LLaMA}    & \textbf{Alpaca}   & \textbf{Vicuna}   & \textbf{ChatGPT}  & \multicolumn{1}{c}{\textbf{\begin{tabular}[c]{@{}c@{}}ChatGPT\\ (MT)\end{tabular}}}     & \textbf{SoTA}     & \textbf{SoTA study}\\ \midrule
Sent-eng\textsubscript{Pan}      & Acc.      & 51.80     & 81.60     & 86.07     & 85.20     & 86.93     & 51.60     & \textbf{97.20}    & 97.20     & 96.80     & 56.40     & 51.40     & 76.80     & 76.80     & 61.80     & 60.20     & 55.40     & 88.40     & 88.40     & \underline{90.82}     & \citet{ke-2020-sentilare}    \\ 
Sent-zho\textsubscript{Tan}      & M-F1      & 49.04     & 95.72     & 95.85     & \textbf{96.85}    & 95.92     & 36.40     & 90.52     & 87.19     & 87.73     & 58.32     & 31.88     & 86.57     & 90.14     & 56.83     & 49.59     & 45.31     & 90.16     & 88.39     & \underline{95.80}     & \citet{sun-2020-ernie2}  \\ 
Sent-T-eng\textsubscript{The}    & Acc.      & 47.00     & 79.20     & 88.73     & \textbf{91.40}    & 89.60     & 47.20     & 75.60     & 75.60     & 78.40     & 45.20     & 41.20     & 73.80     & 73.80     & 43.20     & 54.00     & 55.60     & 90.00     & 90.00     & \underline{88.00}     & \citet{felbo2017using}   \\ 
Sent-Y-eng\textsubscript{The}    & Acc.      & 50.20     & 86.00     & 90.80     & 92.87     & \textbf{93.33}    & 42.40     & 84.80     & 84.80     & 85.20     & 36.80     & 30.60     & 81.60     & 81.60     & 32.60     & 47.40     & 33.20     & 90.00     & 90.00     & \underline{93.00}     & \citet{felbo2017using}   \\ 
Sent-5-eng\textsubscript{Soc}    & Acc.      & 19.80     & 49.73     & \textbf{53.93}    & 51.73     & 53.67     & 24.80     & 48.60     & 48.60     & 48.40     & 19.80     & 27.60     & 45.60     & 45.60     & 20.00     & 40.40     & 19.60     & 49.60     & 49.60     & \underline{58.59}     & \citet{ke-2020-sentilare}    \\ 
Sent-kor\textsubscript{Jan}      & M-F1      & 20.79     & 38.64     & 42.56     & 41.05     & \textbf{44.70}    & 2.96      & 29.03     & 3.70      & 28.08     & 1.11      & 13.02     & 25.91     & 19.91     & 12.78     & 25.37     & 8.01      & 42.32     & 36.49     & -     & ---  \\ 
Sent-eng\textsubscript{Soc}      & Acc.      & 46.80     & 84.47     & 88.40     & 86.87     & 88.73     & 56.40     & 92.20     & 92.20     & \textbf{93.00}    & 53.60     & 49.00     & 76.80     & 76.80     & 53.40     & 58.20     & 49.80     & 91.80     & 91.80     & \underline{96.70}     & \citet{tian-2020-skep}   \\ 
Sent-ita\textsubscript{Bas}      & M-F1      & 46.79     & 78.06     & 85.22     & 85.33     & \textbf{88.77}    & 40.60     & 54.18     & 39.08     & 74.32     & 41.25     & 38.50     & 53.07     & 38.50     & 54.20     & 51.58     & 60.85     & 87.26     & 87.01     & \underline{67.71}     & \citet{basile2014overview}   \\ 
Sent-ita\textsubscript{Bas}      & M-F1      & 49.61     & 66.68     & 78.51     & 79.93     & 84.73     & 46.61     & 45.37     & 42.04     & 63.71     & 42.63     & 40.76     & 44.99     & 40.76     & 57.97     & 52.71     & 67.62     & 83.09     & \textbf{85.21}    & \underline{66.38}     & \citet{barbieri2016overview} \\ 
Sent-mlt\textsubscript{Din}      & M-F1      & 47.51     & 63.14     & 39.79     & 65.96     & 68.01     & 39.36     & 36.75     & 39.36     & 47.70     & 41.14     & 39.36     & 39.25     & 25.97     & 53.99     & 39.36     & 48.98     & \textbf{78.47}    & 77.45     & \underline{54.70}     & \citet{dingli-2016-sentiment}    \\ 
Sent-bul\textsubscript{Moz}      & M-F1      & 32.48     & 62.01     & 64.22     & 63.09     & \textbf{65.09}    & 22.05     & 22.61     & 15.15     & 27.35     & 29.54     & 11.75     & 19.37     & 29.02     & 13.24     & 16.88     & 23.69     & 59.32     & 54.39     & \underline{52.00}     & \citet{mozetivc2016multilingual} \\ 
Sent-bos\textsubscript{Moz}      & M-F1      & 34.37     & 64.75     & 68.12     & 65.83     & \textbf{68.31}    & 24.58     & 26.39     & 16.31     & 31.80     & 24.05     & 16.31     & 21.16     & 50.69     & 27.92     & 19.32     & 16.72     & 63.22     & 48.15     & \underline{60.60}     & \citet{mozetivc2016multilingual} \\ 
Sent-deu\textsubscript{Moz}      & M-F1      & 32.60     & 61.63     & 61.54     & \textbf{62.80}    & 62.53     & 17.36     & 26.96     & 9.39      & 26.13     & 20.10     & 9.79      & 20.54     & 21.93     & 22.50     & 15.88     & 25.51     & 49.70     & 47.64     & \underline{53.60}     & \citet{mozetivc2016multilingual} \\ 
Sent-eng\textsubscript{Moz}      & M-F1      & 30.33     & 62.81     & 68.53     & 68.59     & \textbf{68.78}    & 26.73     & 36.94     & 36.94     & 36.52     & 27.25     & 12.90     & 22.92     & 22.92     & 15.77     & 28.86     & 23.87     & 60.61     & 60.61     & \underline{63.00}     & \citet{mozetivc2016multilingual} \\ 
Sent-spa\textsubscript{Moz}      & M-F1      & 29.19     & 51.39     & 55.80     & 55.79     & \textbf{55.87}    & 25.84     & 31.85     & 10.32     & 33.31     & 24.18     & 7.25      & 26.36     & 7.25      & 19.94     & 14.69     & 20.63     & 39.39     & 42.48     & \underline{38.60}     & \citet{mozetivc2016multilingual} \\ 
Sent-hrv\textsubscript{Moz}      & M-F1      & 30.23     & 64.79     & \textbf{69.44}    & 66.68     & 67.83     & 19.35     & 27.96     & 11.02     & 35.18     & 20.68     & 11.02     & 27.47     & 11.02     & 24.32     & 14.45     & 13.67     & 62.02     & 55.66     & \underline{60.60}     & \citet{mozetivc2016multilingual} \\ 
Sent-hun\textsubscript{Moz}      & M-F1      & 27.80     & 67.27     & \textbf{71.62}    & 70.16     & 68.37     & 15.80     & 28.87     & 8.61      & 30.31     & 19.62     & 6.93      & 27.19     & 39.94     & 12.18     & 8.66      & 16.39     & 53.22     & 43.89     & \underline{64.10}     & \citet{mozetivc2016multilingual} \\ 
Sent-pol\textsubscript{Moz}      & M-F1      & 30.95     & 67.09     & 66.96     & 67.91     & \textbf{68.22}    & 18.69     & 27.58     & 14.36     & 34.78     & 26.65     & 14.34     & 25.57     & 14.91     & 24.98     & 20.83     & 14.91     & 61.84     & 60.58     & \underline{67.70}     & \citet{mozetivc2016multilingual} \\ 
Sent-por\textsubscript{Moz}      & M-F1      & 29.81     & \textbf{57.14}    & 56.21     & 56.37     & 56.71     & 32.20     & 26.22     & 18.15     & 27.65     & 34.78     & 18.22     & 18.03     & 18.22     & 31.48     & 26.59     & 18.94     & 44.60     & 42.76     & \underline{55.30}     & \citet{mozetivc2016multilingual} \\ 
Sent-rus\textsubscript{Moz}      & M-F1      & 32.33     & 75.24     & 78.35     & 78.84     & \textbf{80.37}    & 26.77     & 30.42     & 28.64     & 31.33     & 30.70     & 14.99     & 20.60     & 30.55     & 18.19     & 21.73     & 22.35     & 61.67     & 59.27     & \underline{61.50}     & \citet{mozetivc2016multilingual} \\ 
Sent-slk\textsubscript{Moz}      & M-F1      & 28.57     & 71.78     & 74.78     & 71.82     & \textbf{75.71}    & 20.49     & 27.36     & 20.32     & 31.89     & 18.14     & 14.09     & 27.85     & 23.54     & 25.01     & 17.93     & 10.46     & 57.46     & 56.18     & \underline{68.20}     & \citet{mozetivc2016multilingual} \\ 
Sent-slv\textsubscript{Moz}      & M-F1      & 35.90     & 59.53     & 61.57     & 61.27     & \textbf{61.96}    & 26.34     & 19.79     & 14.66     & 26.76     & 22.95     & 14.66     & 20.75     & 14.66     & 27.09     & 20.27     & 20.69     & 58.64     & 57.26     & \underline{55.30}     & \citet{mozetivc2016multilingual} \\ 
Sent-sqi\textsubscript{Moz}      & M-F1      & 29.39     & 43.07     & 46.42     & 45.69     & \textbf{47.04}    & 18.25     & 26.88     & 10.76     & 31.78     & 23.15     & 9.98      & 26.59     & 41.09     & 27.51     & 9.98      & 16.11     & 46.82     & 33.84     & \underline{39.10}     & \citet{mozetivc2016multilingual} \\ 
Sent-srp\textsubscript{Moz}      & M-F1      & 34.09     & 53.16     & 56.62     & 52.51     & \textbf{56.85}    & 28.06     & 20.66     & 19.82     & 27.45     & 24.20     & 16.99     & 20.55     & 44.27     & 31.95     & 22.73     & 19.05     & 55.84     & 52.54     & \underline{60.60}     & \citet{mozetivc2016multilingual} \\ 
Sent-swe\textsubscript{Moz}      & M-F1      & 33.35     & 64.61     & 68.84     & 69.93     & \textbf{70.98}    & 24.85     & 25.84     & 19.32     & 29.31     & 28.14     & 19.66     & 20.77     & 26.30     & 24.57     & 22.56     & 16.05     & 60.10     & 62.58     & \underline{65.70}     & \citet{mozetivc2016multilingual} \\ 
Sent-deu\textsubscript{Rei}      & M-F1      & 19.05     & 47.05     & 52.99     & 55.93     & \textbf{60.94}    & 11.67     & 14.46     & 3.01      & 13.61     & 16.78     & 4.26      & 15.78     & 18.18     & 13.82     & 8.76      & 30.32     & 30.95     & 33.58     & -     & ---  \\ 
Sent-spa\textsubscript{Rei}      & M-F1      & 19.78     & 39.59     & 51.26     & 51.53     & \textbf{52.83}    & 17.86     & 6.70      & 6.60      & 8.12      & 12.91     & 1.69      & 8.49      & 1.69      & 15.72     & 14.71     & 31.50     & 25.01     & 28.54     & -     & ---  \\ 
Sent-ita\textsubscript{Rei}      & M-F1      & 23.76     & 47.29     & 49.57     & 47.14     & \textbf{51.20}    & 27.92     & 10.75     & 3.30      & 12.45     & 25.39     & 3.45      & 11.30     & 3.30      & 14.80     & 10.59     & 31.17     & 25.39     & 30.61     & -     & ---  \\ 
Sent-eng\textsubscript{Ros}      & M-Rec     & 36.77     & 64.52     & 67.50     & \textbf{70.90}    & 70.39     & 33.60     & 52.75     & 52.75     & 57.60     & 32.29     & 33.33     & 38.89     & 38.89     & 33.97     & 38.72     & 39.31     & 69.94     & 69.94     & \underline{72.60}     & \citet{barbieri-2020-tweeteval}  \\ 
Sent-ben\textsubscript{Pat}      & M-F1      & 28.06     & 54.20     & 58.28     & 57.56     & \textbf{59.45}    & 23.88     & 34.29     & 30.34     & 34.73     & 22.12     & 13.67     & 28.91     & 26.36     & 19.53     & 18.53     & 21.01     & 55.00     & 29.14     & \underline{52.60}     & \citet{patra-2018-sentiment} \\ 
Sent-hin\textsubscript{Pat}      & M-F1      & 31.67     & 55.46     & 58.31     & \textbf{62.30}    & 59.76     & 28.73     & 28.74     & 11.59     & 30.45     & 28.23     & 11.57     & 23.96     & 26.66     & 13.80     & 20.58     & 22.45     & 59.40     & 48.30     & \underline{56.90}     & \citet{patra-2018-sentiment} \\ 
Sent-heb\textsubscript{Amr}      & Acc.      & 47.60     & 93.27     & 95.27     & 95.40     & \textbf{95.80}    & 64.80     & 71.00     & 32.20     & 71.20     & 60.80     & 32.20     & 76.60     & 32.20     & 67.00     & 34.80     & 40.80     & 84.20     & 57.40     & \underline{89.06}     & \citet{amram-etal-2018-representations}  \\ 
Sent-por\textsubscript{Bru}      & M-F1      & 35.59     & 59.05     & \textbf{60.00}    & 57.85     & 59.29     & 24.38     & 26.73     & 20.55     & 29.33     & 34.23     & 19.68     & 20.31     & 19.65     & 28.19     & 25.63     & 15.49     & 42.64     & 42.39     & \underline{62.14}     & \citet{brum-2018-building}   \\ 
Sent-fin\textsubscript{Kaj}      & M-F1      & 50.11     & 78.88     & 83.48     & 79.86     & 82.32     & 35.48     & 36.55     & 35.65     & 41.24     & 41.33     & 35.65     & 58.17     & 74.27     & 38.38     & 40.41     & 57.76     & \textbf{83.77}    & 83.32     & -     & ---  \\ 
Sent-fra\textsubscript{Kaj}      & M-F1      & 47.70     & 78.44     & 84.21     & 86.12     & 87.50     & 35.72     & 71.71     & 31.32     & 70.88     & 42.54     & 35.23     & 58.67     & 63.50     & 49.58     & 60.49     & 68.81     & \textbf{87.96}    & 87.78     & -     & ---  \\ 
Sent-ita\textsubscript{Kaj}      & M-F1      & 49.65     & 79.21     & 85.70     & 84.59     & \textbf{86.17}    & 59.43     & 56.28     & 35.98     & 65.24     & 36.48     & 35.98     & 64.76     & 35.98     & 57.65     & 56.67     & 64.59     & 86.03     & 84.69     & -     & ---  \\ 
Sent-nor\textsubscript{Vel}      & M-F1      & 16.24     & 41.05     & \textbf{51.15}    & 39.73     & 42.36     & 0.00      & 18.43     & 4.09      & 19.84     & 0.00      & 1.93      & 35.73     & 39.75     & 0.00      & 11.76     & 9.80      & 42.11     & 41.41     & -     & ---  \\ 
Sent-pol\textsubscript{Koc}      & M-F1      & 25.03     & 70.99     & 76.03     & 77.05     & \textbf{77.63}    & 10.17     & 32.45     & 13.02     & 32.20     & 9.23      & 13.02     & 39.51     & 40.34     & 21.75     & 30.42     & 9.63      & 50.86     & 40.71     & -     & ---  \\ 
Sent-tha\textsubscript{Sur}      & M-F1      & 31.26     & 65.07     & 72.18     & 71.99     & \textbf{75.17}    & 20.65     & 20.51     & 24.37     & 20.16     & 11.04     & 13.33     & 23.94     & 29.83     & 13.29     & 14.21     & 24.40     & 52.81     & 36.32     & -     & ---  \\ 
Sent-zho\textsubscript{Wan}      & M-F1      & 46.66     & 98.80     & \textbf{99.00}    & 98.67     & 98.93     & 46.49     & 69.80     & 49.89     & 70.65     & 61.31     & 33.51     & 56.51     & 81.16     & 53.89     & 45.32     & 65.70     & 77.83     & 79.08     & \underline{91.20}     & \citet{wan-2020-s2ap}    \\ 
Sent-fas\textsubscript{Ash}      & M-F1      & 30.51     & 82.70     & \textbf{84.80}    & 84.56     & 84.54     & 10.25     & 33.77     & 26.77     & 41.34     & 10.25     & 9.98      & 45.69     & 49.56     & 12.76     & 10.25     & 17.63     & 73.87     & 68.91     & \underline{80.00}     & \citet{ashrafi2020mirasopinion}  \\ 
Sent-ron\textsubscript{Dum}      & M-F1      & 52.34     & 84.07     & 88.60     & 84.12     & 88.23     & 37.82     & 65.40     & 30.26     & 83.18     & 44.66     & 30.26     & 93.21     & 30.26     & 62.09     & 32.21     & 30.26     & 93.49     & \textbf{93.72}    & -     & ---  \\ 
Sent-pcm\textsubscript{Oye}      & M-F1      & 31.09     & 65.11     & 66.86     & \textbf{68.87}    & 68.84     & 24.96     & 33.63     & 30.52     & 38.99     & 23.01     & 8.19      & 25.29     & 25.29     & 9.00      & 15.17     & 15.93     & 52.19     & 31.79     & -     & ---  \\ 
Sent-pol\textsubscript{Ryb}      & M-F1      & 21.56     & 45.45     & 53.98     & 51.12     & \textbf{55.11}    & 8.72      & 11.80     & 3.83      & 13.82     & 5.92      & 3.83      & 16.28     & 21.80     & 10.58     & 15.91     & 5.58      & 43.30     & 31.48     & -     & ---  \\ 
Sent-ind\textsubscript{Wil}      & M-F1      & 30.52     & 88.28     & 92.06     & 92.52     & \textbf{92.70}    & 18.36     & 58.25     & 16.42     & 59.11     & 45.24     & 16.19     & 50.79     & 36.23     & 16.16     & 23.89     & 12.78     & 73.49     & 73.98     & \underline{92.72}     & \citet{wong-2020-indonlu}    \\ 
Sent-ara\textsubscript{Abd}      & M-F1      & 29.17     & 71.96     & 75.75     & \textbf{77.36}    & 77.30     & 28.05     & 42.96     & 24.15     & 44.48     & 32.36     & 24.15     & 26.60     & 49.01     & 28.43     & 27.56     & 8.21      & 61.90     & 60.56     & \underline{80.86}     & \citet{elmadany-2022-orca}   \\ 
Sent-bam\textsubscript{Dia}      & M-F1      & 27.61     & 64.28     & 58.46     & 65.46     & \textbf{65.57}    & 19.89     & 31.54     & 8.74      & 36.70     & 19.66     & 8.74      & 24.31     & 8.74      & 15.47     & 9.94      & 14.40     & 40.27     & 37.11     & 72.00     & \citet{diallo-2021-bambara}  \\ 
Sent-ben\textsubscript{Isl}      & M-F1      & 31.18     & 62.95     & 67.99     & 66.74     & \textbf{68.54}    & 18.50     & 42.84     & 29.71     & 45.18     & 29.11     & 18.50     & 30.75     & 31.68     & 18.50     & 18.50     & 15.08     & 55.76     & 29.38     & \underline{64.61}     & \citet{islam-etal-2021-sentnob-dataset}  \\ 
Sent-mar\textsubscript{Kul}      & Acc.      & 29.60     & 84.00     & 86.40     & 84.40     & \textbf{86.47}    & 35.40     & 46.20     & 35.60     & 43.80     & 35.80     & 35.00     & 35.40     & 39.20     & 36.80     & 38.20     & 34.40     & 68.60     & 50.20     & \underline{84.13}     & \citet{kulkarni-2021-l3cubemahasent} \\ 
Sent-kan\textsubscript{Cha}      & M-F1      & 49.55     & 78.92     & 80.55     & 82.35     & \textbf{82.92}    & 27.44     & 55.22     & 22.77     & 61.86     & 23.82     & 22.77     & 47.51     & 51.22     & 29.28     & 23.75     & 24.07     & 68.44     & 51.15     & \underline{68.50}     & \citet{chakravarthi-2022-dravidian}  \\ 
Sent-mal\textsubscript{Cha}      & M-F1      & 50.66     & 84.00     & 83.50     & \textbf{84.03}    & 83.88     & 25.81     & 62.16     & 22.24     & 66.87     & 24.76     & 22.24     & 56.51     & 64.97     & 35.54     & 23.18     & 25.65     & 68.75     & 58.89     & \underline{60.50}     & \citet{chakravarthi-2022-dravidian}  \\ 
Sent-tam\textsubscript{Cha}      & M-F1      & 44.61     & 70.91     & \textbf{75.98}    & 75.90     & 75.87     & 26.73     & 58.06     & 44.93     & 61.29     & 19.37     & 15.40     & 50.64     & 31.28     & 22.52     & 16.59     & 17.85     & 65.98     & 54.79     & \underline{59.00}     & \citet{chakravarthi-2022-dravidian}  \\ 
Sent-ara\textsubscript{Muh}      & Acc.      & 33.20     & 48.73     & 56.40     & 61.20     & \textbf{62.20}    & 34.80     & 51.60     & 34.40     & 52.80     & 37.00     & 34.40     & 41.60     & 53.40     & 36.60     & 37.80     & 25.00     & 58.40     & 58.20     & \underline{75.16}     & \citet{mageed:2022:nadi} \\ 
Sent-amh\textsubscript{Muh}      & W-F1      & 37.95     & 24.93     & 60.32     & 57.21     & \textbf{65.68}    & 54.53     & 27.76     & 22.74     & 16.05     & 49.04     & 54.53     & 22.49     & 2.18      & 54.53     & 54.53     & 2.99      & 20.62     & 46.82     & \underline{78.42}     & \citet{muhammadSemEval2023}  \\ 
Sent-ary\textsubscript{Muh}      & W-F1      & 35.94     & 46.56     & 50.21     & 52.61     & \textbf{53.44}    & 23.29     & 34.49     & 14.17     & 37.40     & 23.98     & 14.17     & 23.41     & 32.35     & 20.11     & 18.02     & 16.64     & 52.19     & 51.66     & \underline{64.83}     & \citet{muhammadSemEval2023}  \\ 
Sent-arq\textsubscript{Muh}      & W-F1      & 34.23     & 57.16     & 65.09     & 70.21     & \textbf{71.25}    & 35.26     & 49.79     & 34.23     & 52.02     & 37.83     & 34.23     & 18.05     & 53.01     & 34.91     & 39.80     & 5.33      & 63.89     & 67.58     & \underline{74.20}     & \citet{muhammadSemEval2023}  \\ 
Sent-hau\textsubscript{Muh}      & W-F1      & 35.22     & 69.94     & \textbf{73.44}    & 73.11     & 72.18     & 24.86     & 20.66     & 21.67     & 30.14     & 33.26     & 16.55     & 20.93     & 16.55     & 19.83     & 16.97     & 17.39     & 55.52     & 34.13     & \underline{82.62}     & \citet{muhammadSemEval2023}  \\ 
Sent-ibo\textsubscript{Muh}      & W-F1      & 32.90     & 76.37     & 76.52     & \textbf{78.36}    & 76.75     & 25.58     & 16.24     & 27.92     & 23.21     & 37.48     & 12.25     & 11.52     & 27.93     & 14.25     & 14.96     & 28.37     & 57.55     & 33.46     & \underline{82.96}     & \citet{muhammadSemEval2023}  \\ 
Sent-pcm\textsubscript{Muh}      & W-F1      & 35.62     & 60.82     & 62.44     & 63.61     & 64.45     & 28.07     & 53.47     & 45.54     & 55.33     & 8.99      & 39.73     & 23.21     & 22.73     & 40.18     & 42.36     & 11.91     & \textbf{70.08}    & 53.71     & \underline{75.96}     & \citet{muhammadSemEval2023}  \\ 
Sent-kin\textsubscript{Muh}      & W-F1      & 35.83     & 55.32     & 57.44     & \textbf{58.78}    & 56.69     & 19.99     & 23.02     & 18.33     & 28.06     & 26.01     & 18.33     & 14.84     & 28.08     & 24.32     & 19.17     & 22.23     & 53.78     & 29.03     & \underline{72.63}     & \citet{muhammadSemEval2023}  \\ 
Sent-swh\textsubscript{Muh}      & W-F1      & 33.72     & 53.04     & 61.15     & 56.19     & \textbf{61.29}    & 5.09      & 17.80     & 13.20     & 17.60     & 8.05      & 1.96      & 14.02     & 13.20     & 9.45      & 2.36      & 45.55     & 54.39     & 53.84     & \underline{65.68}     & \citet{muhammadSemEval2023}  \\ 
Sent-tso\textsubscript{Muh}      & W-F1      & 34.03     & 42.64     & 48.56     & 52.47     & \textbf{52.55}    & 21.65     & 38.43     & 30.48     & 45.54     & 25.42     & 18.54     & 30.74     & 30.82     & 24.58     & 21.05     & 6.50      & 42.58     & 35.70     & 60.67     & \citet{muhammadSemEval2023}  \\ 
Sent-twi\textsubscript{Muh}      & W-F1      & 34.04     & 63.32     & 64.29     & \textbf{65.43}    & 64.51     & 22.32     & 37.26     & 29.31     & 46.56     & 16.91     & 20.74     & 29.34     & 28.99     & 23.18     & 24.48     & 5.28      & 51.12     & 32.06     & \underline{68.28}     & \citet{muhammadSemEval2023}  \\ 
Sent-yor\textsubscript{Muh}      & W-F1      & 36.56     & \textbf{67.87}    & 64.38     & 66.71     & 64.84     & 14.14     & 33.14     & 14.12     & 31.70     & 18.87     & 7.29      & 28.58     & 7.29      & 9.27      & 8.11      & 19.01     & 57.57     & 37.21     & \underline{80.16}     & \citet{muhammad-2022-naijasenti} \\ 
Sent-yor\textsubscript{Sho}      & M-F1      & 46.80     & \textbf{86.86}    & 83.69     & 85.39     & 84.93     & 33.24     & 56.04     & 34.22     & 54.50     & 34.03     & 33.33     & 43.07     & 35.52     & 40.83     & 33.33     & 33.75     & 71.97     & 47.41     & \underline{87.20}     & \citet{shode-2021-africanlp} \\ 
Sent-jpn\textsubscript{Suz}      & Acc.      & 18.00     & 54.33     & 59.67     & 61.13     & \textbf{61.47}    & 32.00     & 45.20     & 26.00     & 44.20     & 33.40     & 25.80     & 47.60     & 14.20     & 32.00     & 19.60     & 26.80     & 55.80     & 44.40     & \underline{61.50}     & \citet{suzuki-etal-2022-japanese}    \\ 
Sent-ace\textsubscript{Win}      & M-F1      & 34.78     & 76.59     & 74.19     & 75.74     & \textbf{77.36}    & 23.31     & 28.22     & 22.89     & 37.89     & 20.40     & 18.44     & 24.73     & 20.99     & 19.39     & 19.38     & 12.79     & 52.63     & 58.05     & 77.40     & \citet{nusax-2020-genta} \\ 
Sent-ban\textsubscript{Win}      & M-F1      & 30.14     & 75.96     & 74.79     & 76.86     & \textbf{79.49}    & 21.96     & 35.00     & 31.84     & 41.90     & 23.12     & 18.27     & 29.91     & 24.44     & 19.80     & 18.44     & 13.82     & 60.91     & 42.28     & 79.50     & \citet{nusax-2020-genta} \\ 
Sent-bbc\textsubscript{Win}      & M-F1      & 30.60     & 72.66     & 65.57     & \textbf{75.17}    & 73.58     & 19.23     & 24.61     & 36.37     & 36.42     & 23.68     & 18.48     & 19.20     & 18.27     & 18.77     & 18.44     & 13.86     & 38.43     & 40.65     & 76.70     & \citet{nusax-2020-genta} \\ 
Sent-bjn\textsubscript{Win}      & M-F1      & 30.77     & 75.81     & 79.82     & 82.58     & \textbf{84.50}    & 20.52     & 41.73     & 43.86     & 50.51     & 34.89     & 18.44     & 27.78     & 18.74     & 18.44     & 22.07     & 14.89     & 69.34     & 75.43     & 86.30     & \citet{nusax-2020-genta} \\ 
Sent-bug\textsubscript{Win}      & M-F1      & 30.77     & \textbf{74.57}    & 67.72     & 73.85     & 71.55     & 19.04     & 20.69     & 31.51     & 34.60     & 20.92     & 18.39     & 18.27     & 18.27     & 19.59     & 18.44     & 12.90     & 34.63     & 30.86     & 77.20     & \citet{nusax-2020-genta} \\ 
Sent-jav\textsubscript{Win}      & M-F1      & 31.62     & 76.08     & 83.44     & 84.38     & \textbf{84.79}    & 20.11     & 35.48     & 18.44     & 48.06     & 23.68     & 18.44     & 37.65     & 33.97     & 18.95     & 18.44     & 15.21     & 73.03     & 78.56     & 85.60     & \citet{nusax-2020-genta} \\ 
Sent-mad\textsubscript{Win}      & M-F1      & 28.64     & 70.56     & 73.36     & 77.36     & \textbf{78.36}    & 22.42     & 31.23     & 37.89     & 45.44     & 23.12     & 18.44     & 21.85     & 18.27     & 18.36     & 18.92     & 13.14     & 61.07     & 61.14     & 77.80     & \citet{nusax-2020-genta} \\ 
Sent-min\textsubscript{Win}      & M-F1      & 34.41     & 77.25     & 80.89     & 80.19     & \textbf{84.07}    & 18.39     & 41.20     & 35.18     & 49.93     & 28.76     & 18.44     & 32.41     & 21.71     & 18.44     & 18.44     & 14.95     & 69.80     & 62.91     & 83.10     & \citet{nusax-2020-genta} \\ 
Sent-nij\textsubscript{Win}      & M-F1      & 34.86     & 73.82     & 73.47     & \textbf{78.19}    & 77.22     & 19.01     & 35.18     & 39.43     & 42.86     & 27.70     & 18.44     & 22.89     & 19.65     & 18.44     & 18.44     & 15.21     & 57.64     & 57.07     & 75.80     & \citet{nusax-2020-genta} \\ 
Sent-sun\textsubscript{Win}      & M-F1      & 32.18     & 75.32     & 76.61     & 77.70     & \textbf{81.71}    & 18.87     & 31.98     & 18.44     & 44.83     & 27.14     & 18.44     & 37.65     & 12.90     & 18.86     & 18.44     & 12.93     & 64.97     & 68.76     & 86.00     & \citet{nusax-2020-genta} \\ 
Sent-tel\textsubscript{Mar}      & M-F1      & 30.27     & 65.20     & 69.44     & 67.14     & \textbf{69.47}    & 9.98      & 38.81     & 23.32     & 32.08     & 13.52     & 9.99      & 20.40     & 36.96     & 9.98      & 9.98      & 23.32     & 56.77     & 43.14     & \underline{62.00}     & \citet{mounika-2022-resource}    \\ \midrule
Average      & ---   & 34.68 & 66.34 & 69.58 & 70.44          & \textbf{71.64}                             & 26.67                          & 39.03 & 28.61                          & 43.03                                                                         & 28.46                                                                         & 20.77                           & 34.65& 32.76                          & 27.55                            & 25.84                            & 25.02                            & 60.34                           & 54.9     & ---   & ---  \\ 
\bottomrule 
\end{tabular}
}
\caption{Full Test-S results on sentiment analysis. \textbf{SoTA:} Previous SoTA performance on each respective dataset. \textbf{Underscore} indicates that we have different data splits to the SoTA model. Best model of each dataset is in \textbf{bold}. }\label{tab:sentiment_res}
\end{table*}

%% file: table_fig/subject_res.tex
\begin{table*}[]
\centering 
\tiny
\setlength\tabcolsep{2pt}
\resizebox{\textwidth}{!}{
\begin{tabular}{@{}llc|cccc|ccccc:ccc:ccc|cc|cl@{}}
 \toprule
\textbf{Dataset}     & \textbf{Metric}   & \textbf{Random}   & \textbf{mBERT}    & \textbf{XLM-R}    & \textbf{Bernice}  & \textbf{InfoDCL}  & \textbf{BLOOM}    & \textbf{BLOOMZ}   & \multicolumn{1}{c}{\textbf{\begin{tabular}[c]{@{}c@{}}BLOOMZ\\ (MT)\end{tabular}}}  & \multicolumn{1}{c}{\textbf{\begin{tabular}[c]{@{}c@{}}BLOOMZ-\\ P3\end{tabular}}}    & \multicolumn{1}{c}{\textbf{\begin{tabular}[c]{@{}c@{}}BLOOMZ-\\ Bactrian\end{tabular}}}  & \textbf{mT5}  & \textbf{mT0}  & \multicolumn{1}{c}{\textbf{\begin{tabular}[c]{@{}c@{}}mT0\\ (MT)\end{tabular}}}     & \textbf{LLaMA}    & \textbf{Alpaca}   & \textbf{Vicuna}   & \textbf{ChatGPT}  & \multicolumn{1}{c}{\textbf{\begin{tabular}[c]{@{}c@{}}ChatGPT\\ (MT)\end{tabular}}}     & \textbf{SoTA}     & \textbf{SoTA study}\\ \midrule
Subj-eng\textsubscript{Pan}      & Acc.      & 48.60     & 95.00     & 95.13     & 94.53     & \textbf{95.20}    & 56.80     & 49.00     & 49.00     & 49.00     & 51.00     & 49.60     & 48.80     & 48.80     & 52.80     & 49.60     & 50.60     & 73.40     & 73.40     & \underline{97.10}     & \citet{chen-2022-dual}   \\ 
Subj-kor\textsubscript{Jan}      & M-F1      & 10.74     & 36.53     & 36.99     & \textbf{37.78}    & 36.95     & 2.33      & 3.15      & 3.63      & 10.80     & 3.94      & 9.85      & 12.80     & 1.93      & 6.55      & 7.17      & 12.76     & 27.41     & 15.14     & -     & ---  \\ 
Subj-ita\textsubscript{Bas}      & M-F1      & 43.30     & 74.02     & 77.18     & 77.71     & \textbf{80.16}    & 50.00     & 17.76     & 19.57     & 17.76     & 47.24     & 20.68     & 54.09     & 43.82     & 46.67     & 18.89     & 43.88     & 72.48     & 61.84     & \underline{71.40}     & \citet{basile2014overview}   \\ 
Subj-ita\textsubscript{Bas}      & M-F1      & 48.51     & 68.10     & 71.34     & 71.65     & 72.36     & 50.14     & 26.14     & 27.63     & 26.14     & 48.18     & 43.43     & 47.04     & 41.79     & 47.24     & 26.14     & 39.55     & \textbf{73.51}    & 60.37     & \underline{74.44}     & \citet{barbieri2016overview} \\ 
Subj-spa\textsubscript{Bar}      & M-F1      & 51.13     & 70.95     & 73.92     & 75.54     & \textbf{77.33}    & 53.67     & 27.85     & 31.51     & 27.85     & 40.33     & 47.72     & 42.94     & 27.85     & 53.33     & 27.85     & 38.60     & 71.38     & 71.21     & \underline{74.44}     & \citet{barbieri2016overview} \\ 
Subj-ces\textsubscript{Pri}      & Acc.      & 46.20     & 90.67     & 92.13     & 91.60     & \textbf{92.40}    & 51.80     & 52.80     & 52.80     & 52.80     & 47.20     & 52.80     & 44.20     & 52.80     & 47.20     & 53.00     & 47.00     & 79.40     & 74.00     & 93.56     & \citet{priban-2022-czech}   \\ \midrule
Average      & ---   & 41.41     & 72.54     & 74.45     & 74.80     & \textbf{75.73}     & 44.12     & 29.45     & 30.69     & 30.73     & 39.65     & 37.35     & 41.64     & 36.16     & 42.30     & 30.44     & 38.73     & 66.26     & 59.33     & ---   & ---  \\ 
\bottomrule 
\end{tabular}
}
\caption{Full test results on subjectivity analysis. \textbf{SoTA:} Previous SoTA performance on each respective dataset. \textbf{Underscore} indicates that we have different data splits to the SoTA model. Best model of each dataset is in \textbf{bold}. }\label{tab:subjectivity_res}
\end{table*}

%% file: table_fig/language_wise_results.tex
\begin{table*}[t]
\centering
% \tiny
% \setlength\tabcolsep{3pt}
% \renewcommand*{\arraystretch}{0.9}   %to squeeze table content
\resizebox{0.6\linewidth}{!}{
\begin{tabular}{@{}llc|c|ccc:cc@{}}
\toprule
\multicolumn{1}{c}{\textbf{Lang Fam.}} & \multicolumn{1}{c}{\textbf{Lang}} & \textbf{Random} & \textbf{InfoDCL} & \textbf{BMZ-P3} & \textbf{mT0} & \textbf{Vicuna} & \textbf{CG} & \textbf{CG-MT} \\  \hline
\multirow{7}{*}{Afro-Asiatic}          & ara                               & 34.05         & \bf 73.53         & 36.78           & 35.61        & \textcolor{red}{33.61}         & \colorbox{green!30}{60.81}       & 52.53          \\
                                       & amh                               & 37.95         & \bf 65.68         & \textcolor{red}{16.05}           & \textcolor{red}{22.49}        & \textcolor{red}{2.99}          & \textcolor{red}{20.62}       & \colorbox{green!30}{46.82}          \\
                                       & arq                               & 34.23         & \bf 71.25         & 52.02           & \textcolor{red}{18.05}        & \textcolor{red}{5.33}          & 63.89       & \colorbox{green!30}{67.58}          \\
                                       & ary                               & 35.94         & \bf 53.44         & 37.40           & \textcolor{red}{23.41}        & \textcolor{red}{16.64}         & \colorbox{green!30}{52.19}       & 51.66          \\
                                           & hau                               & 35.22         & \bf 72.18         & \textcolor{red}{30.14}           & \textcolor{red}{20.93}        & \textcolor{red}{17.39}         & \colorbox{green!30}{55.52}       & \textcolor{red}{34.13}          \\
                                       & heb                               & 47.60         & \bf 95.80         & 71.20           & 76.60        & \textcolor{red}{40.80}         & \colorbox{green!30}{84.20}       & 57.40          \\
                                       & mlt                               & 47.51         & 68.01         & 47.70           & \textcolor{red}{39.25}        & 48.98         & \bf 78.47       & \colorbox{green!30}{77.45}          \\ \hline
\multirow{7}{*}{Atlantic-C.}           & bam                               & 27.61         & \bf 65.57         & 36.70           & \textcolor{red}{24.31}        & \textcolor{red}{14.40}         & \colorbox{green!30}{40.27}       & 37.11          \\
                                       & ibo                               & 32.90         & \bf 76.75         & \textcolor{red}{23.21}           & \textcolor{red}{11.52}        & \textcolor{red}{28.37}         & \colorbox{green!30}{57.55}       & 33.46          \\
                                       & kin                               & 35.83         & \bf 56.69         & \textcolor{red}{28.06}           & \textcolor{red}{14.84}        & \textcolor{red}{22.23}         & \colorbox{green!30}{53.78}       & \textcolor{red}{29.03}          \\
                                       & swh                               & 33.72         & \bf 61.29         & \textcolor{red}{17.60}           & \textcolor{red}{14.02}        & 45.55         & \colorbox{green!30}{54.39}       & 53.84          \\
                                       & twi                               & 34.04         & \bf 64.51         & 46.56           & \textcolor{red}{29.34}        & \textcolor{red}{5.28}          & \colorbox{green!30}{51.12}       & \textcolor{red}{32.06}          \\
                                       & tso                               & 34.03         & \bf 52.55         & \colorbox{green!30}{45.54}           & \textcolor{red}{30.74}        & \textcolor{red}{6.50}          & 42.58       & 35.70          \\
                                       & yor                               & 41.68         & \bf 74.88         & 43.10           & \textcolor{red}{35.83}        & \textcolor{red}{26.38}         & \colorbox{green!30}{64.77}       & 42.31          \\ \hline
Austroasi.                             & vie                               & 16.12         & \bf 64.58         & \textcolor{red}{12.22}           & 27.81        & \textcolor{red}{9.52}          & \colorbox{green!30}{54.69}       & 32.96          \\  \hline
\multirow{12}{*}{Austrones.}           & ace                               & 34.78         & \bf 77.36         & 37.89           & \textcolor{red}{24.73}        & \textcolor{red}{12.79}         & 52.63       & \colorbox{green!30}{58.05}          \\
                                       & ban                               & 30.14         & \bf 79.49         & 41.90           & \textcolor{red}{29.91}        & \textcolor{red}{13.82}         & \colorbox{green!30}{60.91}       & 42.28          \\
                                       & bjn                               & 30.77         & \bf 84.50         & 50.51           & \textcolor{red}{27.78}        & \textcolor{red}{14.89}         & 69.34       & \colorbox{green!30}{75.43}          \\
                                       & bug                               & 30.77         & \bf 71.55         & 34.60           & \textcolor{red}{18.27}        & \textcolor{red}{12.90}         & \colorbox{green!30}{34.63}       & 30.86          \\
                                       & fil                               & 52.37         & \bf 79.01         & \textcolor{red}{34.47}           & \textcolor{red}{34.47}        & \textcolor{red}{34.47}         & \colorbox{green!30}{69.13}       & 66.67          \\
                                       & ind                               & 22.85         & \bf 83.05         & 42.86           & 36.77        & \textcolor{red}{20.69}         & \colorbox{green!30}{75.29}       & 64.14          \\
                                       & jav                               & 31.62         & \bf 84.79         & 48.06           & 37.65        & \textcolor{red}{15.21}         & 73.03       & \colorbox{green!30}{78.56}          \\
                                       & mad                               & 28.64         & \bf 78.36         & 45.44           & \textcolor{red}{21.85}        & \textcolor{red}{13.14}         & 61.07       & \colorbox{green!30}{61.14}          \\
                                       & min                               & 34.41         & \bf 84.07         & 49.93           & \textcolor{red}{32.41}        & \textcolor{red}{14.95}         & \colorbox{green!30}{69.80}       & 62.91          \\
                                       & nij                               & 34.86         & \bf 77.22         & 42.86           & \textcolor{red}{22.89}        & \textcolor{red}{15.21}         & \colorbox{green!30}{57.64}       & 57.07          \\
                                       & sun                               & 32.18         & \bf 81.71         & 44.83           & 37.65        & \textcolor{red}{12.93}         & 64.97       & \colorbox{green!30}{68.76}          \\
                                       & bbc                               & 30.60         & \bf 73.58         & 36.42           & \textcolor{red}{19.20}        & \textcolor{red}{13.86}         & 38.43       & \colorbox{green!30}{40.65}          \\ \hline
\multirow{4}{*}{Dravidian}             & kan                               & 32.14         & \bf 61.79         & 39.23           & \textcolor{red}{27.80}        & \textcolor{red}{19.12}         & \colorbox{green!30}{44.81}       & \textcolor{red}{30.39}          \\
                                       & mal                               & 31.68         & \bf 82.70         & 43.84           & 41.65        & \textcolor{red}{24.85}         & \colorbox{green!30}{44.03}       & \textcolor{red}{31.44}          \\
                                       & tam                               & 29.45         & \bf 57.81         & 38.53           & 34.27        & \textcolor{red}{17.60}         & \colorbox{green!30}{45.85}       & 33.20          \\
                                       & tel                               & 32.29         & \bf 59.61         & 40.99           & 35.15        & 36.61         & \colorbox{green!30}{53.33}       & 38.68          \\ \hline
\multirow{26}{*}{Indo-Euro.}           & sqi                               & 29.39         & \bf 47.04         & 31.78           & \textcolor{red}{26.59}        & \textcolor{red}{16.11}         & \colorbox{green!30}{46.82}       & 33.84          \\
                                       & bos                               & 34.37         & \bf 68.31         & \textcolor{red}{31.80}           & \textcolor{red}{21.16}        & \textcolor{red}{16.72}         & \colorbox{green!30}{63.22}       & 48.15          \\
                                       & bul                               & 32.48         & \bf 65.09         & \textcolor{red}{27.35}           & \textcolor{red}{19.37}        & \textcolor{red}{23.69}         & \colorbox{green!30}{59.32}       & 54.39          \\
                                       & ben                               & 24.53         & \bf 69.49         & 24.71           & 25.18        & \textcolor{red}{15.62}         & \colorbox{green!30}{53.25}       & 37.33          \\
                                       & hrv                               & 30.23         & \bf 67.83         & 35.18           & \textcolor{red}{27.47}        & \textcolor{red}{13.67}         & \colorbox{green!30}{62.02}       & 55.66          \\
                                       & ces                               & 41.06         & \bf 80.15         & 50.91           & 46.61        & 51.00         & \colorbox{green!30}{65.30}       & 63.24          \\
                                       & dan                               & 41.14         & \bf 82.09         & 46.68           & 46.68        & 51.84         & \colorbox{green!30}{66.91}       & 66.79          \\
                                       & eng                               & 37.90         & \bf 75.48         & 43.32           & 39.23        & 39.75         & \colorbox{green!30}{66.51}       & ---          \\
                                       & fra                               & 24.65         & \bf 66.58         & \textcolor{red}{23.29}           & 32.20        & 30.46         & \colorbox{green!30}{62.08}       & 56.84          \\
                                       & deu                               & 24.40         & \bf 67.55         & \textcolor{red}{20.00}           & 26.51        & 29.31         & \colorbox{green!30}{52.24}       & 35.41          \\
                                       & ell                               & 41.24         & \bf 79.13         & 46.71           & 45.47        & 48.21         & \colorbox{green!30}{60.94}       & \textcolor{red}{34.98}          \\
                                       & hin                               & 35.24         & \bf 67.55         & \textcolor{red}{28.92}           & \textcolor{red}{26.20}        & \textcolor{red}{29.06}         & \colorbox{green!30}{52.63}       & 48.30          \\
                                       & ita                               & 39.83         & \bf 74.78         & \textcolor{red}{38.76}           & 41.45        & 45.05         & \colorbox{green!30}{70.29}       & 64.37          \\
                                       & mar                               & 29.60         & \bf 86.47         & 43.80           & 35.40        & 34.40         & \colorbox{green!30}{68.60}       & 50.20          \\
                                       & pcm                               & 33.36         & \bf 66.65         & 47.16           & \textcolor{red}{24.25}        & \textcolor{red}{13.92}         & \colorbox{green!30}{61.13}       & 42.75          \\
                                       & nor                               & 16.24         & \bf 42.36         & 19.84           & 35.73        & \textcolor{red}{9.80}          & \colorbox{green!30}{42.11}       & 41.41          \\
                                       & fas                               & 33.03         & \bf 62.43         & 35.01           & 38.73        & \textcolor{red}{26.64}         & \colorbox{green!30}{55.02}       & 51.17          \\
                                       & por                               & 29.17         & \bf 66.10         & \textcolor{red}{24.45}           & \textcolor{red}{20.14}        & \textcolor{red}{19.16}         & \colorbox{green!30}{41.10}       & 39.37          \\
                                       & pol                               & 30.43         & \bf 68.04         & 31.82           & 31.96        & \textcolor{red}{19.15}         & \colorbox{green!30}{58.25}       & 49.43          \\
                                       & ron                               & 38.16         & \bf 89.83         & 48.97           & 67.14        & \textcolor{red}{30.54}         & \colorbox{green!30}{89.11}       & 77.34          \\
                                       & rus                               & 31.85         & \bf 84.31         & \textcolor{red}{25.49}           & \textcolor{red}{28.33}        & \textcolor{red}{28.51}         & \colorbox{green!30}{71.34}       & 69.08          \\
                                       & spa                               & 35.98         & \bf 70.20         & \textcolor{red}{30.80}           & \textcolor{red}{32.17}        & 37.71         & \colorbox{green!30}{57.16}       & 56.64          \\
                                       & srp                               & 34.09         & \bf 56.85         & \textcolor{red}{27.45}           & \textcolor{red}{20.55}        & \textcolor{red}{19.05}         & \colorbox{green!30}{55.84}       & 52.54          \\
                                       & slk                               & 28.57         & \bf 75.71         & 31.89           & \textcolor{red}{27.85}        & \textcolor{red}{10.46}         & \colorbox{green!30}{57.46}       & 56.18          \\
                                       & slv                               & 26.32         & \bf 58.62         & \textcolor{red}{14.37}           & \textcolor{red}{16.84}        & \textcolor{red}{16.64}         & \colorbox{green!30}{46.25}       & 36.97          \\
                                       & swe                               & 33.35         & \bf 70.98         & \textcolor{red}{29.31}           & \textcolor{red}{20.77}        & \textcolor{red}{16.05}         & 60.10       & \colorbox{green!30}{62.58}          \\ \hline
Japonic                                & jpn                               & 18.00         & \bf 61.47         & 44.20           & 47.60        & 26.80         & \colorbox{green!30}{55.80}       & 44.40          \\ \hline
Koreanic                               & kor                               & 28.55         & \bf 57.46         & \textcolor{red}{19.48}           & \textcolor{red}{21.24}        & \textcolor{red}{16.26}         & \colorbox{green!30}{42.90}       & 37.06          \\ \hline
Sino-Tib.                              & zho                               & 36.77         & \bf 76.10         & 48.19           & 46.94        & 38.19         & \colorbox{green!30}{63.43}       & 61.39          \\ \hline
Tai-Kadai                              & tha                               & 31.26         & \bf 75.17         & \textcolor{red}{20.16}           & \textcolor{red}{23.94}        & \textcolor{red}{24.40}         & \colorbox{green!30}{52.81}       & 36.32          \\ \hline
Turkic                                 & tur                               & 29.56         & \bf 87.71         & 32.32           & 48.07        & 35.44         & \colorbox{green!30}{82.27}       & 64.84          \\ \hline
\multirow{2}{*}{Uralic}                & fin                               & 25.68         & \bf 63.85         & \textcolor{red}{15.84}           & 30.45        & \textcolor{red}{23.08}         & \colorbox{green!30}{63.17}       & 57.48          \\
                                       & hun                               & 27.80         & \bf 68.37         & 30.31           & \textcolor{red}{27.19}        & \textcolor{red}{16.39}         & \colorbox{green!30}{53.22}       & 43.89          \\ \bottomrule
\end{tabular} 
}
% \vspace{-3pt}
\caption{Language-wise model performance. The best performance in each language is \textbf{bold}, and the second best is in \colorbox{green!30}{green highlight}. The \textcolor{red}{red font} denotes a performance lower than the random baseline. \textbf{BMZ:} BLOOMZ, \textbf{CG:} ChatGPT, \textbf{MT:} using machine translated prompts.} \label{tab:full_lanuage_wise}
\end{table*}

%% file: table_fig/prompt_sensitive.tex
\begin{table*}[h]
\centering
\tiny
\begin{tabular}{@{}lcccccc|cccccc@{}}
\toprule
\multicolumn{1}{c}{\multirow{2}{*}{\textbf{Task}}} & \multicolumn{6}{c}{\textbf{lm-evaluation-harness Prompts}}                                                                                                                                                                & \multicolumn{6}{c}{\textbf{ChatGPT Prompts}}                                                                                                                                                                              \\ \cmidrule(l){2-7}\cmidrule(l){8-13} 
\multicolumn{1}{c}{}                               & \multicolumn{1}{c}{\textbf{BLOOM}} & \multicolumn{1}{c}{\textbf{BLOOMZ}} & \multicolumn{1}{c}{\textbf{mT5}} & \multicolumn{1}{c}{\textbf{mT0}} & \multicolumn{1}{c}{\textbf{LLaMA}} & \multicolumn{1}{c}{\textbf{Vicuna}} & \multicolumn{1}{c}{\textbf{BLOOM}} & \multicolumn{1}{c}{\textbf{BLOOMZ}} & \multicolumn{1}{c}{\textbf{mT5}} & \multicolumn{1}{c}{\textbf{mT0}} & \multicolumn{1}{c}{\textbf{LLaMA}} & \multicolumn{1}{c}{\textbf{Vicuna}} \\ \midrule
Hate                                               & 39.83                              & 38.52                               & 23.29                            & 37.33                            & 37.80                              & \textbf{41.59}                      & 18.39                              & 31.34                               & 28.96                            & 38.19                            & 18.37                              & 37.33                               \\
Emotion                                            & 9.71                               & 15.07                               & 7.75                             & 27.87                            & 15.14                              & 18.12                               & 8.61                               & 20.07                               & 7.57                             & \textbf{29.63}                   & 17.61                              & 8.65                                \\
Humor                                              & 41.78                              & 33.04                               & 43.60                            & 33.12                            & 39.78                              & \textbf{46.19}                      & 41.59                              & 44.99                               & 41.34                            & 34.13                            & 41.59                              & 33.12                               \\
Irony                                              & 36.63                              & 44.46                               & 36.52                            & 34.69                            & 40.78                              & \textbf{47.48}                      & 27.33                              & 26.02                               & 36.08                            & 34.31                            & 26.02                              & 34.70                               \\ \hline
\textbf{Aveage}                                    & 25.21                              & 28.01                               & 19.58                            & 32.12                            & 27.72                              & 31.80                               & 16.92                              & 26.14                               & 20.91                            & \textbf{33.21}                   & 20.95                              & 23.03                               \\ \bottomrule
\end{tabular}
\caption{Study on model sensitivity to prompts used for zero-shot evaluation.}\label{tab:prompt_sensitive}
\end{table*}